\documentclass{article}

\usepackage[final]{neurips_2021}

\usepackage[
	activate={true,nocompatibility},
	tracking=true
]{microtype}

\usepackage{geometry}
\usepackage{algorithm}
\usepackage{algorithmic}
\usepackage{graphicx}
\usepackage{subcaption}
\usepackage[utf8]{inputenc}
\usepackage{makecell}
\usepackage[T1]{fontenc}
\usepackage{hyperref}
\usepackage{url}
\usepackage{booktabs}
\usepackage{amsfonts}
\usepackage{nicefrac}
\usepackage{microtype}
\usepackage{xcolor}
\usepackage{xspace}
\usepackage{amsmath,amssymb,mathtools,amsthm}
\usepackage{multirow}
\usepackage{pifont}
\usepackage{xparse}
\usepackage{pythonhighlight}
\usepackage{array}
\newcolumntype{H}{>{\setbox0=\hbox\bgroup}c<{\egroup}@{}}

\DeclareCaptionSubType * [alph]{table}
\DeclarePairedDelimiterX{\infdivx}[2]{(}{)}{%
  #1\;\delimsize|\delimsize|\;#2%
}

\newcommand{\shortcite}[1]{[\citenum{#1}]}
\newcommand{\namecite}[1]{\citeauthor{#1} [\citenum{#1}]}
\newcommand{\remark}[1]{}
\setcitestyle{citesep={,}}
\newcommand{\modelname}{unCLIP}

\NewDocumentCommand{\grad}{e{_^}}{%
  \mathop{}\!
  \nabla
  \IfValueT{#1}{_{\!#1}}
  \IfValueT{#2}{^{#2}}
}

\title{Hierarchical Text-Conditional\\ Image Generation with CLIP Latents}

\author{
  Aditya Ramesh\thanks{Equal contribution} \\
  OpenAI \\
  \texttt{aramesh@openai.com} \\
  \And
  Prafulla Dhariwal\footnotemark[1] \\
  OpenAI \\
  \texttt{prafulla@openai.com} \\
  \And
  Alex Nichol\footnotemark[1]\\
  OpenAI \\
  \texttt{alex@openai.com}
  \And
  Casey Chu\footnotemark[1]\\
  OpenAI \\
  \texttt{casey@openai.com}
  \And
  Mark Chen\\
  OpenAI \\
  \texttt{mark@openai.com}
}

\begin{document}

\maketitle

\begin{abstract}
Contrastive models like CLIP have been shown to learn robust representations of images that capture both semantics and style. To leverage these representations for image generation, we propose a two-stage model: a prior that generates a CLIP image embedding given a text caption, and a decoder that generates an image conditioned on the image embedding. We show that explicitly generating image representations improves image diversity with minimal loss in photorealism and caption similarity. Our decoders conditioned on image representations can also produce variations of an image that preserve both its semantics and style, while varying the non-essential details absent from the image representation. Moreover, the joint embedding space of CLIP enables language-guided image manipulations in a zero-shot fashion. We use diffusion models for the decoder and experiment with both autoregressive and diffusion models for the prior, finding that the latter are computationally more efficient and produce higher-quality samples.
\end{abstract}

\section{Introduction}

Recent progress in computer vision has been driven by scaling models on large datasets of captioned images collected from the internet \shortcite{virtex,icmlm,convirt,clip,slip,cloob}. Within this framework, CLIP \shortcite{clip} has emerged as a successful representation learner for images. CLIP embeddings have a number of desirable properties: they are robust to image distribution shift, have impressive zero-shot capabilities, and have been fine-tuned to achieve state-of-the-art results on a wide variety of vision and language tasks \shortcite{clipbenefit}. Concurrently, diffusion models \shortcite{dickstein,improvedscore,ddpm} have emerged as a promising generative modeling framework, pushing the state-of-the-art on image and video generation tasks \shortcite{sotapaper,cascaded,uncond}. To achieve best results, diffusion models leverage a guidance technique \shortcite{sotapaper,uncond} which improves sample fidelity (for images, photorealism) at the cost of sample diversity.

In this work, we combine these two approaches for the problem of text-conditional image generation. We first train a diffusion \textit{decoder} to invert the CLIP image \textit{encoder}. Our inverter is non-deterministic, and can produce multiple images corresponding to a given image embedding. The presence of an encoder and its approximate inverse (the decoder) allows for capabilities beyond text-to-image translation. As in GAN inversion \shortcite{ganinversion,ganinversionsurvey}, encoding and decoding an input image produces semantically similar output images (Figure \ref{fig:variations_examples}). We can also interpolate between input images by inverting interpolations of their image embeddings (Figure \ref{fig:interpolations_examples}). However, one notable advantage of using the CLIP latent space is the ability to semantically modify images by moving in the direction of any encoded text vector (Figure \ref{fig:text_diffs}), whereas discovering these directions in GAN latent space involves
\newgeometry{left=0cm,top=0cm,right=0cm,bottom=0cm}
\begin{figure}[t!]
    \centering
    \setlength{\tabcolsep}{2.0pt}
    \begin{tabular}{ccc}
    \includegraphics[width=0.31\textwidth]{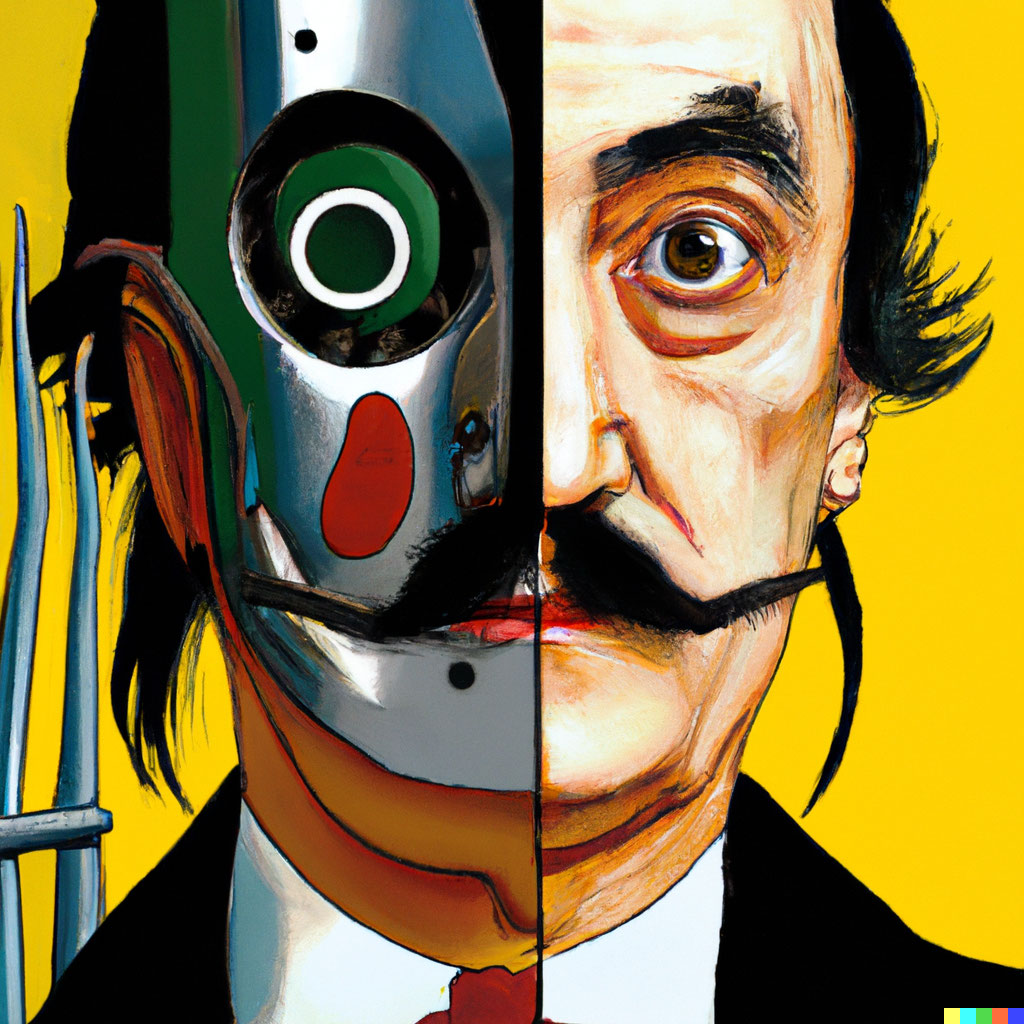} &
    \includegraphics[width=0.31\textwidth]{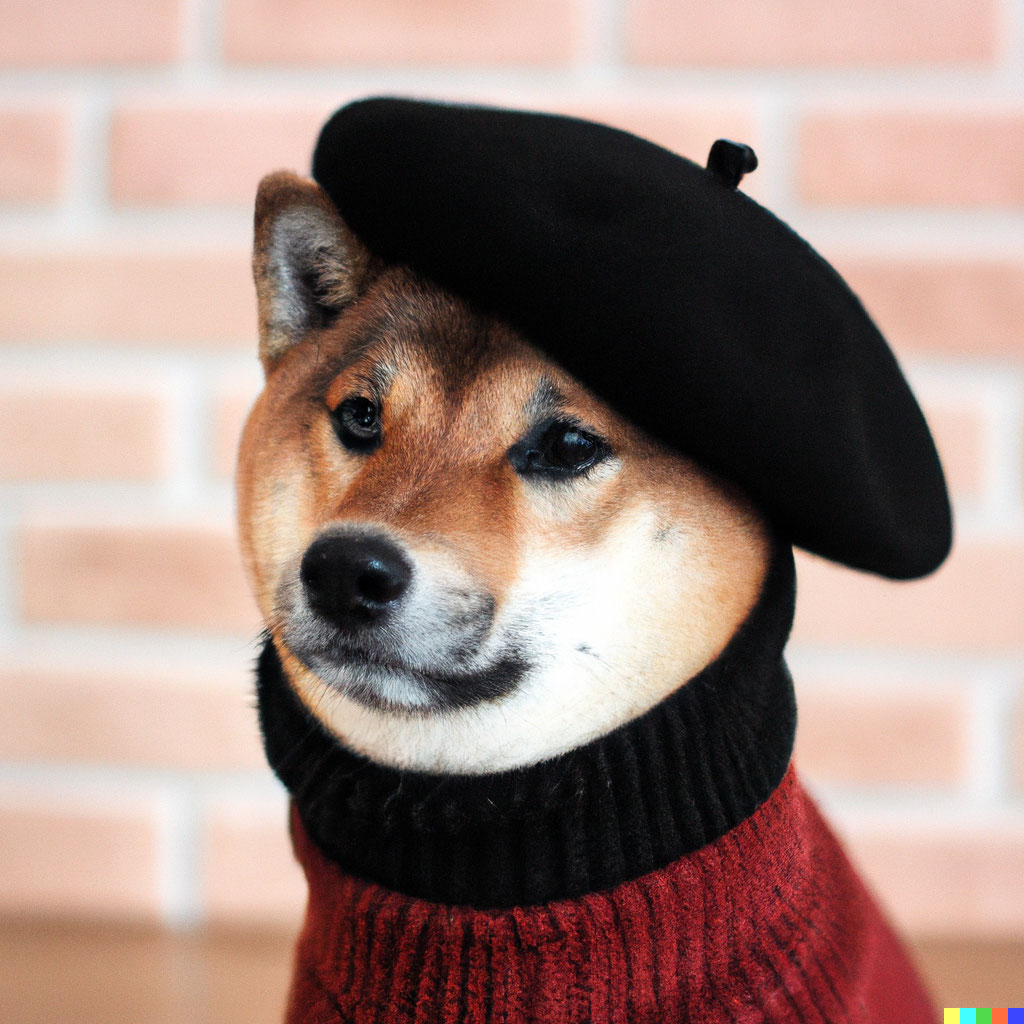} &
    \includegraphics[width=0.31\textwidth]{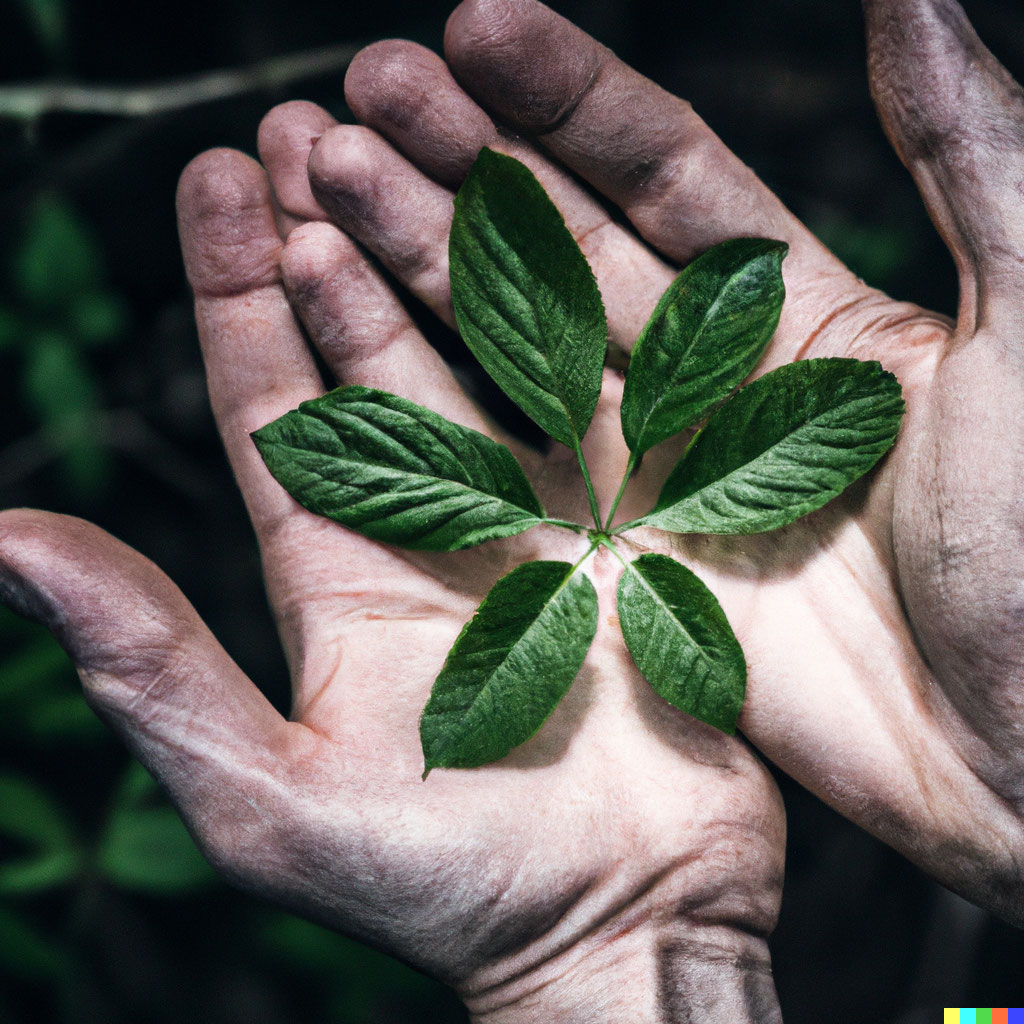} \\
    \scriptsize \makecell{vibrant portrait painting of Salvador Dalí with a robotic half face} &
    \scriptsize \makecell{a shiba inu wearing a beret and black turtleneck} &
    \scriptsize \makecell{a close up of a handpalm with leaves growing from it} \\
    \rule{0pt}{0.0pt} \\
    \includegraphics[width=0.31\textwidth]{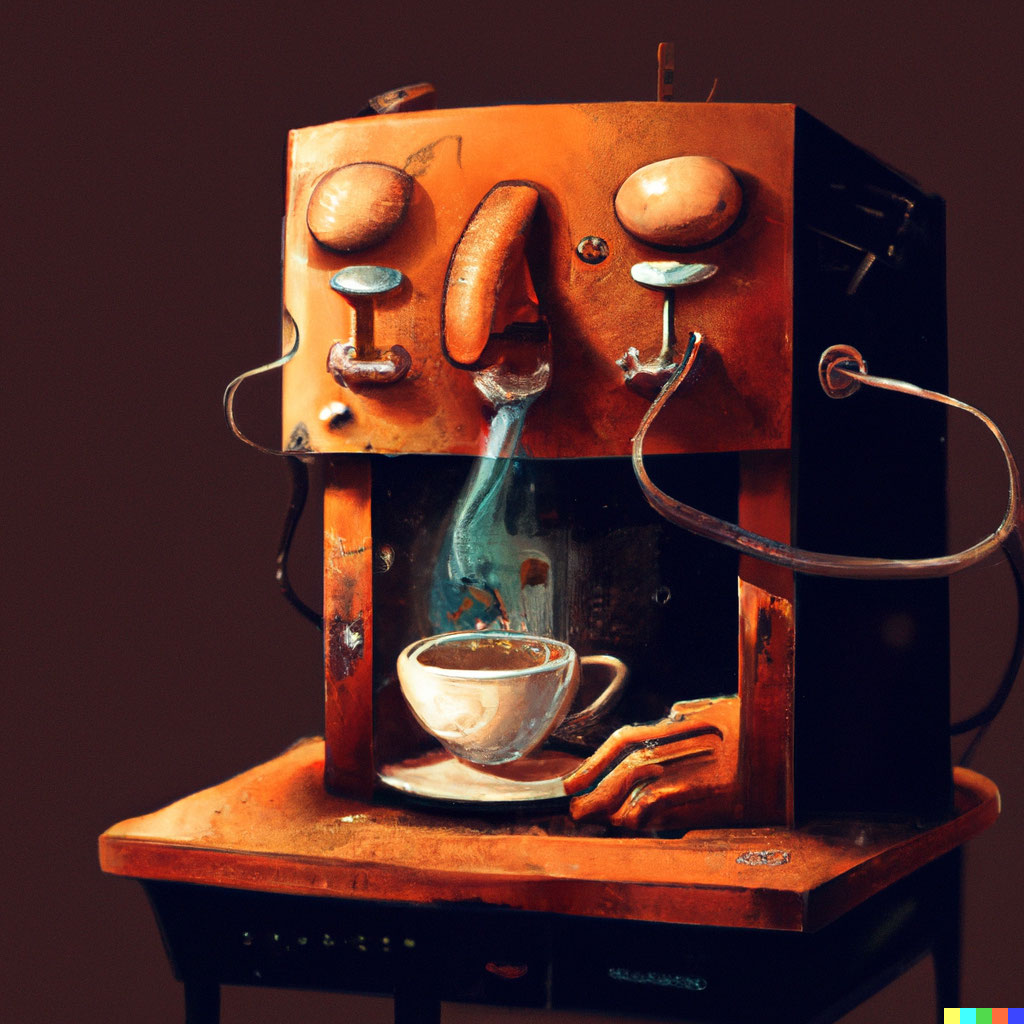} &
    \includegraphics[width=0.31\textwidth]{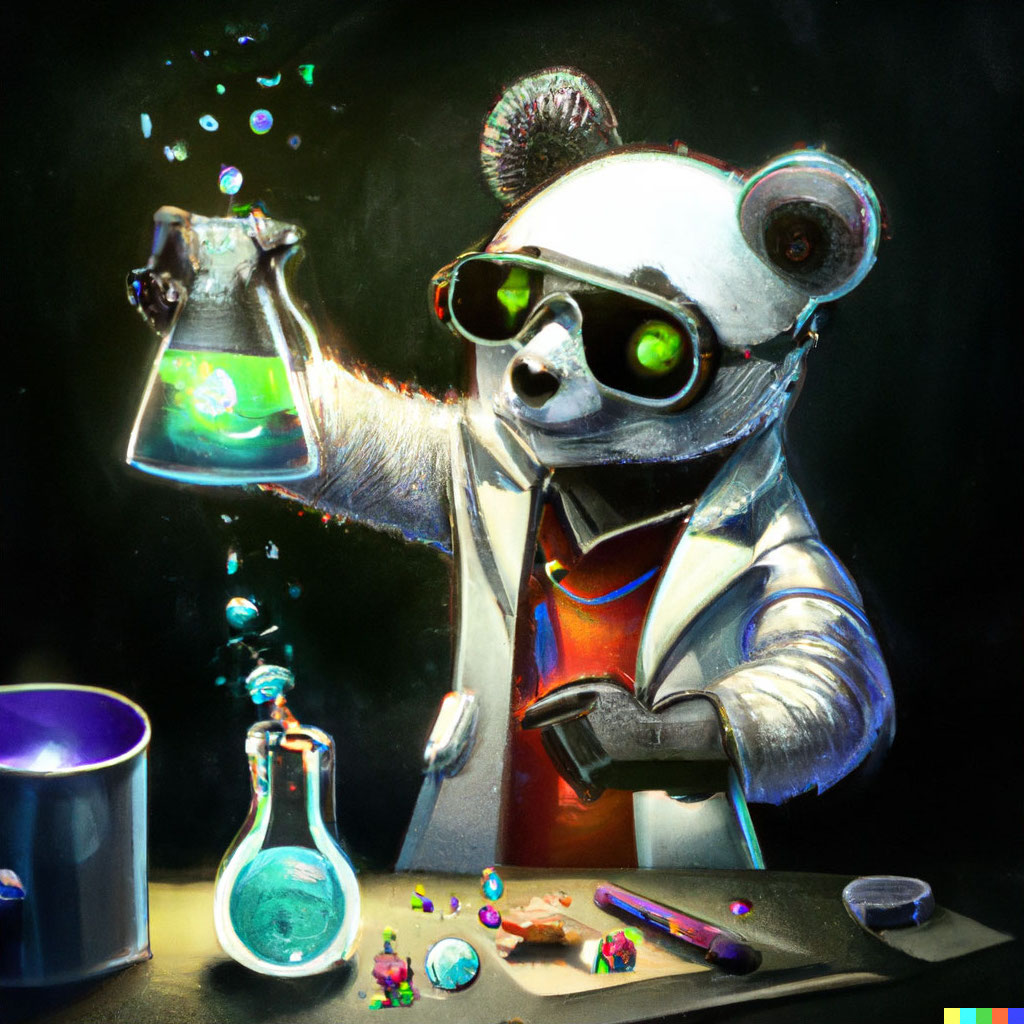} &
    \includegraphics[width=0.31\textwidth]{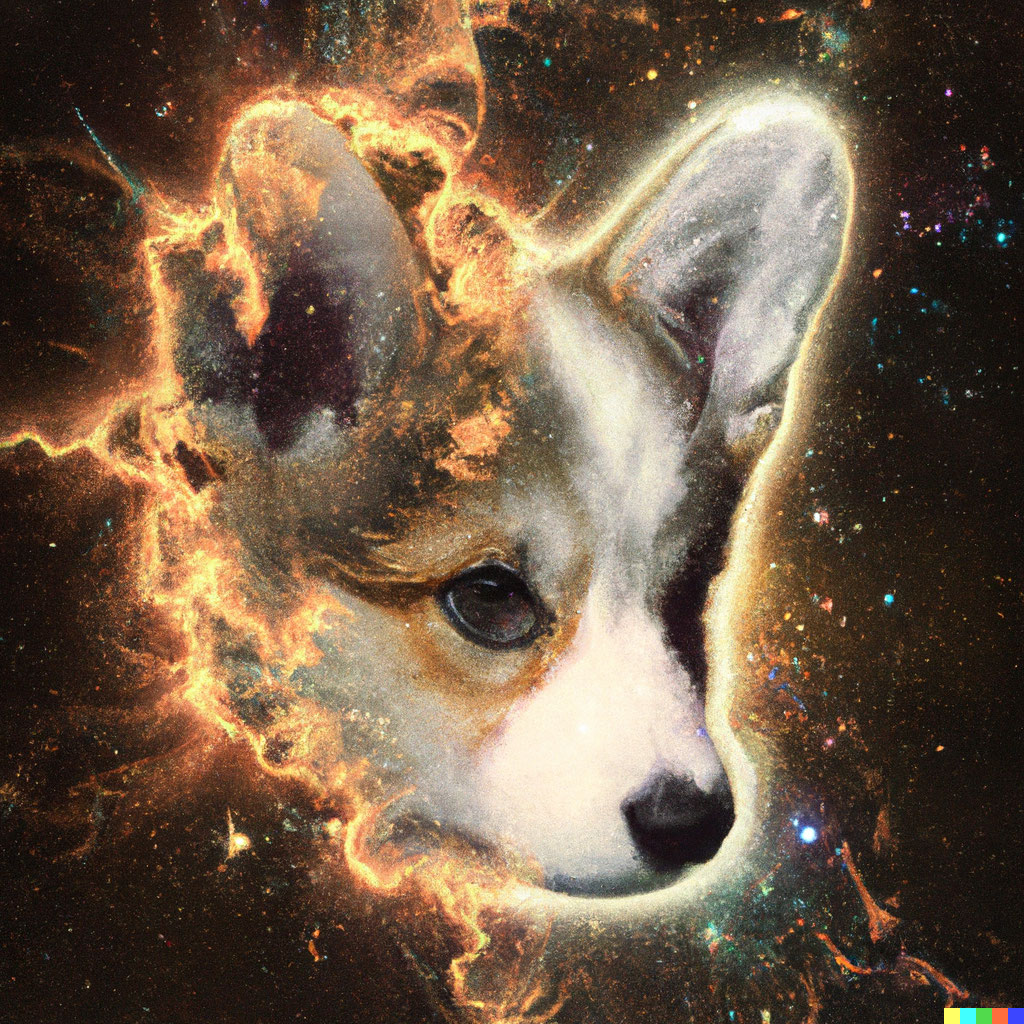} \\
    \scriptsize \makecell{an espresso machine that makes coffee from human souls, artstation} &
    \scriptsize \makecell{panda mad scientist mixing sparkling chemicals, artstation} &
    \scriptsize \makecell{a corgi's head depicted as an explosion of a nebula} \\
    \rule{0pt}{0.0pt} \\
    \includegraphics[width=0.31\textwidth]{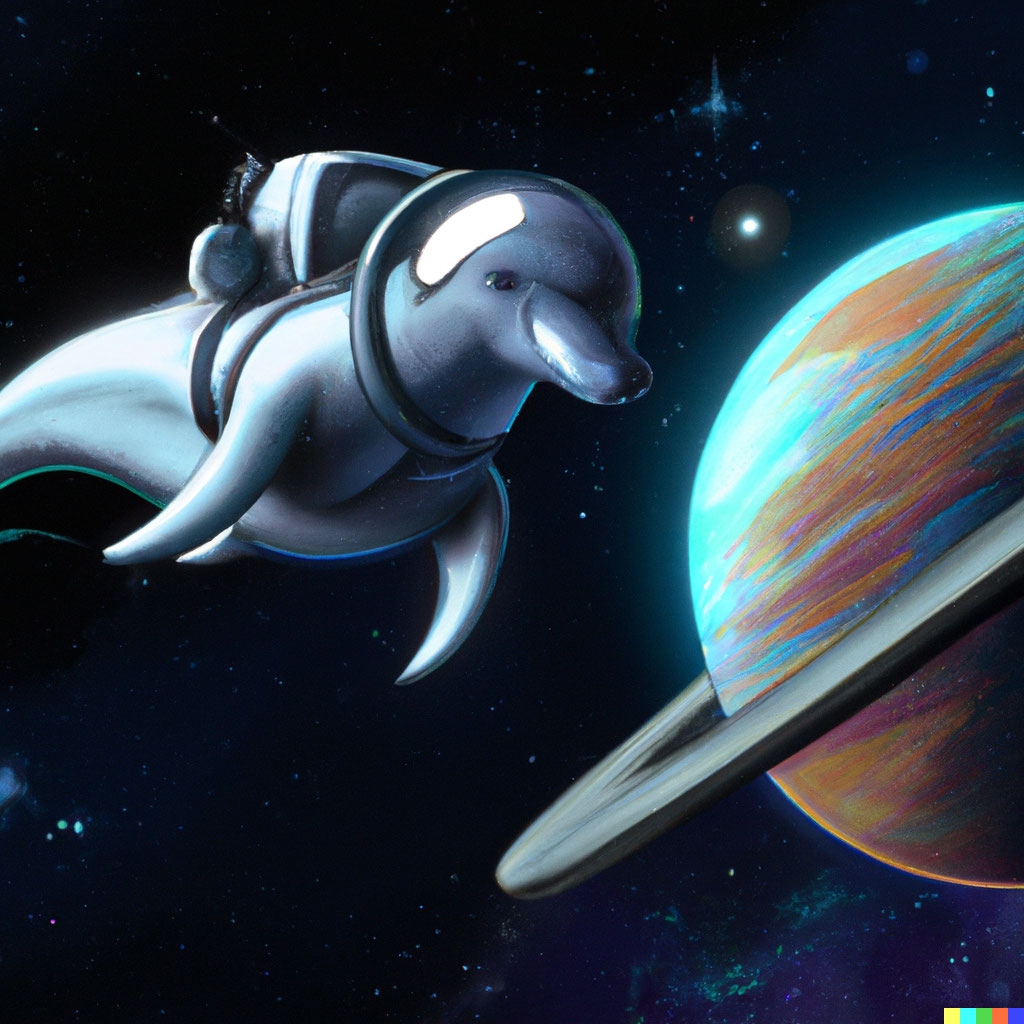} &
    \includegraphics[width=0.31\textwidth]{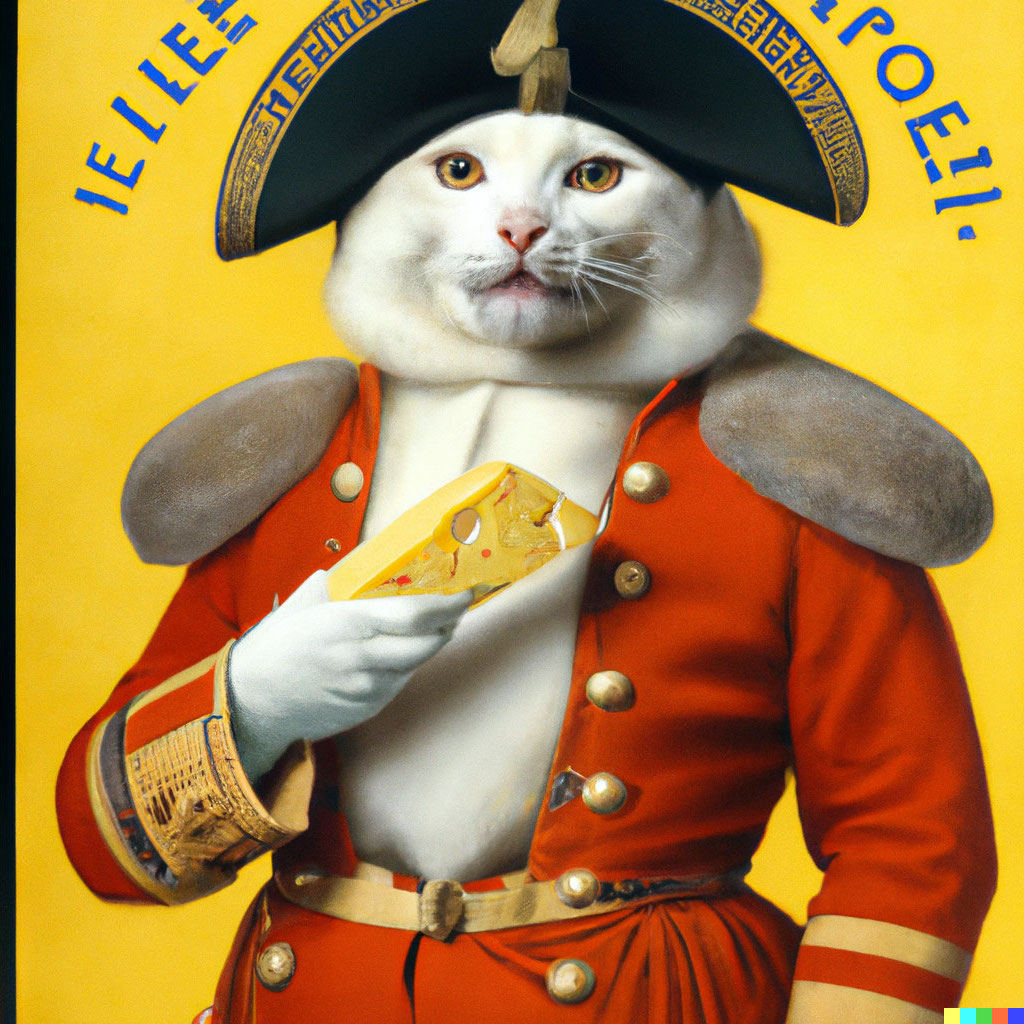} &
    \includegraphics[width=0.31\textwidth]{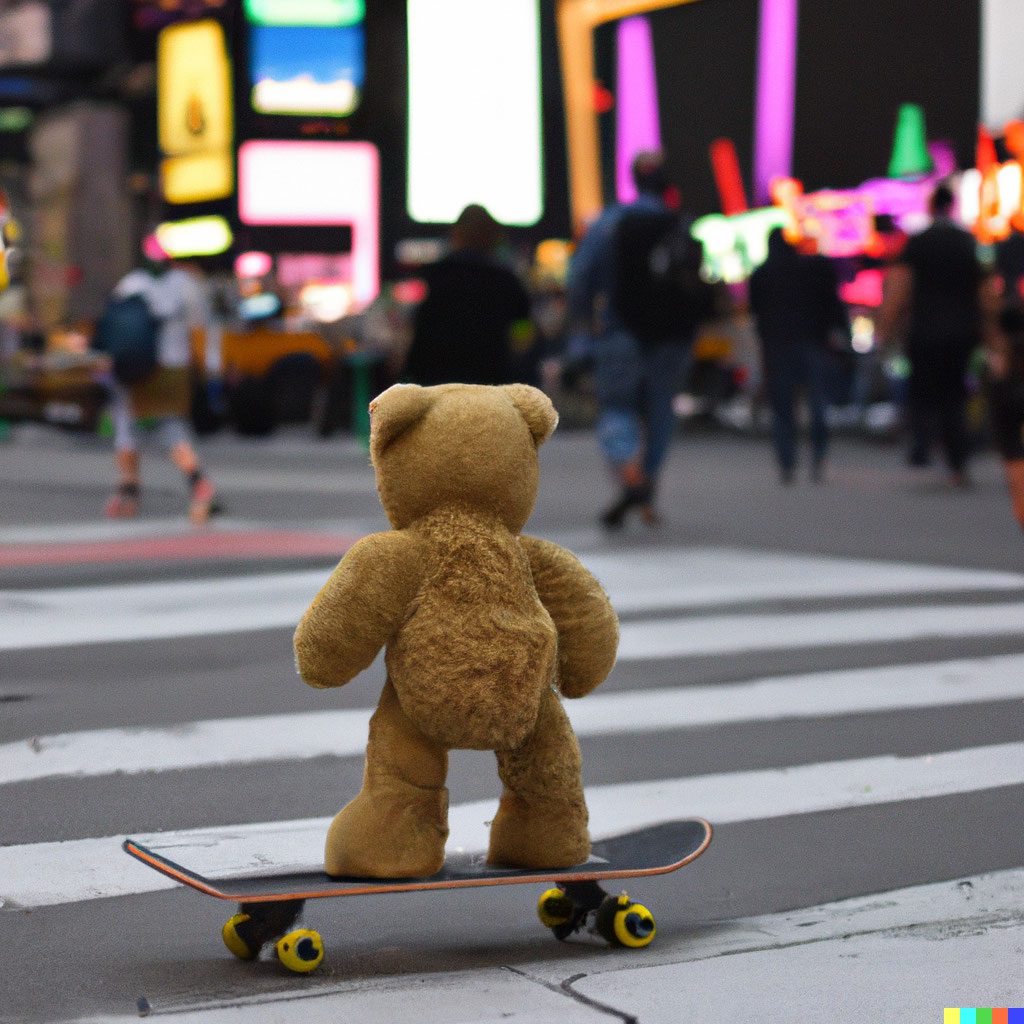} \\
    \scriptsize \makecell{a dolphin in an astronaut suit on saturn, artstation} &
    \scriptsize \makecell{a propaganda poster depicting a cat dressed as french emperor\\ napoleon holding a piece of cheese} &
    \scriptsize \makecell{a teddy bear on a skateboard in times square} \\
    \rule{0pt}{0.0pt} \\
    \end{tabular}
    
    \caption{Selected $1024 \times 1024$ samples from a production version of our model.}
    \label{fig:header_samples}
    \vskip -0.2in
\end{figure}%
\newpage%
\restoregeometry%
luck and diligent manual examination. Furthermore, encoding and decoding images also provides us with a tool for observing which features of the image are recognized or disregarded by CLIP.

To obtain a full generative model of images, we combine the CLIP image embedding \textit{decoder} with a \textit{prior} model, which generates possible CLIP image embeddings from a given text caption. We compare our text-to-image system with other systems such as DALL-E \shortcite{dalle} and GLIDE \shortcite{glide}, finding that our samples are comparable in quality to GLIDE, but with greater diversity in our generations. We also develop methods for training diffusion priors in latent space, and show that they achieve comparable performance to autoregressive priors, while being more compute-efficient. We refer to our full text-conditional image generation stack as \textit{\modelname{}}, since it generates images by inverting the CLIP image encoder.

\begin{figure}[t]
    \centering
    \setlength{\tabcolsep}{2.0pt}
    \includegraphics[width=0.88\textwidth]{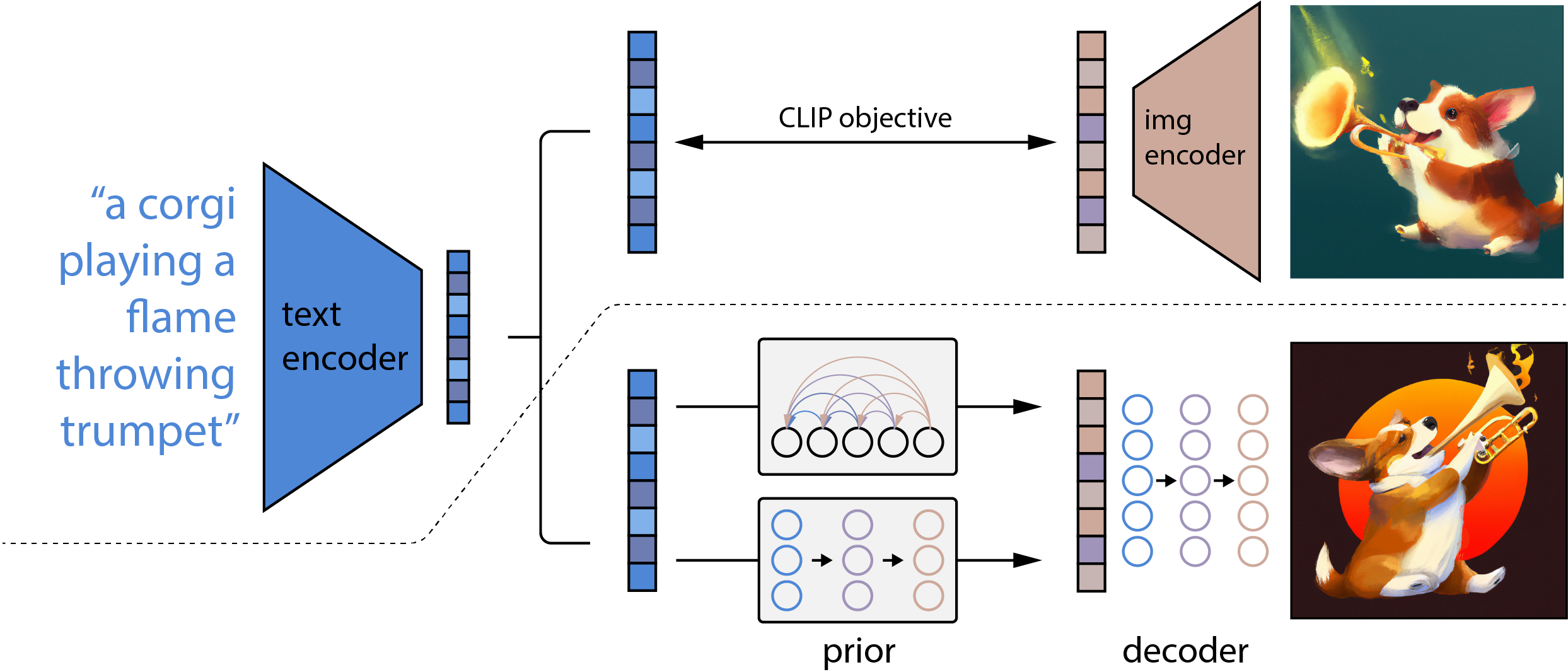}

    \caption{A high-level overview of \modelname{}. Above the dotted line, we depict the CLIP training process, through which we learn a joint representation space for text and images. Below the dotted line, we depict our text-to-image generation process: a CLIP text embedding is first fed to an autoregressive or diffusion prior to produce an image embedding, and then this embedding is used to condition a diffusion decoder which produces a final image. Note that the CLIP model is frozen during training of the prior and decoder.}
    \label{fig:figurehead}
    \vskip -0.1in 
\end{figure}

\section{Method}
Our training dataset consists of pairs~$(x,y)$ of images~$x$ and their corresponding captions~$y$. Given an image~$x$, let~$z_i$ and~$z_t$ be its CLIP image and text embeddings, respectively. We design our generative stack to produce images from captions using two components:

\begin{itemize}
    \item A \textit{prior} $P(z_i|y)$ that produces CLIP image embeddings $z_i$ conditioned on captions $y$.
    \item A \textit{decoder} $P(x|z_i,y)$ that produces images $x$ conditioned on CLIP image embeddings $z_i$ (and optionally text captions $y$). 
\end{itemize}

The decoder allows us to invert images given their CLIP image embeddings, while the prior allows us to learn a generative model of the image embeddings themselves. Stacking these two components yields a generative model $P(x|y)$ of images~$x$ given captions~$y$:
$$P(x|y) = P(x,z_i|y) = P(x|z_i,y) P(z_i|y).$$

The first equality holds because~$z_i$ is a deterministic function of~$x$. The second equality holds because of the chain rule. Thus, we can sample from the true conditional distribution $P(x|y)$ by first sampling~$z_i$ using the prior, and then sampling~$x$ using the decoder. In the following sections, we describe our decoder and prior stacks. For training details and hyperparameters, refer to Appendix \ref{app:hps}.

\subsection{Decoder}
We use diffusion models \shortcite{ddpm,improvedscore} to produce images conditioned on CLIP image embeddings (and optionally text captions). Specifically, we modify the architecture described in ~\cite{glide} by projecting and adding CLIP embeddings to the existing timestep embedding, and by projecting CLIP embeddings into four extra tokens of context that are concatenated to the sequence of outputs from the GLIDE text encoder. We retained the text conditioning pathway present in the original GLIDE model, hypothesizing that it could allow the diffusion model to learn aspects of natural language that CLIP fails to capture (e.g. variable binding), but find that it offers little help in this regard (Section~\ref{sec:limitations}). 

While we can sample from the conditional distribution of the decoder directly, past work using diffusion models shows using guidance on the conditioning information \shortcite{sotapaper,uncond,glide} improves sample quality a lot. We enable classifier-free guidance \shortcite{uncond} by randomly setting the CLIP embeddings to zero (or a learned embedding) 10\%~of the time, and randomly dropping the text caption 50\%~of the time during training. 

To generate high resolution images, we train two diffusion upsampler models \shortcite{improved,sr3}: one to upsample images from $64 \times 64$ to $256 \times 256$ resolution, and another to further upsample those to $1024 \times 1024$ resolution. To improve the robustness of our upsamplers, we slightly corrupt the conditioning images during training. For the first upsampling stage, we use gaussian blur~\shortcite{sr3}, and for the second, we use a more diverse BSR degradation \shortcite{latentdiffusion,bsr}. To reduce training compute and improve numerical stability, we follow \namecite{latentdiffusion} and train on random crops of images that are one-fourth the target size. We use only spatial convolutions in the model (i.e., no attention layers) and at inference time directly apply the model at the target resolution, observing that it readily generalizes to the higher resolution. We found no benefit from conditioning the upsamplers on the caption, and use unconditional ADMNets~\shortcite{sotapaper} with no guidance. 

\begin{figure*}[t!]
    \centering
    \setlength{\tabcolsep}{2.0pt}
    \begin{tabular}{cc}
        \includegraphics[width=0.16\textwidth]{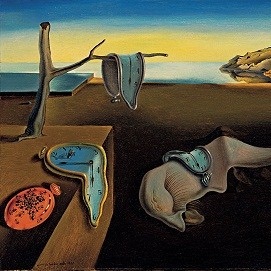} & 
        \includegraphics[width=0.16\textwidth]{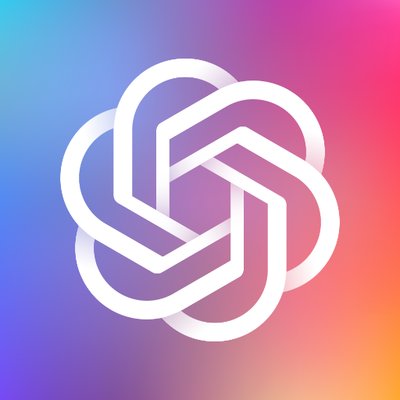} 
        \rule{0pt}{0.0pt} \\
        \includegraphics[width=0.48\textwidth]{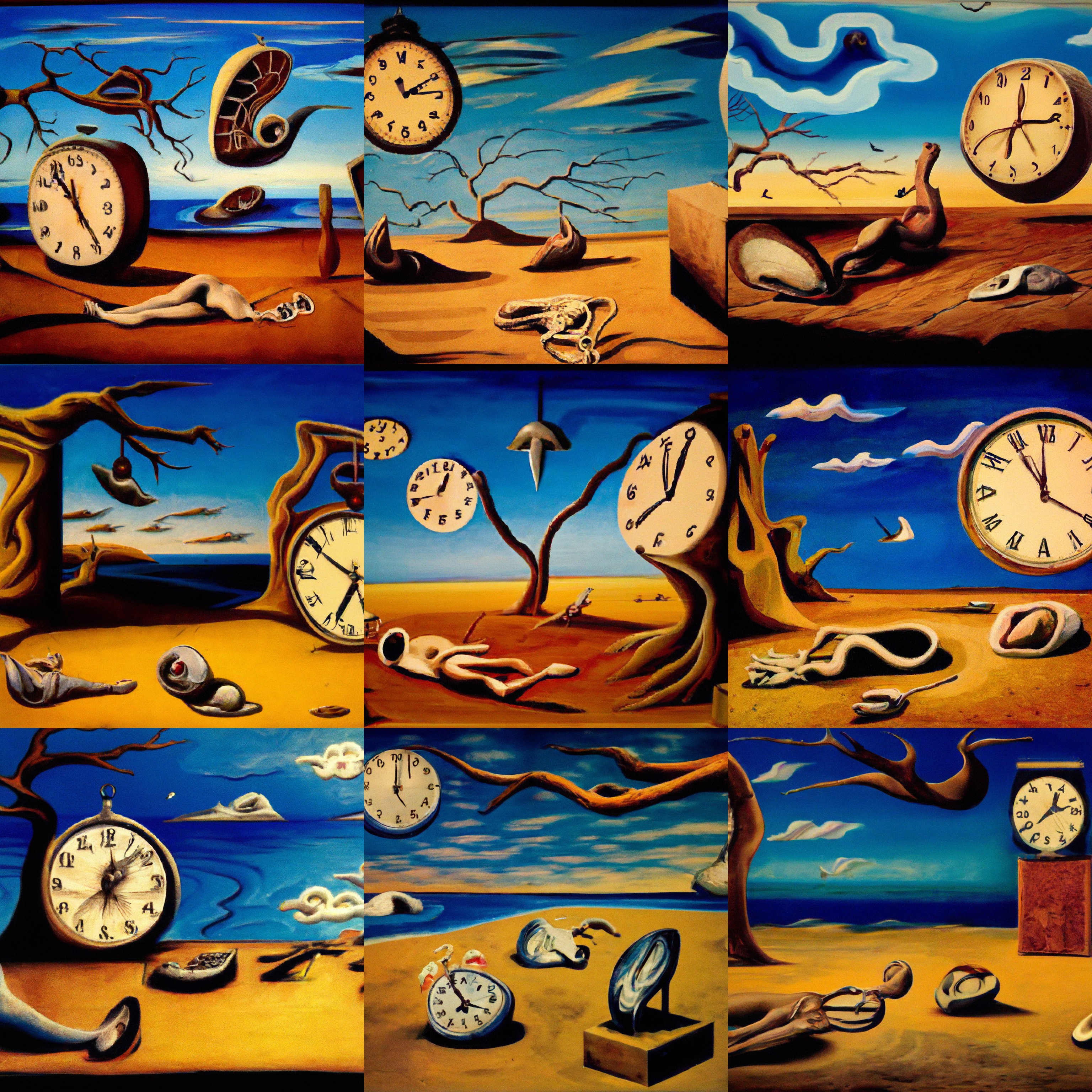} &
        \includegraphics[width=0.48\textwidth]{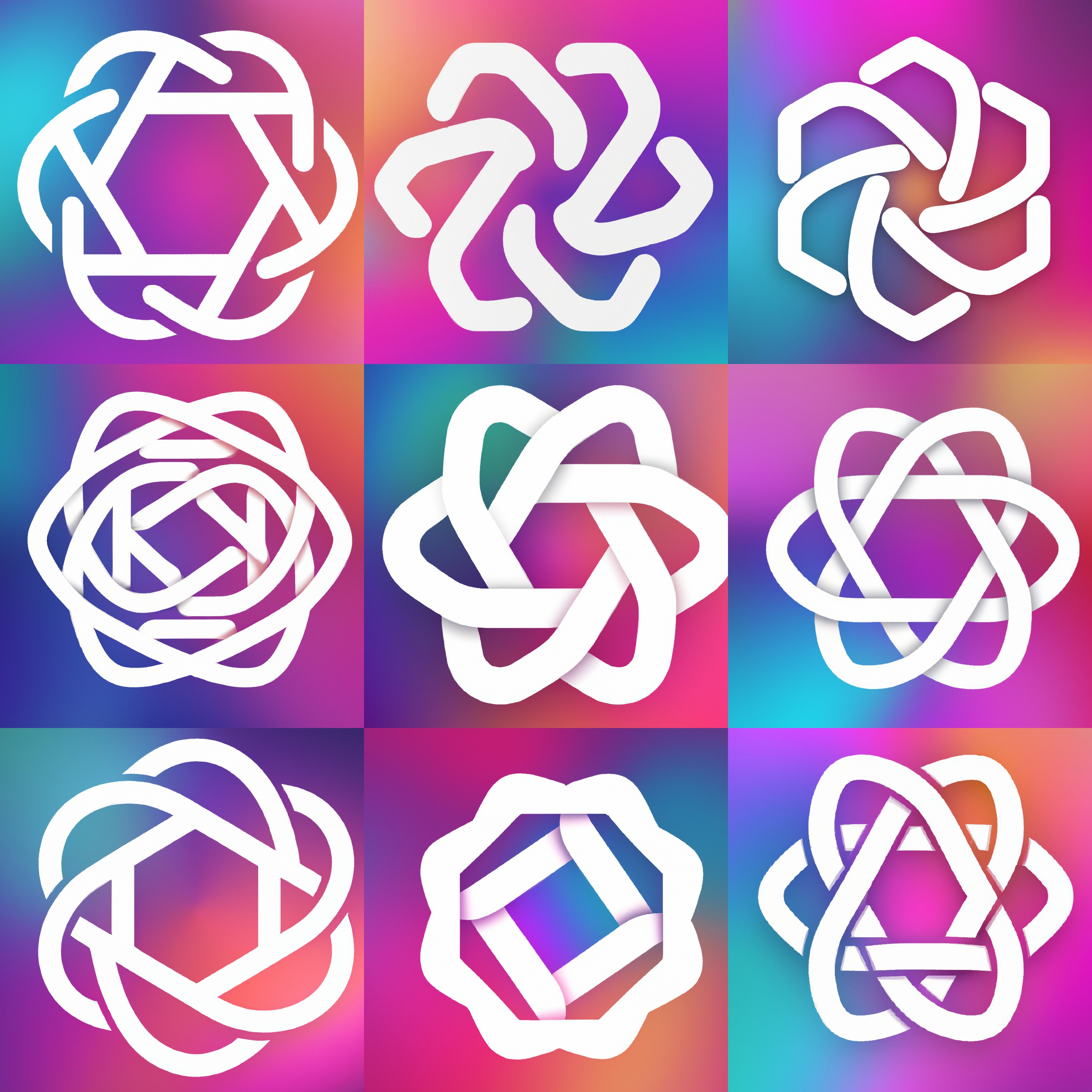} 
        \rule{0pt}{0.5pt}
    \end{tabular}

    \caption{Variations of an input image by encoding with CLIP and then decoding with a diffusion model. The variations preserve both semantic information like presence of a clock in the painting and the overlapping strokes in the logo, as well as stylistic elements like the surrealism in the painting and the color gradients in the logo, while varying the non-essential details.}
    \label{fig:variations_examples}
    \vskip -0.1in 
\end{figure*}

\subsection{Prior}
While a decoder can invert CLIP image embeddings~$z_i$ to produce images~$x$, we need a prior model that produces~$z_i$ from captions~$y$ to enable image generations from text captions. We explore two different model classes for the prior model:

\begin{itemize}
    \item \textit{Autoregressive (AR)} prior: the CLIP image embedding $z_i$ is converted into a sequence of discrete codes and predicted autoregressively conditioned on the caption $y$.
    \item \textit{Diffusion} prior: The continuous vector $z_i$ is directly modelled using a Gaussian diffusion model conditioned on the caption $y$.
\end{itemize}

In addition to the caption, we can condition the prior on the CLIP text embedding~$z_t$ since it is a deterministic function of the caption. To improve sample quality we also enable sampling using classifier-free guidance for both the AR and diffusion prior, by randomly dropping this text conditioning information 10\%~of the time during training.

To train and sample from the AR prior more efficiently, we first reduce the dimensionality of the CLIP image embeddings~$z_i$ by applying Principal Component Analysis~(PCA)~\shortcite{pca}. In particular, we find that the rank of the CLIP representation space is drastically reduced when training CLIP with SAM \shortcite{sam} while slightly improving evaluation metrics. We are able to preserve nearly all of the information\footnote{I.e., less than 1\% average mean-squared error in reconstructing the image representations.} by retaining only 319 principal components out of the original 1,024. After applying PCA, we order the principal components by decreasing eigenvalue magnitude, quantize each of the 319 dimensions into 1,024~discrete buckets, and predict the resulting sequence using a Transformer \shortcite{transformer} model with a causal attention mask. This results in a threefold reduction in the number of tokens predicted during inference, and improves training stability.

We condition the AR~prior on the text caption and the CLIP text embedding by encoding them as a prefix to the sequence. Additionally, we prepend a token indicating the (quantized) dot product between the text embedding and image embedding, $z_i \cdot z_t$. This allows us to condition the model on a higher dot product, since higher text-image dot products correspond to captions which better describe the image. In practice, we find it beneficial to sample the dot product from the top half of the distribution.\footnote{We swept over percentiles 50\%, 70\%, 85\%, 95\% and found 50\% to be optimal in all experiments.} 

For the diffusion prior, we train a decoder-only Transformer with a causal attention mask on a sequence consisting of, in order: the encoded text, the CLIP text embedding, an embedding for the diffusion timestep, the noised CLIP image embedding, and a final embedding whose output from the Transformer is used to predict the unnoised CLIP image embedding. We choose not to condition the diffusion prior on $z_i \cdot z_t$ like in the AR prior; instead, we improve quality during sampling time by generating two samples of $z_i$ and selecting the one with a higher dot product with~$z_t$. Instead of using the $\epsilon$-prediction formulation from \namecite{ddpm}, we find it better to train our model to predict the unnoised $z_i$ directly, and use a mean-squared error loss on this prediction:
$$L_{\text{prior}} = \mathbb{E}_{t \sim [1,T], z_i^{(t)} \sim q_t} \big[\|f_{\theta}(z_i^{(t)}, t, y) - z_i\|^2\big]$$

\section{Image Manipulations}

Our approach allows us to encode any given image~$x$ into a bipartite latent representation~$(z_i, x_T)$ that is sufficient for the decoder to produce an accurate reconstruction. The latent~$z_i$ describes the aspects of the image that are recognized by CLIP, while the latent~$x_T$ encodes all of the residual information necessary for the decoder to reconstruct~$x$. The former is obtained by simply encoding the image with the CLIP image encoder. The latter is obtained by applying DDIM inversion (Appendix~F in \shortcite{sotapaper}) to~$x$ using the decoder, while conditioning on~$z_i$. We describe three different kinds of manipulations that are enabled by this bipartite representation.

    

\subsection{Variations}
\label{sec:variations}

Given an image~$x$, we can produce related images that share the same essential content but vary in other apects, such as shape and orientation (Figure \ref{fig:variations_examples}). To do this, we apply the decoder to the bipartite representation~$(z_i, x_T)$ using DDIM with~$\eta > 0$ for sampling. With~$\eta = 0$, the decoder becomes deterministic and will reconstruct the given image~$x$. Larger values of~$\eta$ introduce stochasticity into successive sampling steps, resulting in variations that are perceptually ``centered'' around the original image~$x$. As $\eta$ increases, these variations tell us what information was captured in the CLIP image embedding (and thus is preserved across samples), and what was lost (and thus changes across the samples).

\begin{figure*}[t]
    \centering
    \setlength{\tabcolsep}{0.6pt}
    \begin{tabular}{ccc}
        \raisebox{\height}{\includegraphics[width=0.11\textwidth]{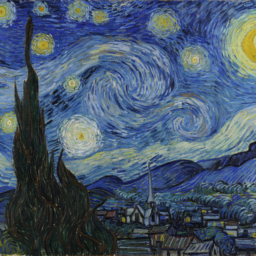}} &
        \includegraphics[width=0.77\textwidth]{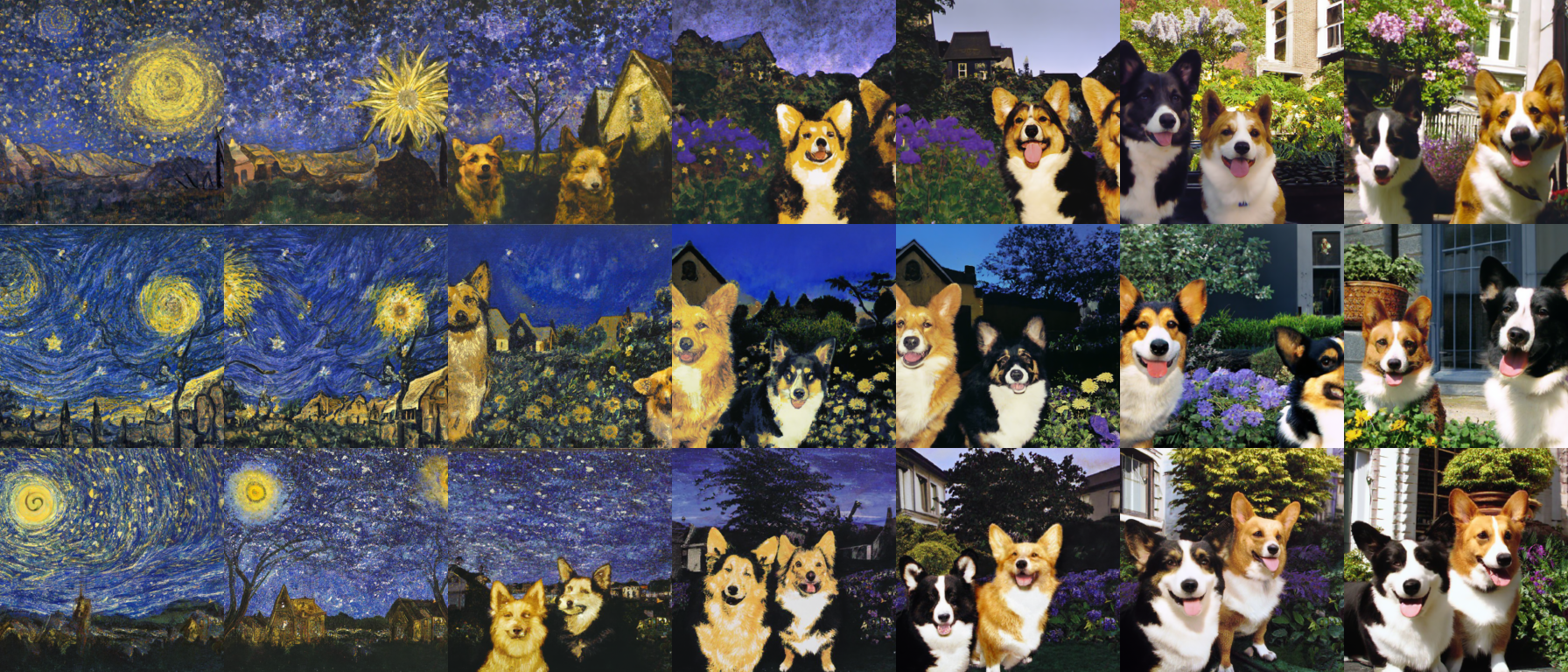} &
        \raisebox{\height}{\includegraphics[width=0.11\textwidth]{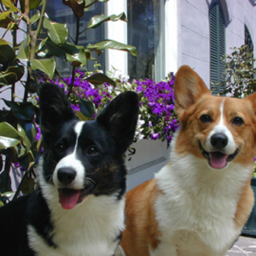}} \\
        \raisebox{\height}{\includegraphics[width=0.11\textwidth]{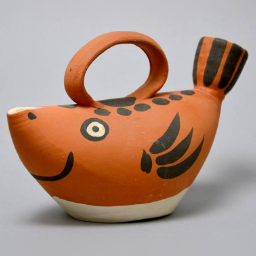}} &
        \includegraphics[width=0.77\textwidth]{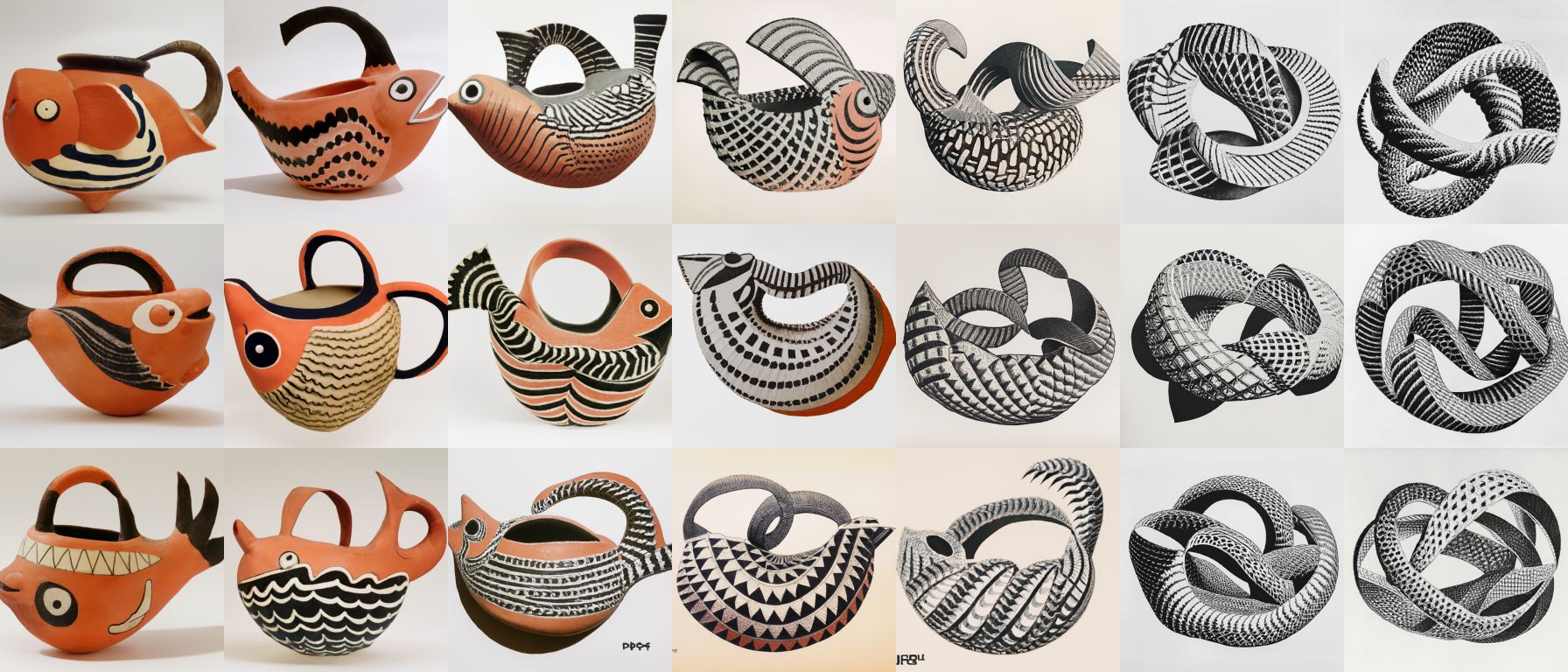} &
        \raisebox{\height}{\includegraphics[width=0.11\textwidth]{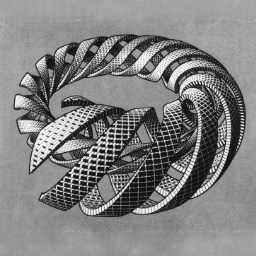}}
    \end{tabular}
    \caption{Variations between two images by interpolating their CLIP image embedding and then decoding with a diffusion model. We fix the decoder seed across each row. The intermediate variations naturally blend the content and style from both input images.}
    \label{fig:interpolations_examples}
    \vskip -0.1in 
\end{figure*}

\subsection{Interpolations}
\label{sec:interpolations}
It is also possible to blend two images~$x_1$ and~$x_2$ for variations (Figure \ref{fig:interpolations_examples}), traversing all of the concepts in CLIP's embedding space that occur between them. To do this, we rotate between their CLIP embeddings~$z_{i_1}$ and~$z_{i_2}$ using spherical interpolation, yielding intermediate CLIP representations $z_{i_\theta} = \text{slerp}(z_{i_1}, z_{i_2}, \theta)$ as~$\theta$ is varied from~0 to~1. There are two options for producing the intermediate DDIM latents along the trajectory. The first option involves interpolating between their DDIM inverted latents~$x_{T_1}$ and~$x_{T_2}$ (by setting $x_{T_\theta} = \text{slerp}(x_{T_1}, x_{T_2}, \theta)$), which yields a single trajectory whose endpoints reconstruct~$x_1$ and~$x_2$. The second option involves fixing the DDIM latent to a randomly-sampled value for all interpolates in the trajectory. This results in an infinite number of trajectories between~$x_1$ and~$x_2$, though the endpoints of these trajectories will generally no longer coincide with the original images. We use this approach in Figure~\ref{fig:interpolations_examples}.

\begin{figure*}[t]
    \centering
    \includegraphics[width=\textwidth]{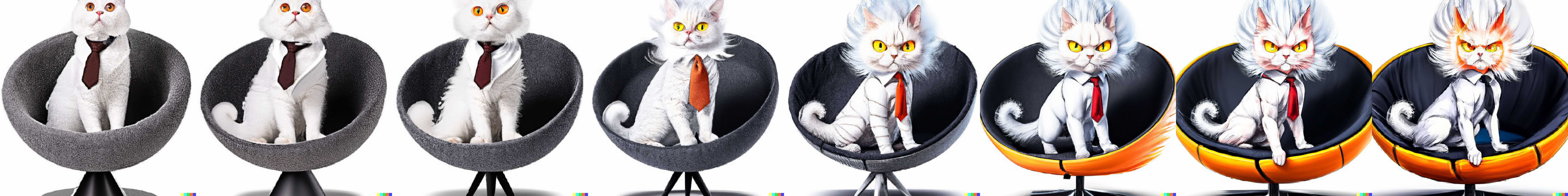}
    \vspace{0.05in}
    \scriptsize \makecell{a photo of a cat $\rightarrow$ an anime drawing of a super saiyan cat, artstation}

    \includegraphics[width=\textwidth]{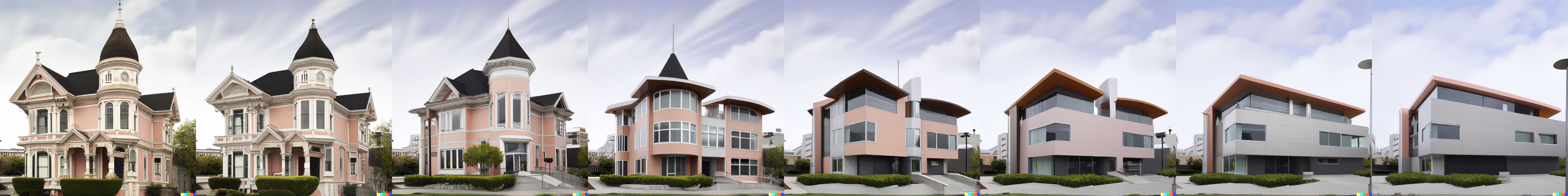}
    \vspace{0.05in}
    \scriptsize \makecell{a photo of a victorian house $\rightarrow$ a photo of a modern house}

    \includegraphics[width=\textwidth]{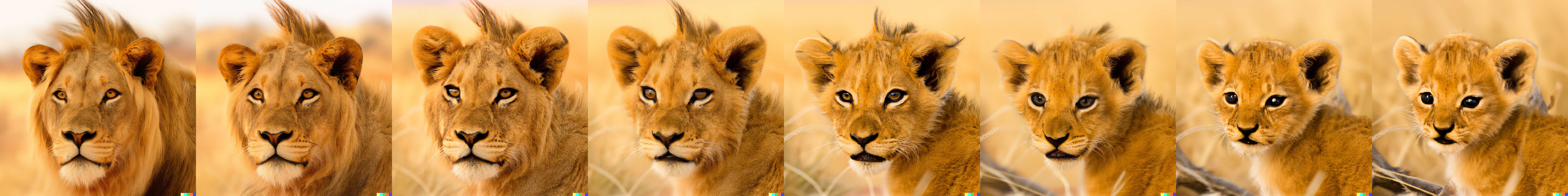}
    \vspace{0.05in}
    \scriptsize \makecell{a photo of an adult lion $\rightarrow$ a photo of lion cub}
    
    \includegraphics[width=\textwidth]{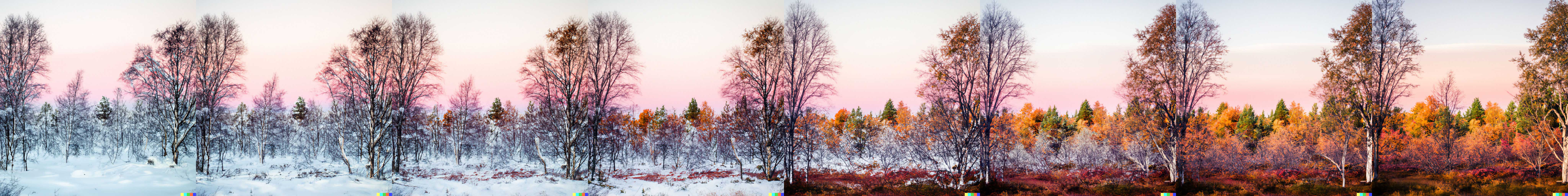}
    \vspace{0.05in}
    \scriptsize \makecell{a photo of a landscape in winter $\rightarrow$ a photo of a landscape in fall}
    
    \caption{Text diffs applied to images by interpolating between their CLIP image embeddings and a normalised difference of the CLIP text embeddings produced from the two descriptions. We also perform DDIM inversion to perfectly reconstruct the input image in the first column, and fix the decoder DDIM noise across each row. }
    \label{fig:text_diffs}
    \vskip -0.2in 
\end{figure*}

\subsection{Text Diffs}
\label{sec:text_diffs}
A key advantage of using CLIP compared to other models for image representations is that it embeds images and text to the same latent space, thus allowing us to apply language-guided image manipulations (i.e., text diffs), which we show in Figure~\ref{fig:text_diffs}. To modify the image to reflect a new text description~$y$, we first obtain its CLIP text embedding~$z_t$, as well as the CLIP text embedding $z_{t_0}$ of a caption describing the current image\footnote{Instead of a description of the current image, we also experimented with using a dummy caption like ``a photo'' for the baseline, or removing it altogether. These also worked well.}. We then compute a \textit{text diff} vector $z_d = \text{norm}(z_t - z_{t_0})$ from these by taking their difference and normalizing. Now, we can rotate between the image CLIP embedding~$z_i$ and the text diff vector~$z_d$ using spherical interpolation, yielding intermediate CLIP representations $z_\theta = \text{slerp}(z_i, z_d, \theta)$, where~$\theta$ is increased linearly from~0 to a maximum value that is typically in~$[0.25, 0.50]$. We produce the final outputs by decoding the interpolates $z_\theta$, fixing the base DDIM noise to~$x_T$ throughout the entire trajectory.

\begin{figure*}[t]
    \centering
    \setlength{\tabcolsep}{4.0pt}
    \begin{tabular}{ccc}
        \includegraphics[width=0.3\textwidth]{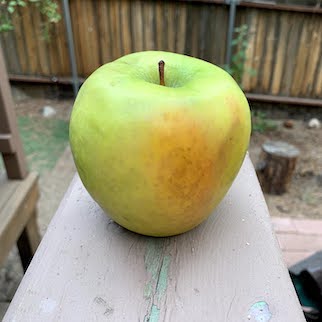} & 
        \includegraphics[width=0.3\textwidth]{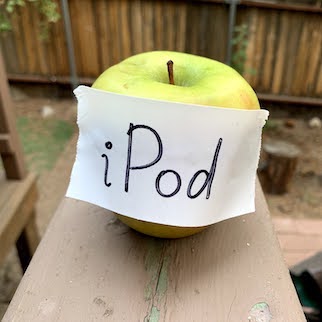} &
        \includegraphics[width=0.3\textwidth]{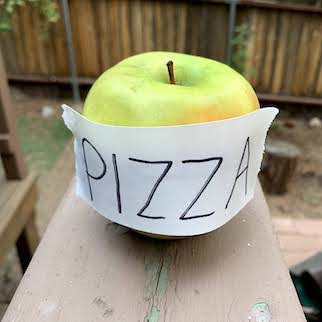} \\
        \rule{0pt}{0.0pt} \\
        \includegraphics[width=0.3\textwidth]{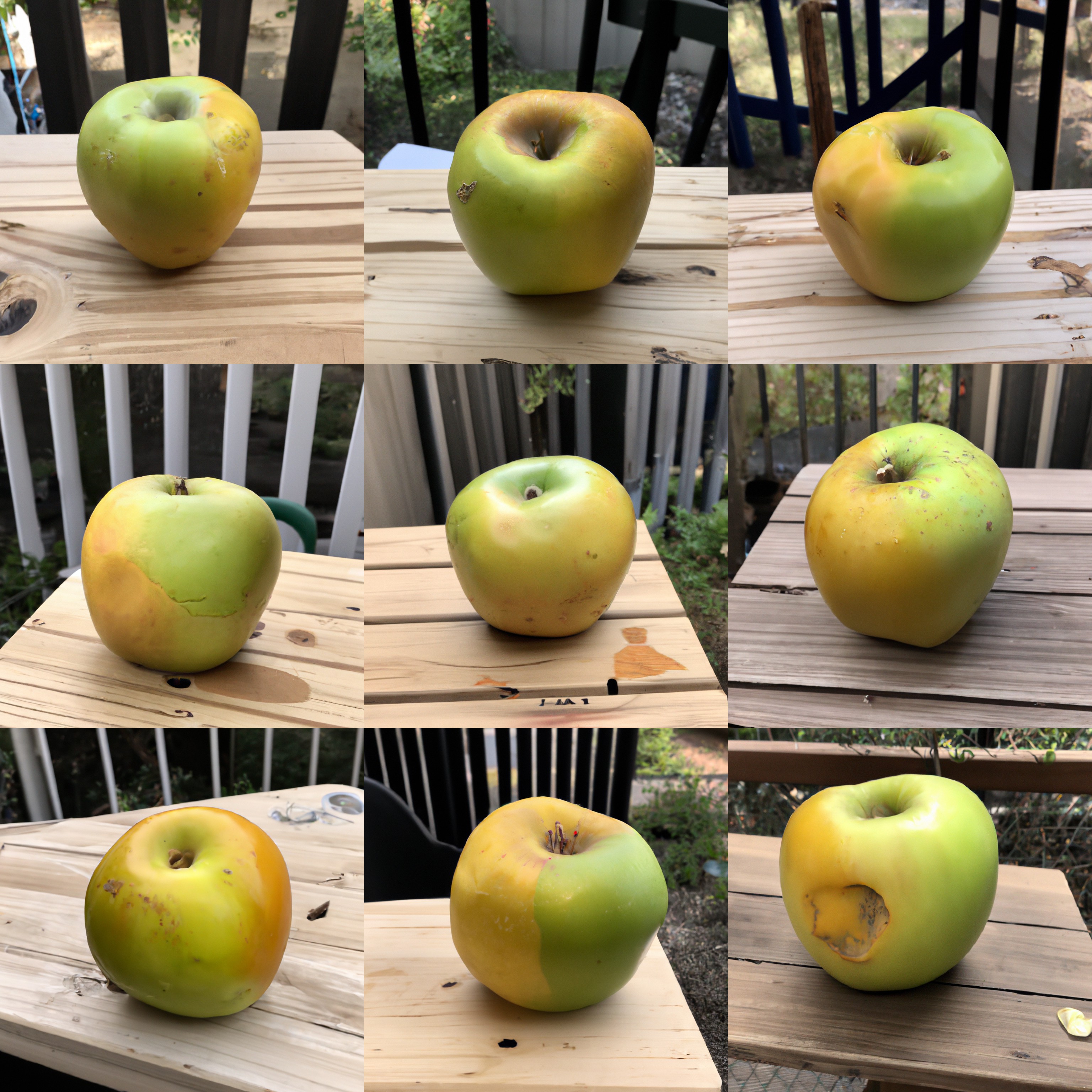} &
        \includegraphics[width=0.3\textwidth]{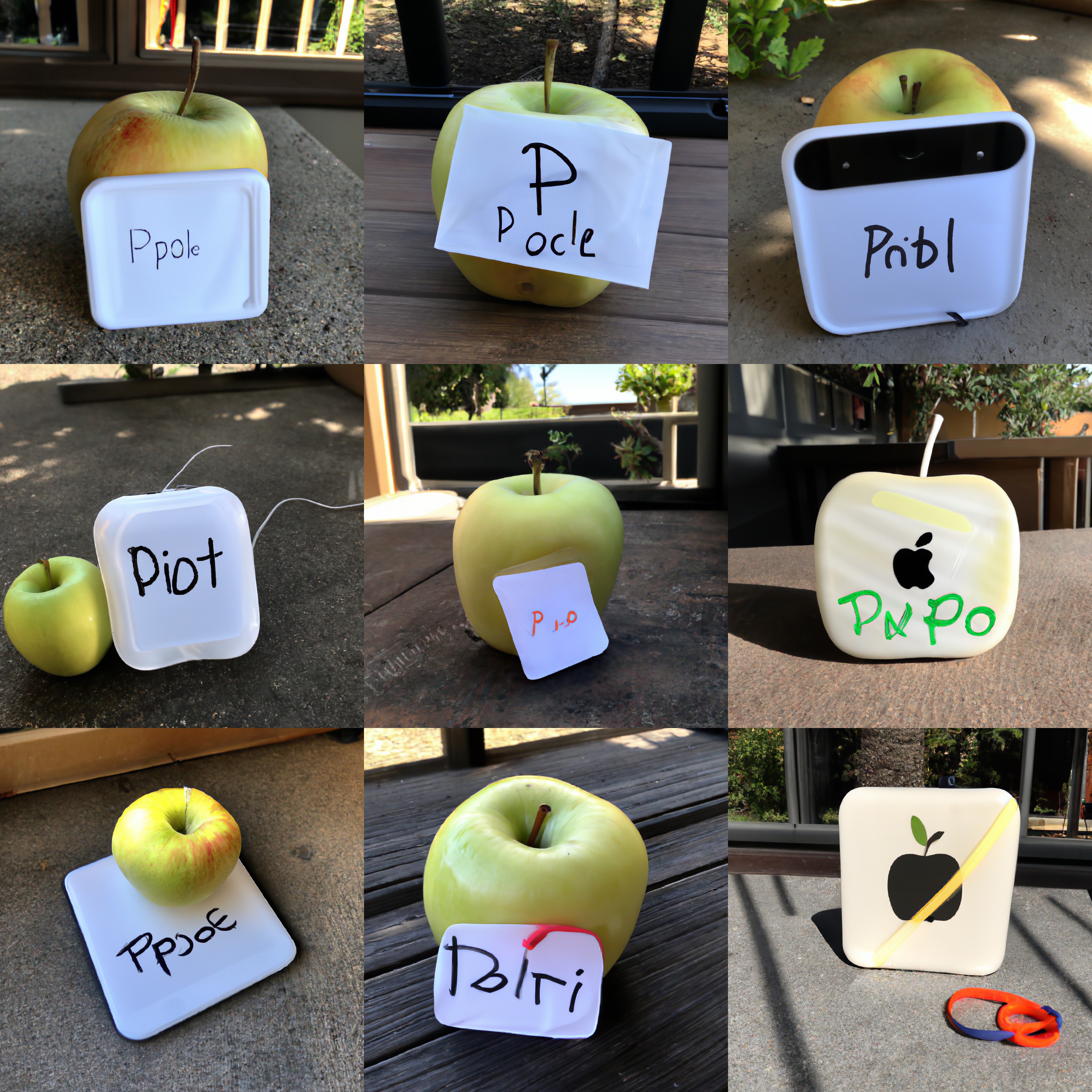} &
        \includegraphics[width=0.3\textwidth]{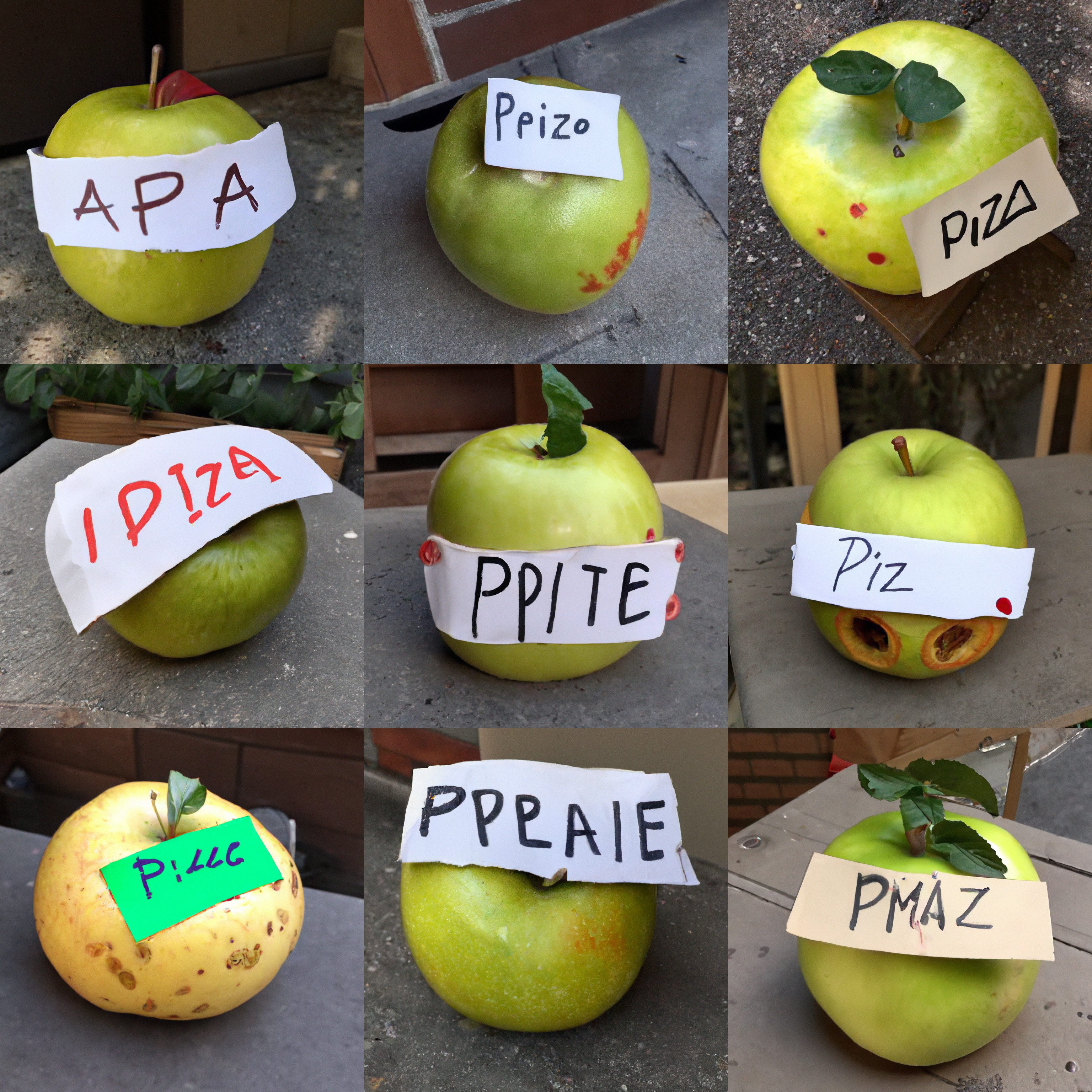} \\
        \makecell{Granny Smith: 100\% \\ iPod: 0\% \\ Pizza: 0\%} & \makecell{Granny Smith: 0.02\% \\ iPod: 99.98\% \\ Pizza: 0\%} & \makecell{Granny Smith: 94.33\% \\ iPod: 0\% \\ Pizza: 5.66\%} \\
        \rule{0pt}{0.5pt}
    \end{tabular}
    \vskip -0.1in
    \caption{\label{fig:variation_adversarial} Variations of images featuring typographic attacks \shortcite{multimodalneurons} paired with the CLIP model's predicted probabilities across three labels. Surprisingly, the decoder still recovers Granny Smith apples even when the predicted probability for this label is near 0\%. We also find that our CLIP model is slightly less susceptible to the ``pizza'' attack than the models investigated in \shortcite{multimodalneurons}.}
    \vskip -0.1in
\end{figure*}
\section{Probing the CLIP Latent Space}
\label{sec:latentspace}

Our decoder model provides a unique opportunity to explore CLIP latent space by allowing us to directly visualize what the CLIP image encoder is seeing. 
As an example use case, we can revisit cases where CLIP makes incorrect predictions, such as typographic attacks \shortcite{multimodalneurons}. In these adversarial images, a piece of text is overlayed on top of an object, which causes CLIP to predict the object described by the text rather than the object depicted in the image. This piece of text essentially hides the original object in terms of output probabilities. In Figure~\ref{fig:variation_adversarial}, we show an example of this attack from~\shortcite{multimodalneurons}, wherein an apple can be misclassified as an iPod. Surprisingly, we find that our decoder still generates pictures of apples with high probability even though the predicted probability of ``Granny Smith'' is near zero. Even more notable, the model never produces pictures of iPods, despite the very high relative predicted probability of this caption.

\begin{figure*}[t]
    \centering
    \setlength{\tabcolsep}{0.6pt}
    \renewcommand{\arraystretch}{0}
    \begin{tabular}{cc}
        \includegraphics[width=0.88\textwidth]{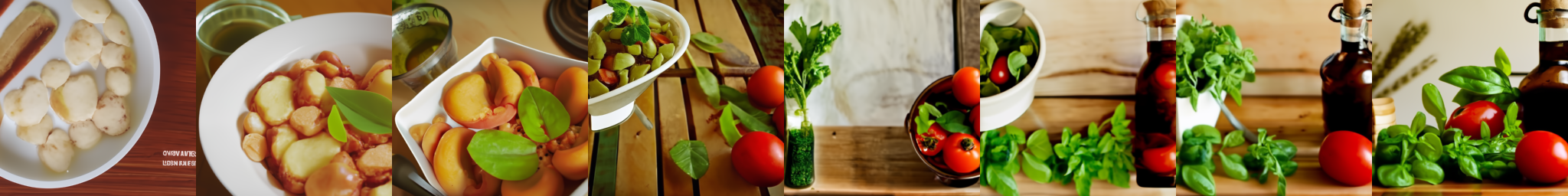} & 
        \includegraphics[width=0.11\textwidth]{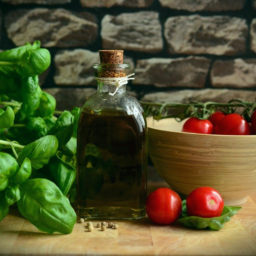}
         \\
        \includegraphics[width=0.88\textwidth]{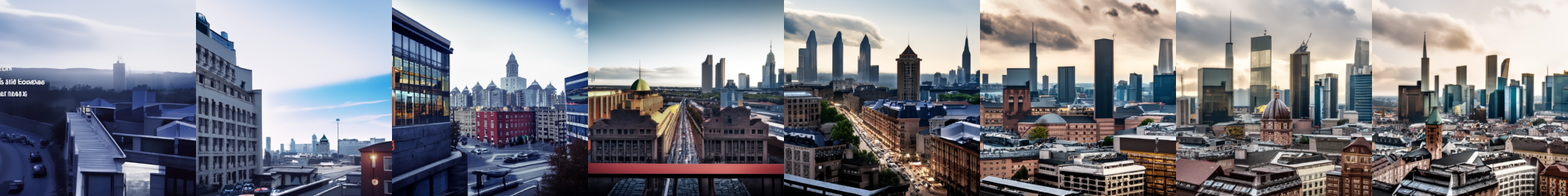} &
        \includegraphics[width=0.11\textwidth]{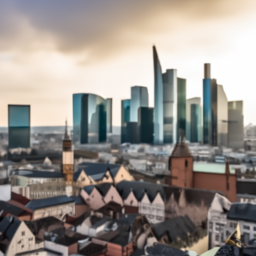}
         \\
         \includegraphics[width=0.88\textwidth]{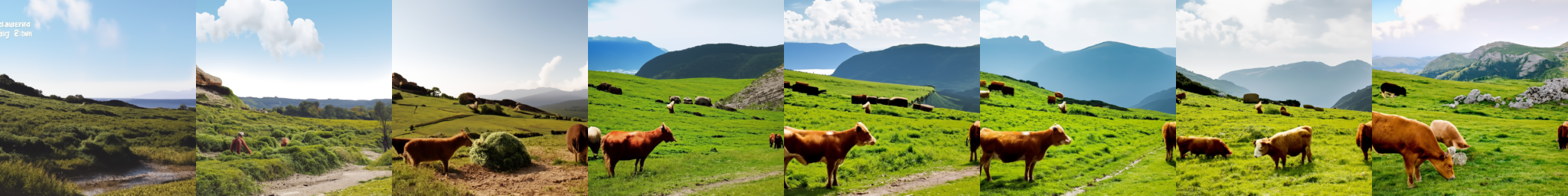} &
        \includegraphics[width=0.11\textwidth]{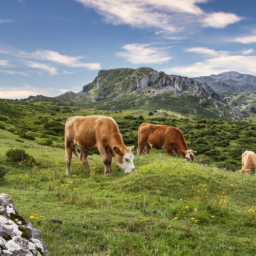}
         \\
    \end{tabular}
    \caption{Visualization of reconstructions of CLIP latents from progressively more PCA dimensions (20, 30, 40, 80, 120, 160, 200, 320 dimensions), with the original source image on the far right. The lower dimensions preserve coarse-grained semantic information, whereas the higher dimensions encode finer-grained details about the exact form of the objects in the scene.}
    \label{fig:pca_reconstructions}
\end{figure*}

PCA reconstructions offer another tool for probing the structure of the CLIP latent space. In Figure~\ref{fig:pca_reconstructions}, we take the CLIP image embeddings of a handful of source images and reconstruct them with progressively more PCA dimensions, and then visualize the reconstructed image embeddings using our decoder with DDIM on a fixed seed. This allows us to see what semantic information the different dimensions encode. We observe that the early PCA dimensions preserve coarse-grained semantic information such as what types of objects are in the scene, whereas the later PCA dimensions encode finer-grained detail such as the shapes and exact form of the objects. For example, in the first scene, the earlier dimensions seem to encode that there is food and perhaps a container present, whereas the later dimensions encode tomatoes and a bottle specifically. Figure~\ref{fig:pca_reconstructions} also serves as a visualization of what the AR prior is modeling, since the AR prior is trained to explicitly predict these principal components in this order.

\begin{figure*}[t]
    \centering
    \setlength{\tabcolsep}{2.0pt}
    \begin{tabular}{cccccc}
        \rotatebox{90}{\scriptsize\phantom{AAAA} Caption} &
        \includegraphics[width=0.18\textwidth]{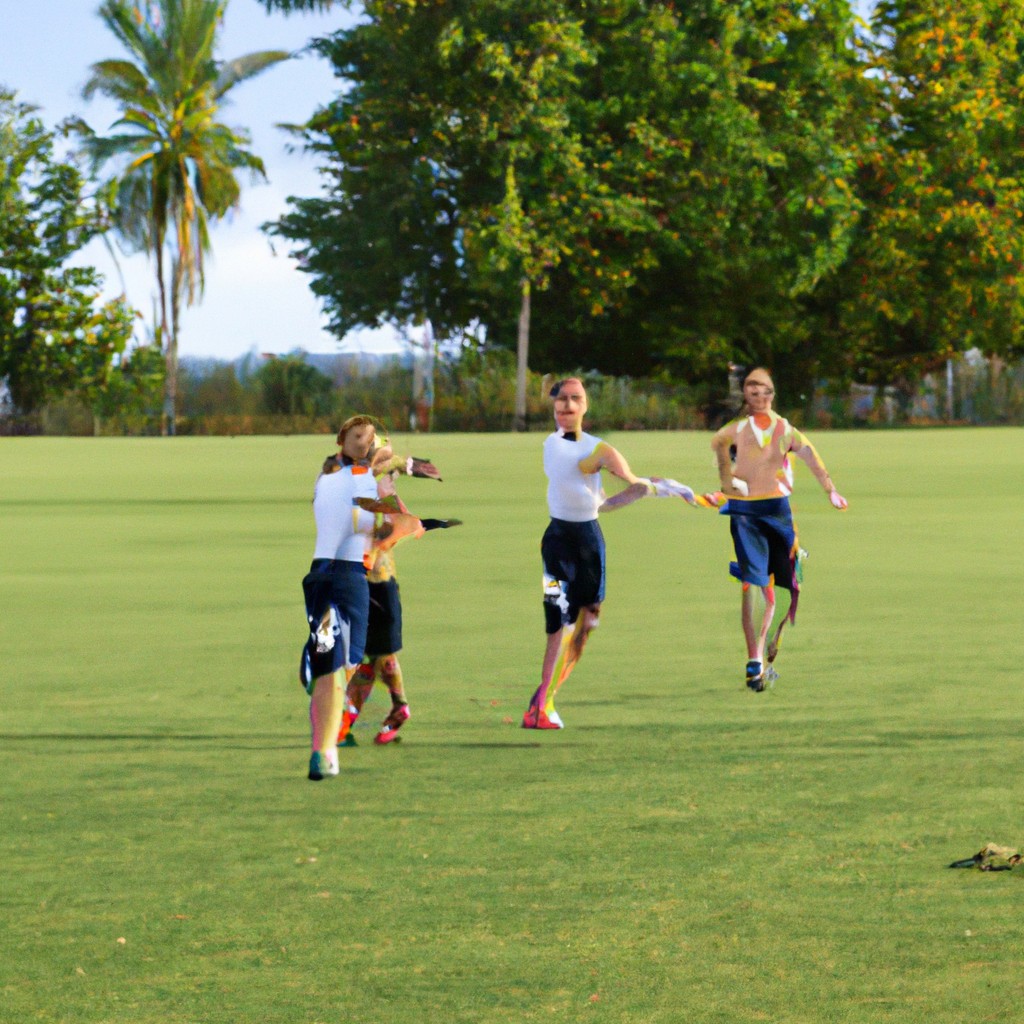} &
        \includegraphics[width=0.18\textwidth]{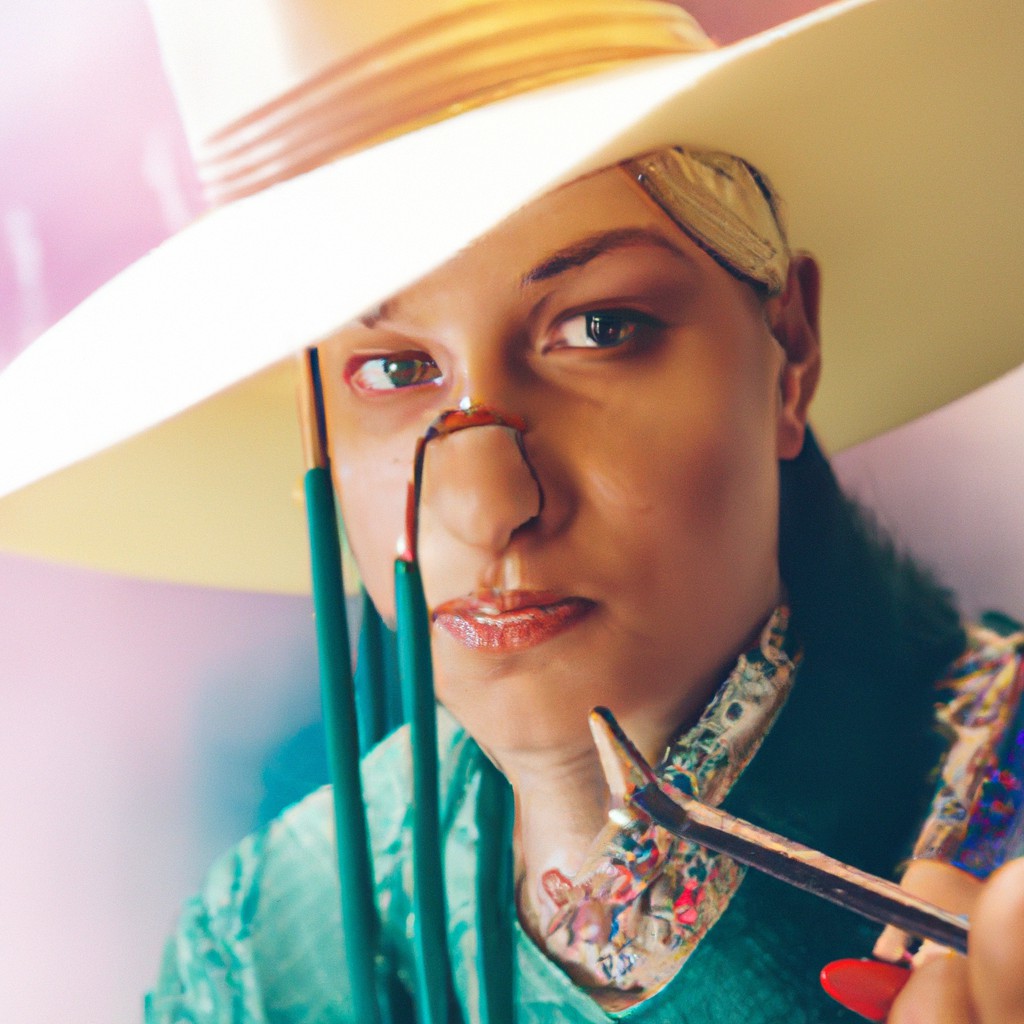} &
        \includegraphics[width=0.18\textwidth]{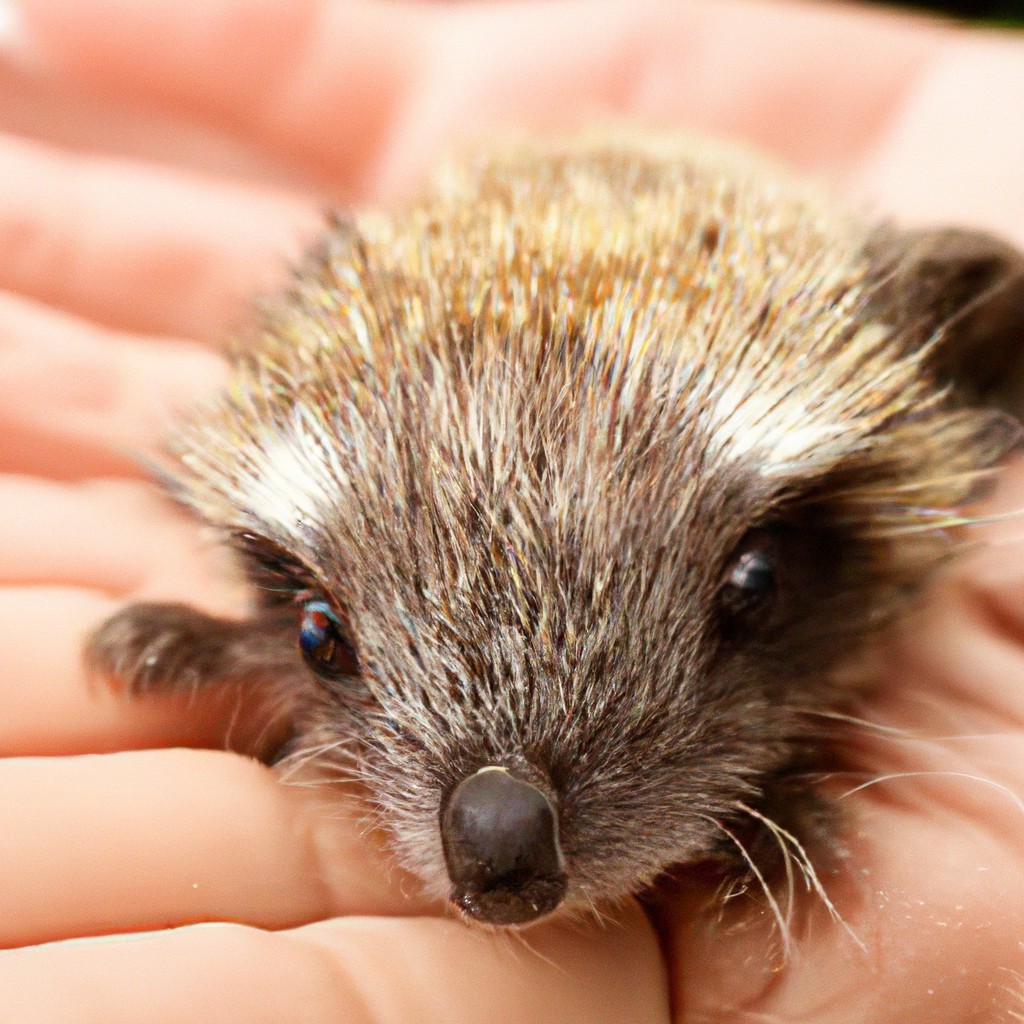} &
        \includegraphics[width=0.18\textwidth]{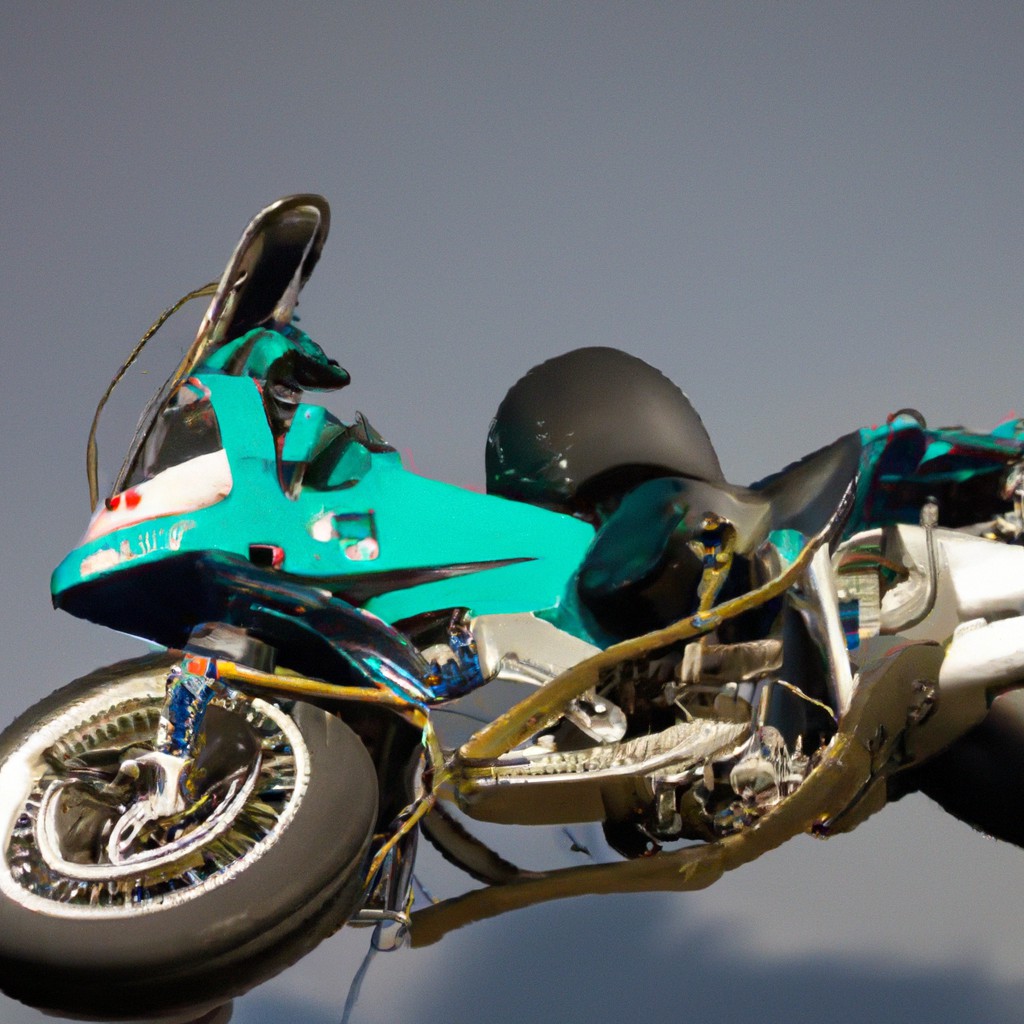} &
        \includegraphics[width=0.18\textwidth]{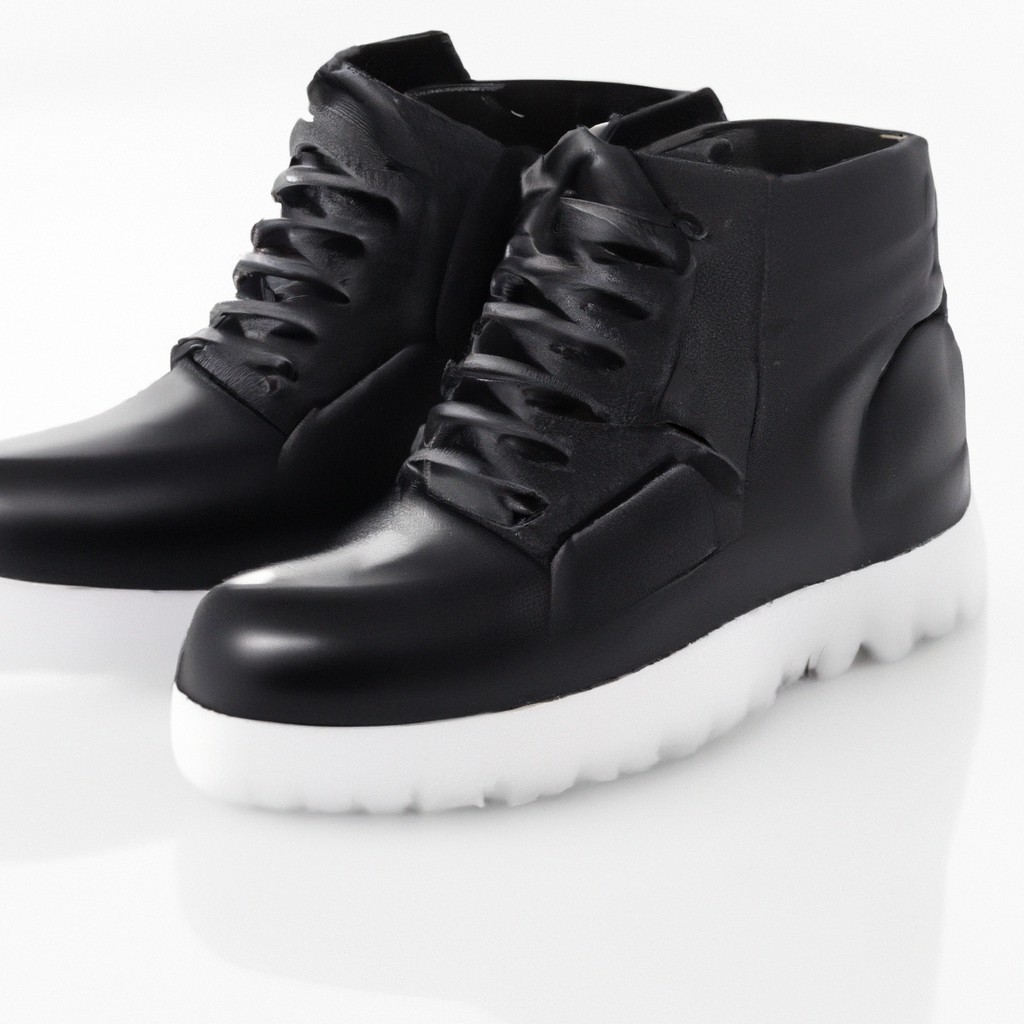} \\

        \rotatebox{90}{\scriptsize\phantom{AA.} Text embedding} &
        \includegraphics[width=0.18\textwidth]{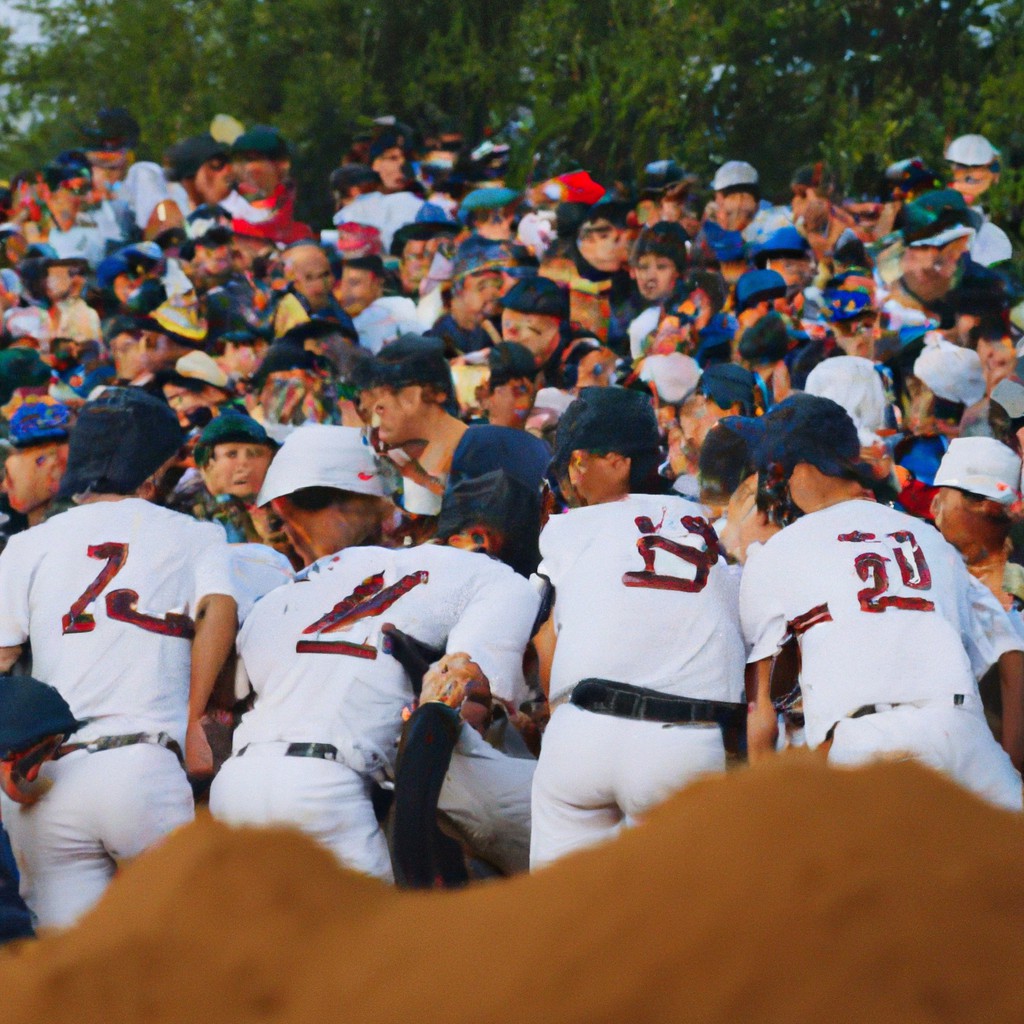} &
        \includegraphics[width=0.18\textwidth]{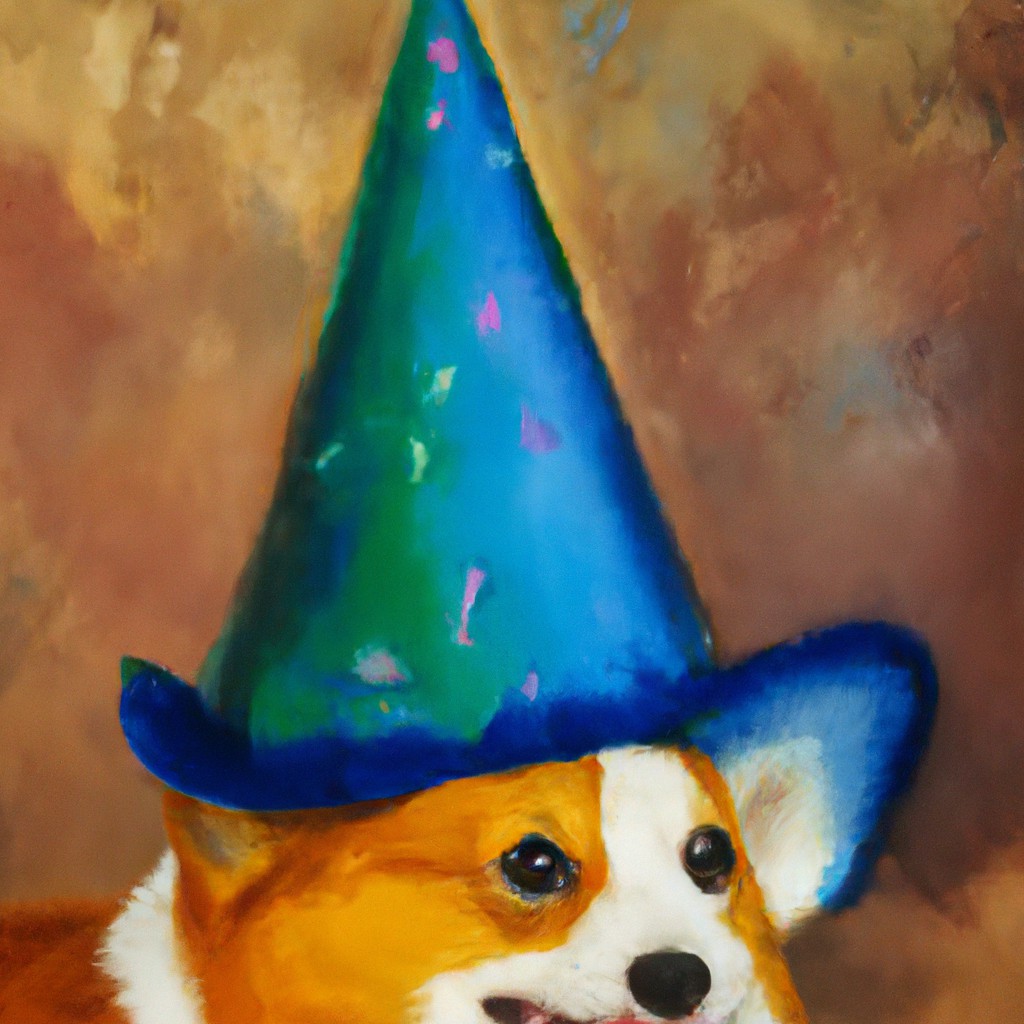} &
        \includegraphics[width=0.18\textwidth]{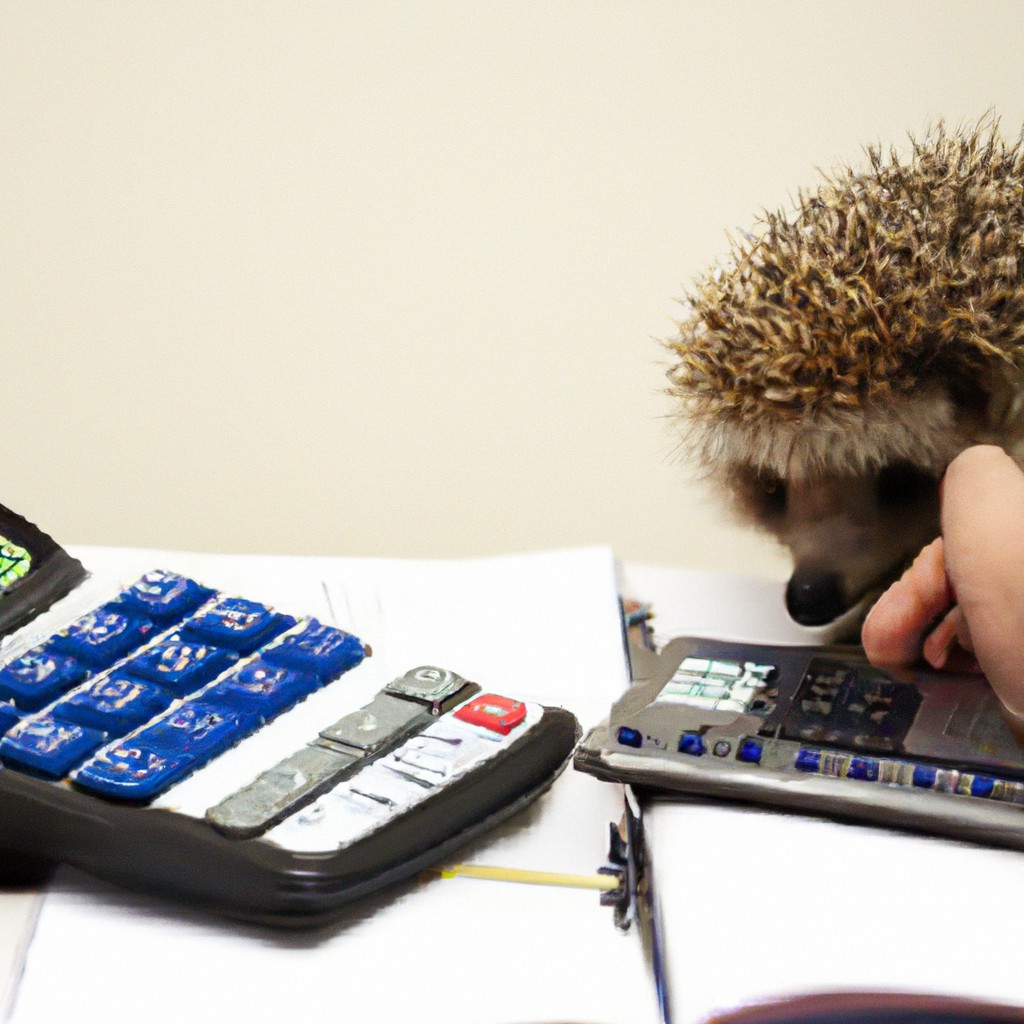} &
        \includegraphics[width=0.18\textwidth]{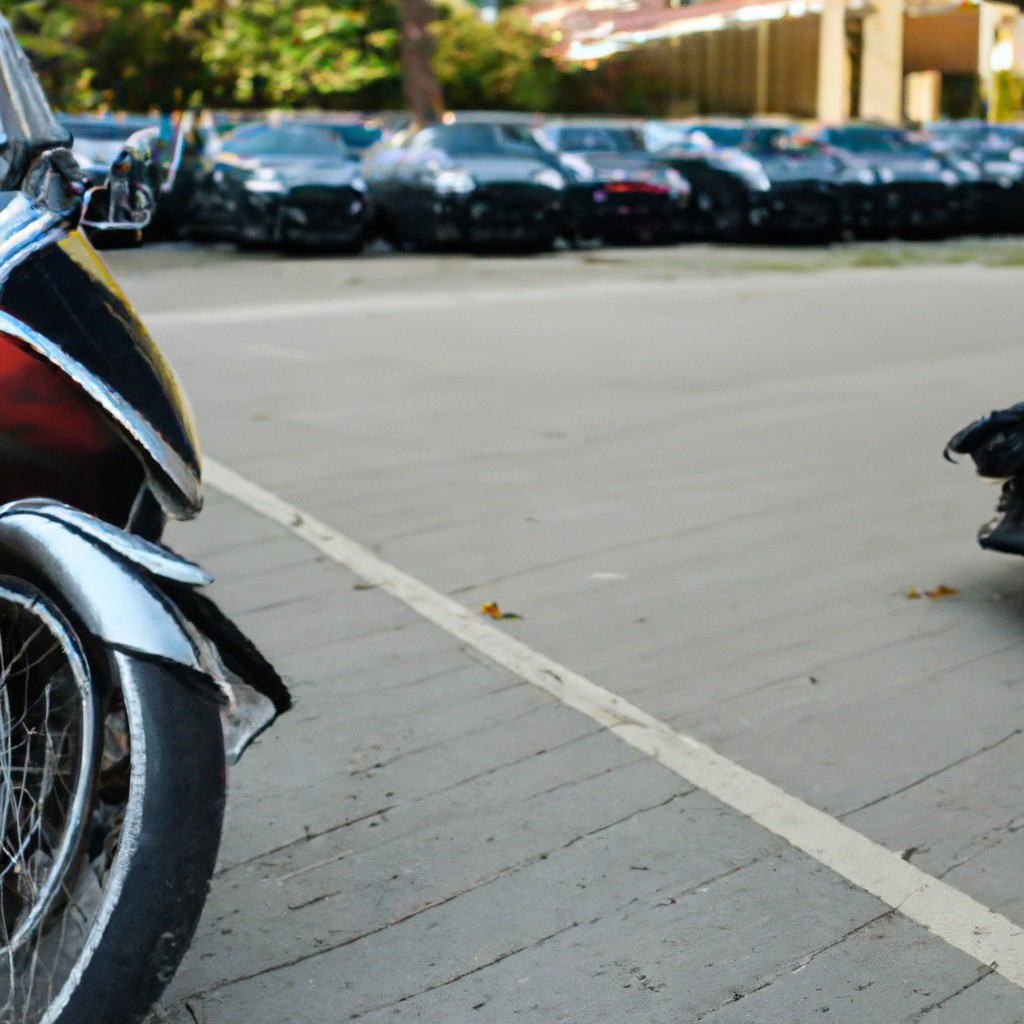} &
        \includegraphics[width=0.18\textwidth]{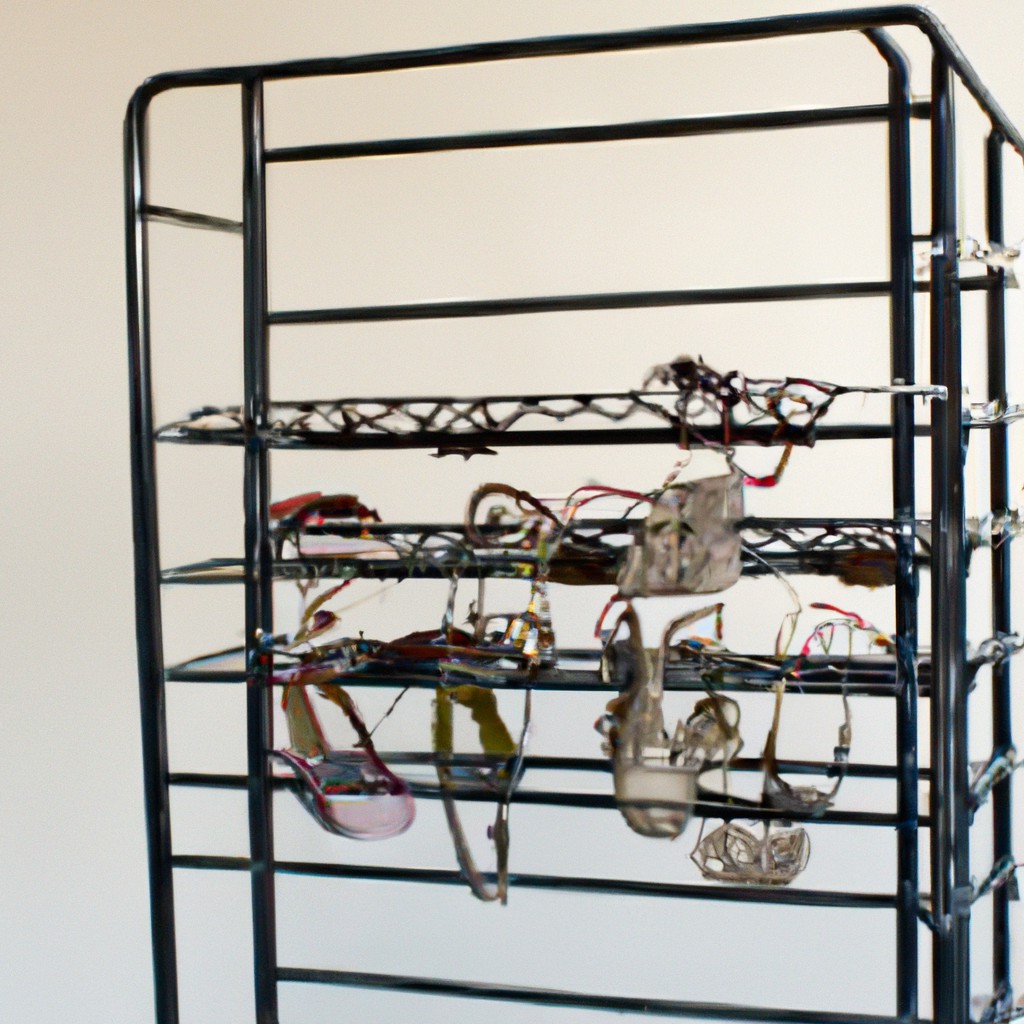} \\
        
        \rotatebox{90}{\scriptsize\phantom{AA} Image embedding} &
        \includegraphics[width=0.18\textwidth]{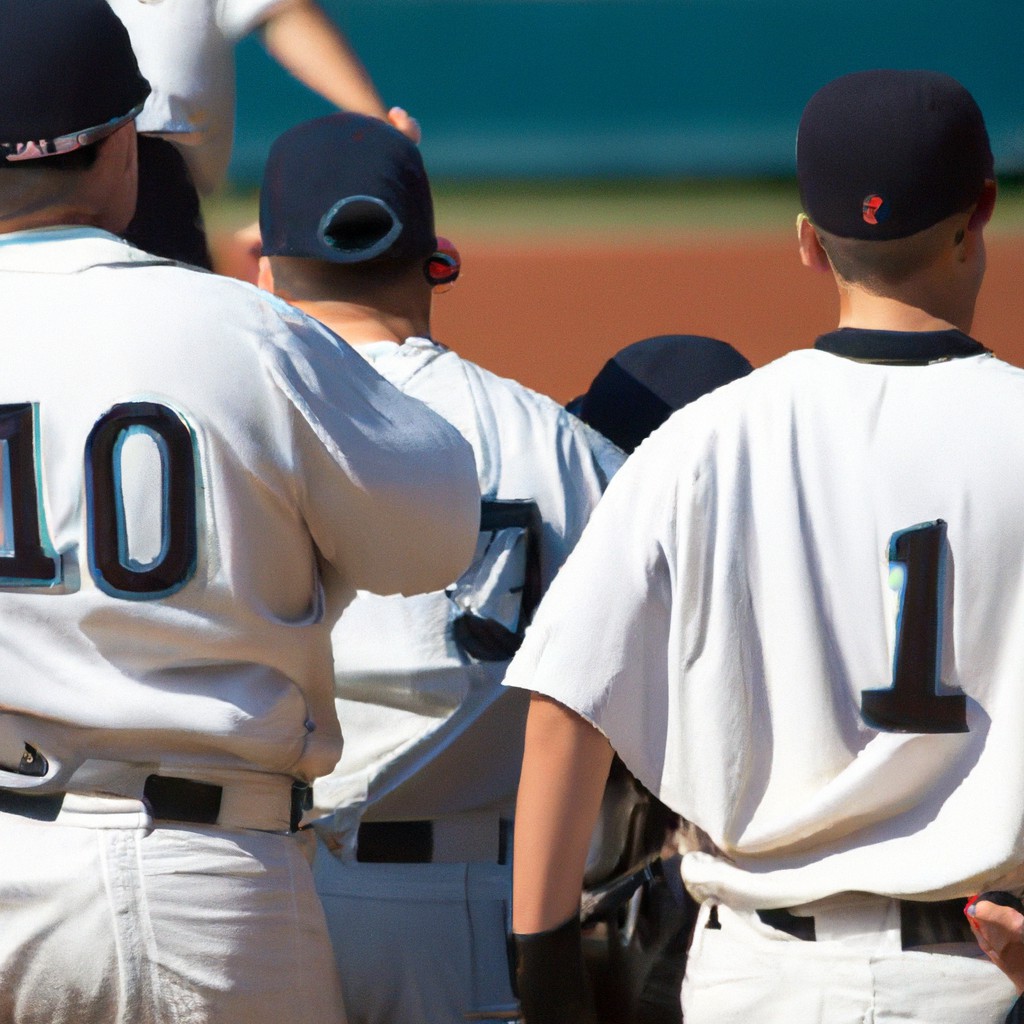} &
        \includegraphics[width=0.18\textwidth]{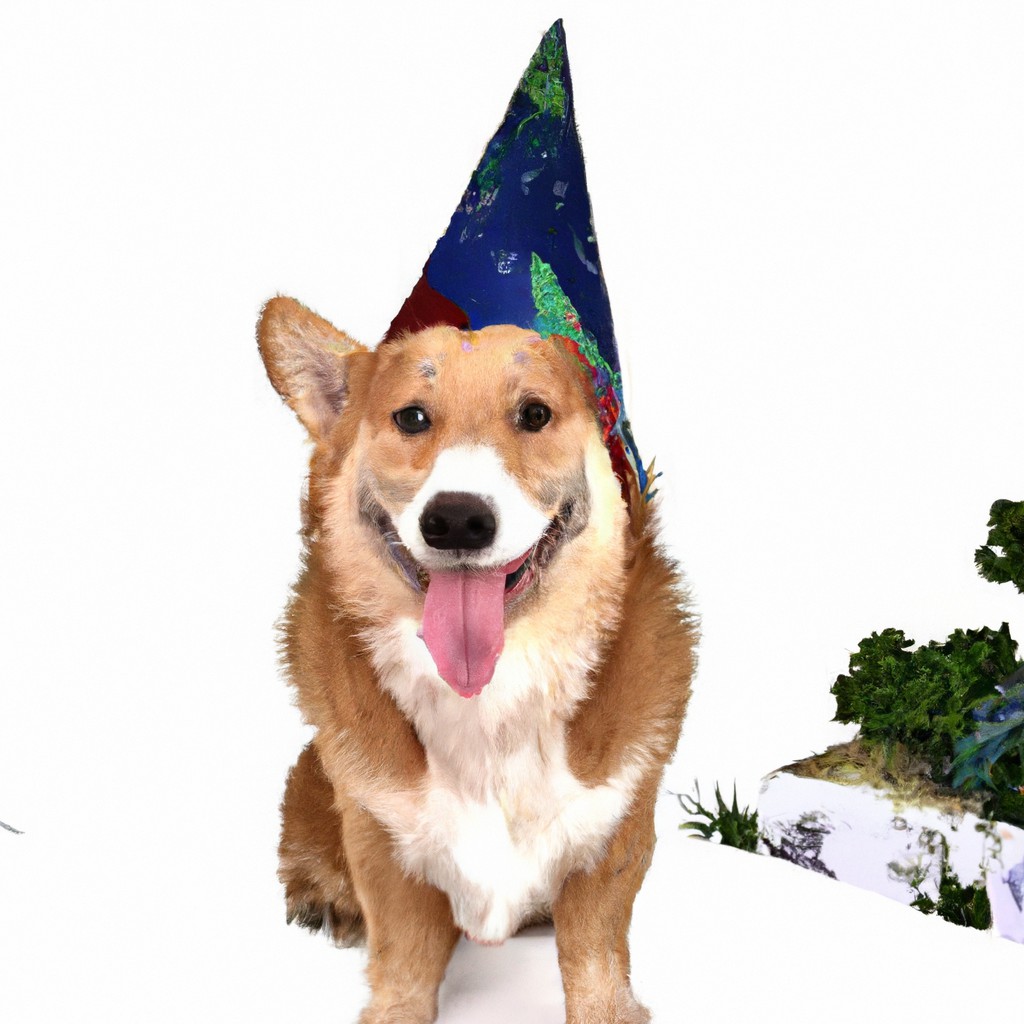} &
        \includegraphics[width=0.18\textwidth]{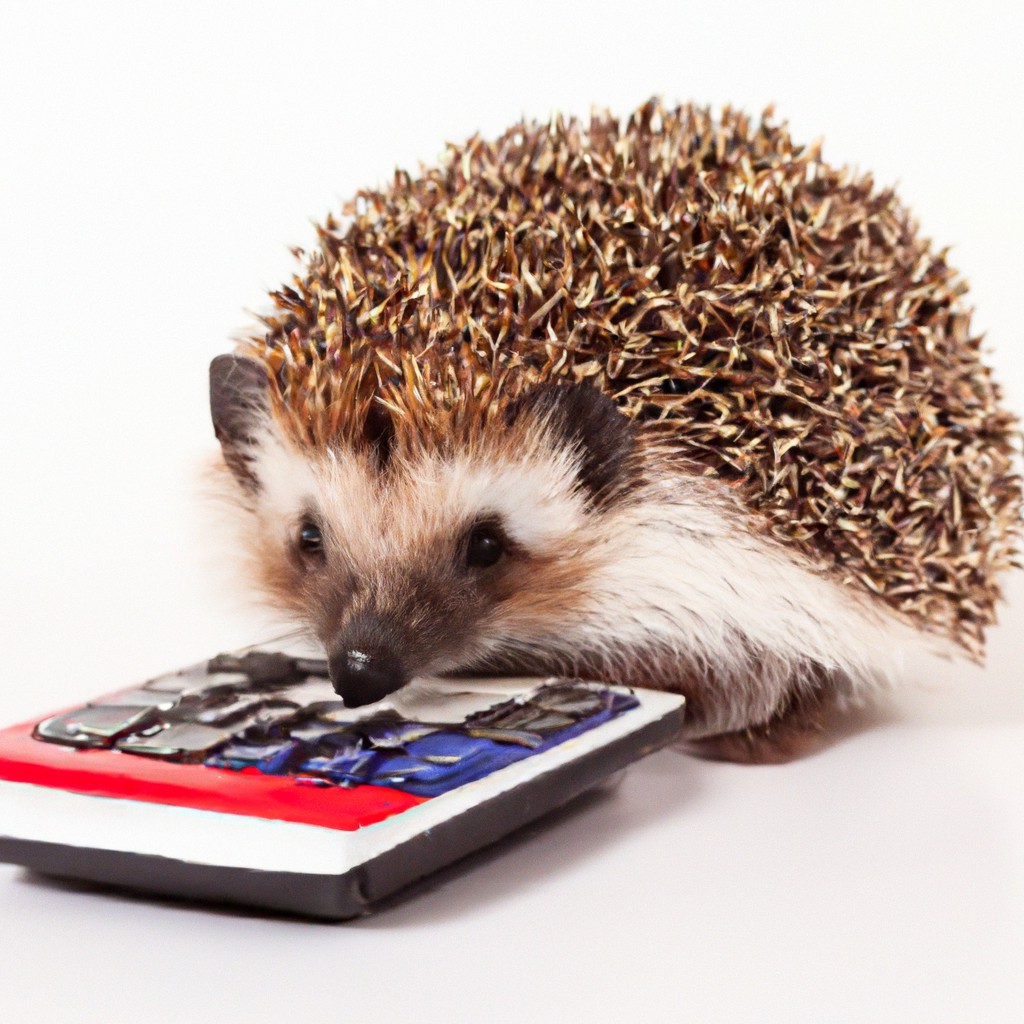} &
        \includegraphics[width=0.18\textwidth]{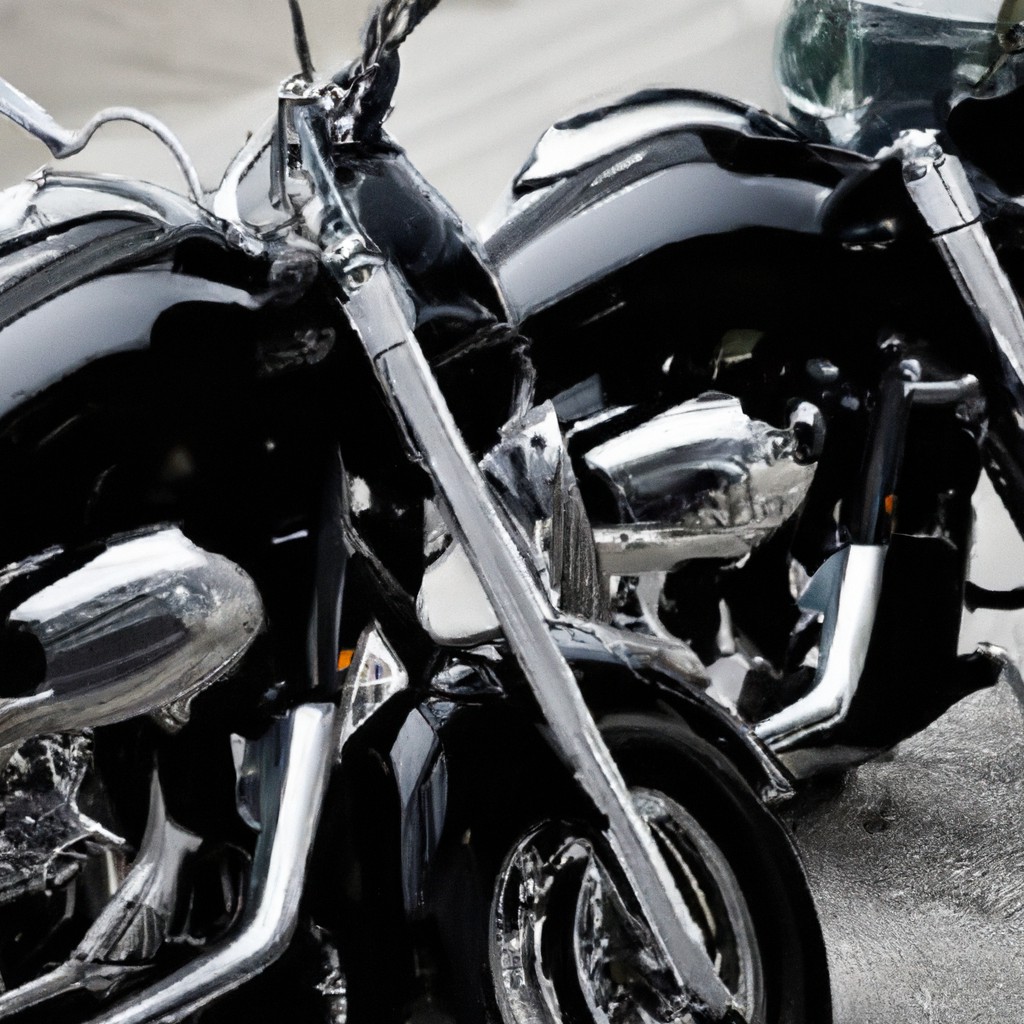} &
        \includegraphics[width=0.18\textwidth]{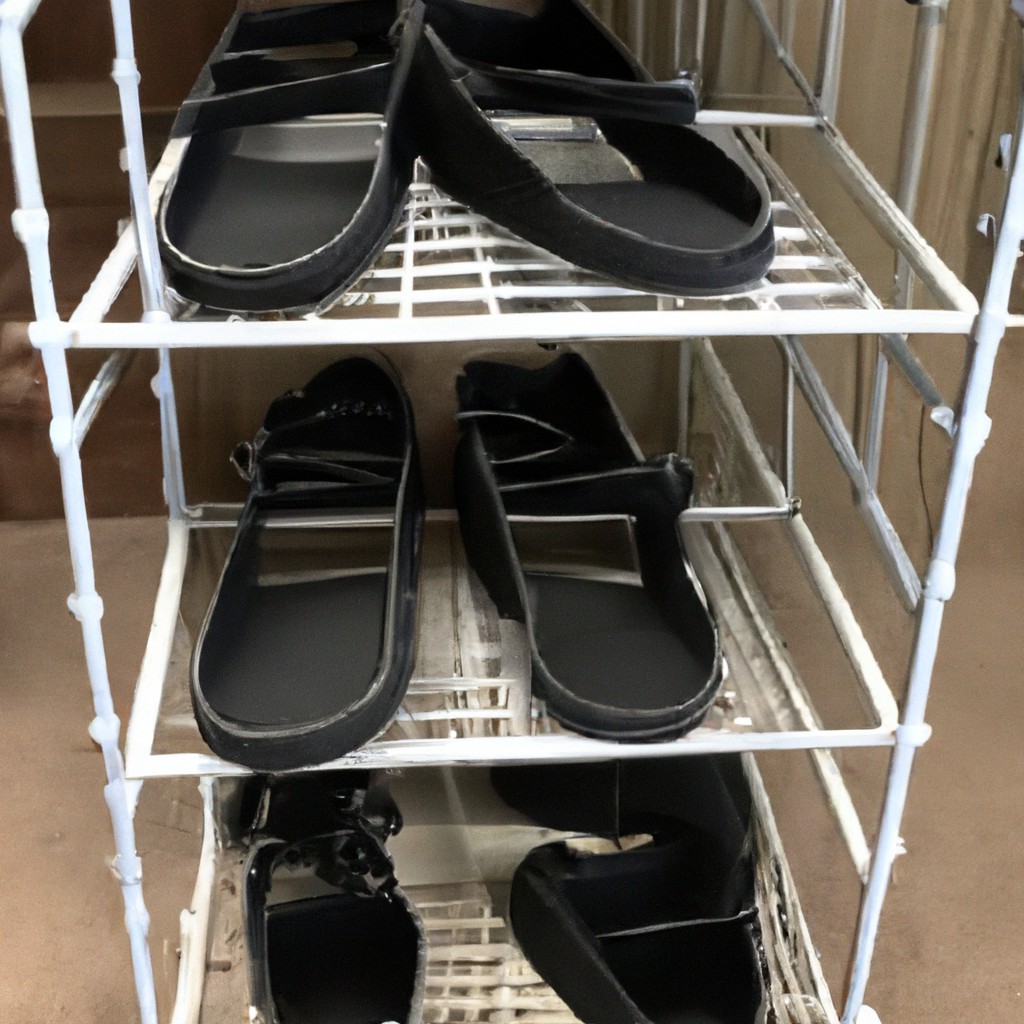} \\

        & \scriptsize \makecell{``A group of baseball \\ players is crowded at \\ the mound.''}
        & \scriptsize \makecell{``an oil painting of a \\ corgi wearing a \\ party hat''}
        & \scriptsize \makecell{``a hedgehog using a \\ calculator''}
        & \scriptsize \makecell{``A motorcycle parked in a \\ parking space next to \\ another motorcycle.''}
        & \scriptsize \makecell{``This wire metal rack \\ holds several pairs of \\ shoes and sandals''}
    \end{tabular}
    \caption{Samples using different conditioning signals for the \emph{same} decoder. In the first row, we pass the text caption to the decoder, and pass a zero vector for the CLIP embedding. In the second row, we pass both the text caption and the CLIP text embedding of the caption. In the third row, we pass the text and a CLIP image embedding generated by an autoregressive prior for the given caption. Note that this decoder is only trained to do the text-to-image generation task (without the CLIP image representation) 5\%~of the time.}
    \label{fig:conditioning_info}
    \vskip -0.1in
\end{figure*}

\section{Text-to-Image Generation}

\subsection{Importance of the Prior}

Although we train a prior to generate CLIP image embeddings from captions, the prior is not strictly necessary for caption-to-image generation. For instance, our decoder can condition on both CLIP image embeddings and captions, but the CLIP image embedding is dropped 5\% of the time during training in order to enable classifier-free guidance. Therefore, at sampling time, we can condition on only the caption, although this underperforms a model trained fully in this way (this model is GLIDE, and we do a thorough comparison with GLIDE in Sections~\ref{sec:human_evals} and~\ref{sec:tradeoff}). Another possibility is to feed the decoder the CLIP text embedding as if it were an image embedding, as previously observed~\shortcite{lafite,clipgen}. The first two rows of Figure \ref{fig:conditioning_info} depicts samples obtained in these two ways; the third row depicts samples obtained with a prior. Conditioning the decoder on just the caption is clearly worst, but conditioning on text embeddings zero-shot does produce reasonable results. Building on this observation, another approach would be to train the decoder to condition on CLIP text embeddings~\shortcite{vdiffusion} instead of CLIP image embeddings (although we would lose the capabilities mentioned in Section~\ref{sec:latentspace}).

To quantify the effectiveness of these alternate approaches, we train two models: a small decoder conditioned on CLIP text embeddings, and a small \modelname{} stack (diffusion prior and decoder). We then compare samples from the text-embedding decoder, samples from the \modelname{} stack, and samples obtained from feeding text embeddings to the \modelname{} decoder zero-shot, sweeping across guidance scales for all models. We find that these approaches respectively score FIDs of 9.16, 7.99, and 16.55 on a test set, suggesting the \modelname{} approach is best. We also run human evaluations comparing the first two settings, sweeping over sampling hyperparameters for each using our human evaluation proxy model (Appendix \ref{app:linear_probes}). We find that humans prefer the full \modelname{} stack 57.0\% $\pm$ 3.1\% of the time for photorealism and 53.1\% $\pm$ 3.1\% of the time for caption similarity.

Given the importance of the prior, it is worth evaluating different approaches for training it. We compare both the AR and diffusion priors throughout our experiments. In all cases (Sections \ref{sec:human_evals}, \ref{sec:fid}, and \ref{sec:aesthetic}), we find that the diffusion prior outperforms the AR prior for comparable model size and reduced training compute.

\subsection{Human Evaluations}
\label{sec:human_evals}

\begin{figure*}[t]
    \centering
    \setlength{\tabcolsep}{4.0pt}
    \begin{tabular}{ccc}
        \rotatebox{90}{\phantom{AA}1.0} & \includegraphics[width=0.46\textwidth]{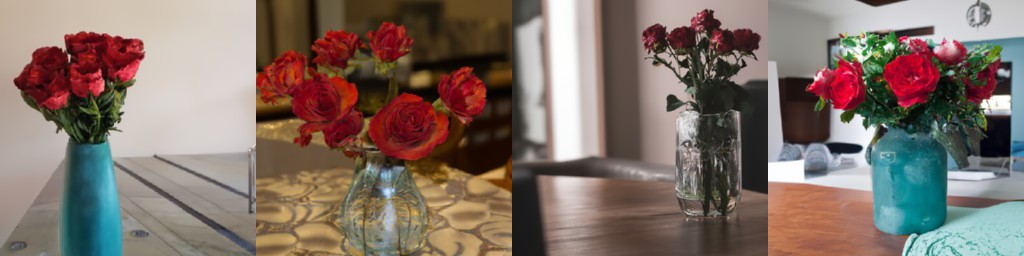} & 
        \includegraphics[width=0.46\textwidth]{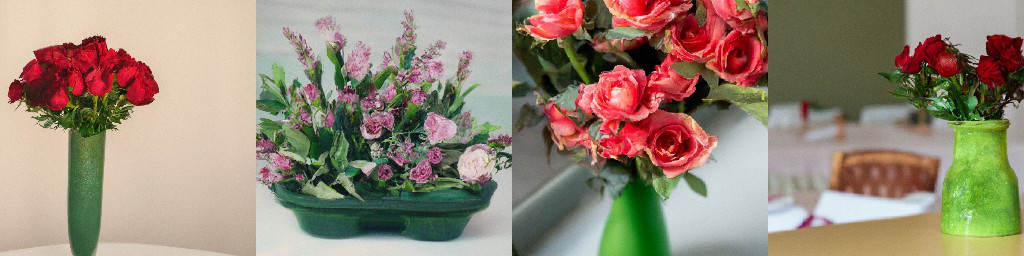} \\
        \rotatebox{90}{\phantom{AA}2.0} & \includegraphics[width=0.46\textwidth]{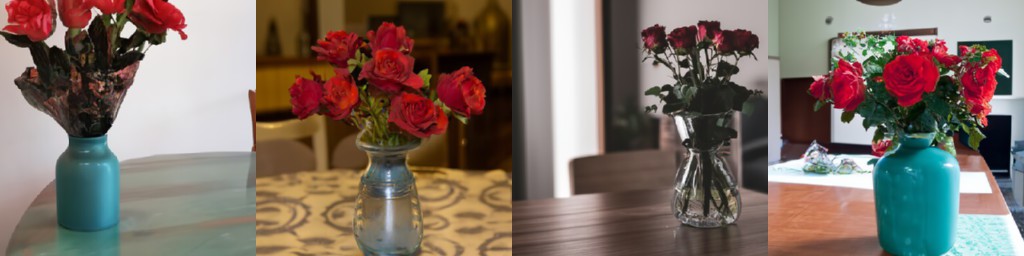} & 
        \includegraphics[width=0.46\textwidth]{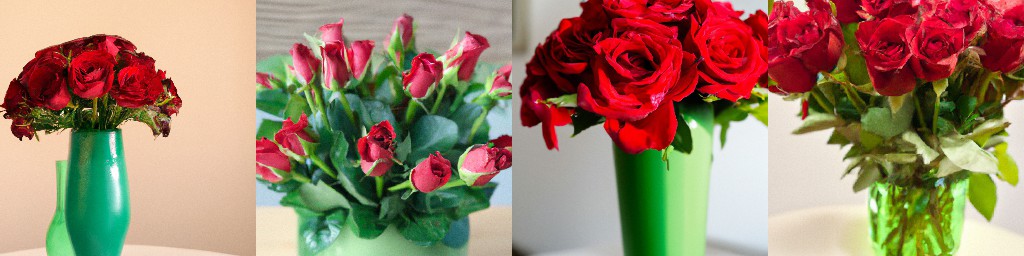} \\
        \rotatebox{90}{\phantom{AA}3.0} & \includegraphics[width=0.46\textwidth]{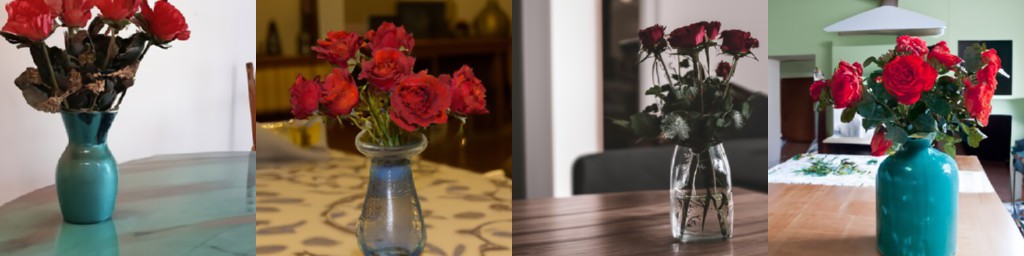} & 
        \includegraphics[width=0.46\textwidth]{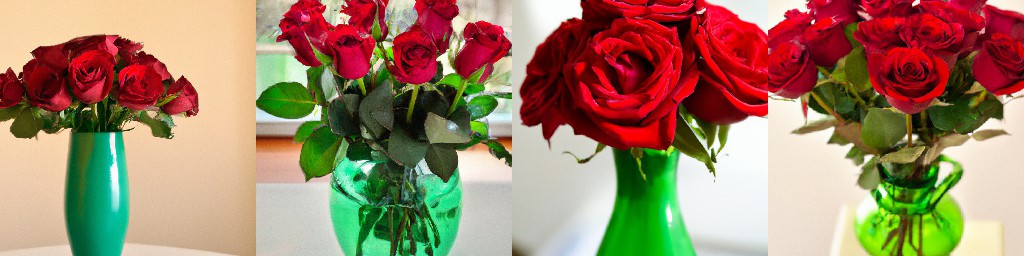} \\
        \rotatebox{90}{\phantom{AA}4.0} & \includegraphics[width=0.46\textwidth]{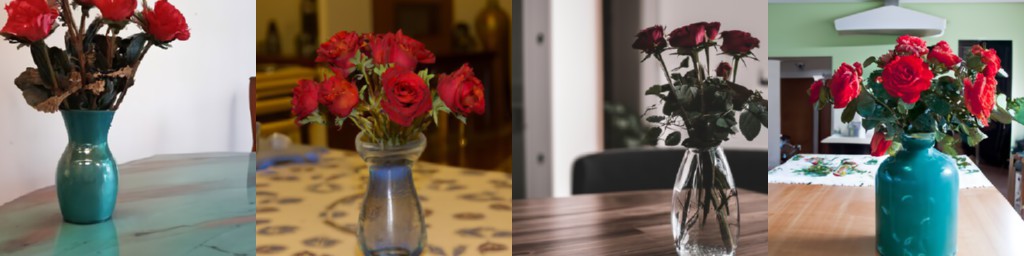} & 
        \includegraphics[width=0.46\textwidth]{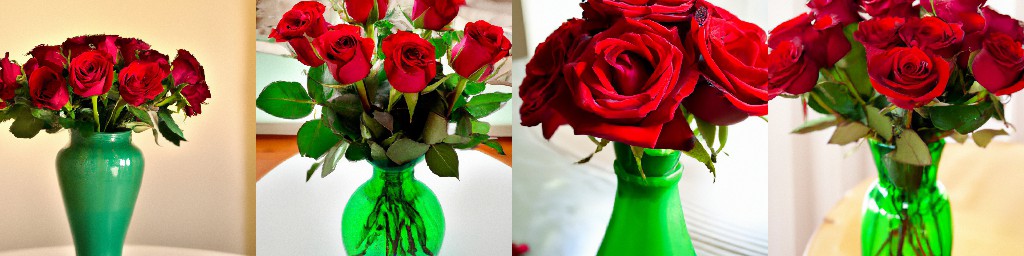} \\
        & unCLIP & GLIDE \\
        \rule{0pt}{0.5pt}
    \end{tabular}
    \vskip -0.2in 
    \caption{Samples when increasing guidance scale for both \modelname{} and GLIDE, using the prompt, ``A green vase filled with red roses sitting on top of table.'' For \modelname{}, we fix the latent vectors sampled from the prior, and only vary the guidance scale of the decoder. For both models, we fix the diffusion noise seed for each column. Samples from \modelname{} improve in quality (more realistic lighting and shadows) but do not change in content as we increase guidance scale, preserving semantic diversity even at high decoder guidance scales.}
    \label{fig:qualitative_diversity}
\end{figure*}

\begin{table}[t]
    \begin{center}
    \begin{tabular}{cccc}
    \toprule
    unCLIP Prior & Photorealism & Caption Similarity & Diversity \\
    \midrule
    AR & 47.1\% $\pm$ 3.1\% & 41.1\% $\pm$ 3.0\% & 62.6\% $\pm$ 3.0\% \\
    Diffusion & 48.9\% $\pm$ 3.1\% & 45.3\% $\pm$ 3.0\% & 70.5\% $\pm$ 2.8\% \\
    \bottomrule
    \end{tabular}
    \end{center}
    \caption{\label{tab:humaneval_realism} Human evaluations comparing \modelname{} to GLIDE. We compare to both the AR and diffusion prior for \modelname{}. Reported figures are 95\% confidence intervals of the probability that the \modelname{} model specified by the row beats GLIDE. Sampling hyperparameters for all models were swept to optimize an automated proxy for human photorealism evaluations.}
    \vskip -0.2in
\end{table}

We observe in Figure \ref{fig:header_samples} that \modelname{} is capable of synthesizing complex, realistic images. While we can compare sample quality to past models using FID, it is not always aligned with human judgment. To better gauge the generation capabilities of our system, we conduct systematic human evaluations comparing \modelname{} to GLIDE for photorealism, caption similarity, and sample diversity. 

We follow the protocol of \namecite{dalle,glide} for the first two evaluations: for photorealism, users are presented with pairs of images and must choose which looks more photorealistic; for caption similarity, users are additionally prompted with a caption, and must choose which image better matches the caption. In both evaluations, there is a third ``Not sure'' option. For diversity, we propose a new evaluation protocol in which humans are presented with two $4 \times 4$ grids of samples and must choose which is more diverse (with a third option, ``Not sure''). For this evaluation, we produce sample grids using 1,000 captions from the MS-COCO validation set, and always compare sample grids for the same caption. Before running human comparisons, we swept over sampling hyperparameters for each model using a CLIP linear probe trained to be a proxy for human photorealism evaluations (Appendix \ref{app:linear_probes}). These hyperparameters are fixed across all three types of evaluation.

We present our results in Table \ref{tab:humaneval_realism}. In general, the diffusion prior performs better than the AR prior in pairwise comparisons against GLIDE. We find that humans still slightly prefer GLIDE to \modelname{} in terms of photorealism, but the gap is very small. Even with similar photorealism, \modelname{} is strongly preferred over GLIDE in terms of diversity, highlighting one of its benefits.

\subsection{Improved Diversity-Fidelity Trade-off with Guidance}
\label{sec:tradeoff}
\begin{figure}[t]
    \begin{center}
    \includegraphics[width=0.7\textwidth]{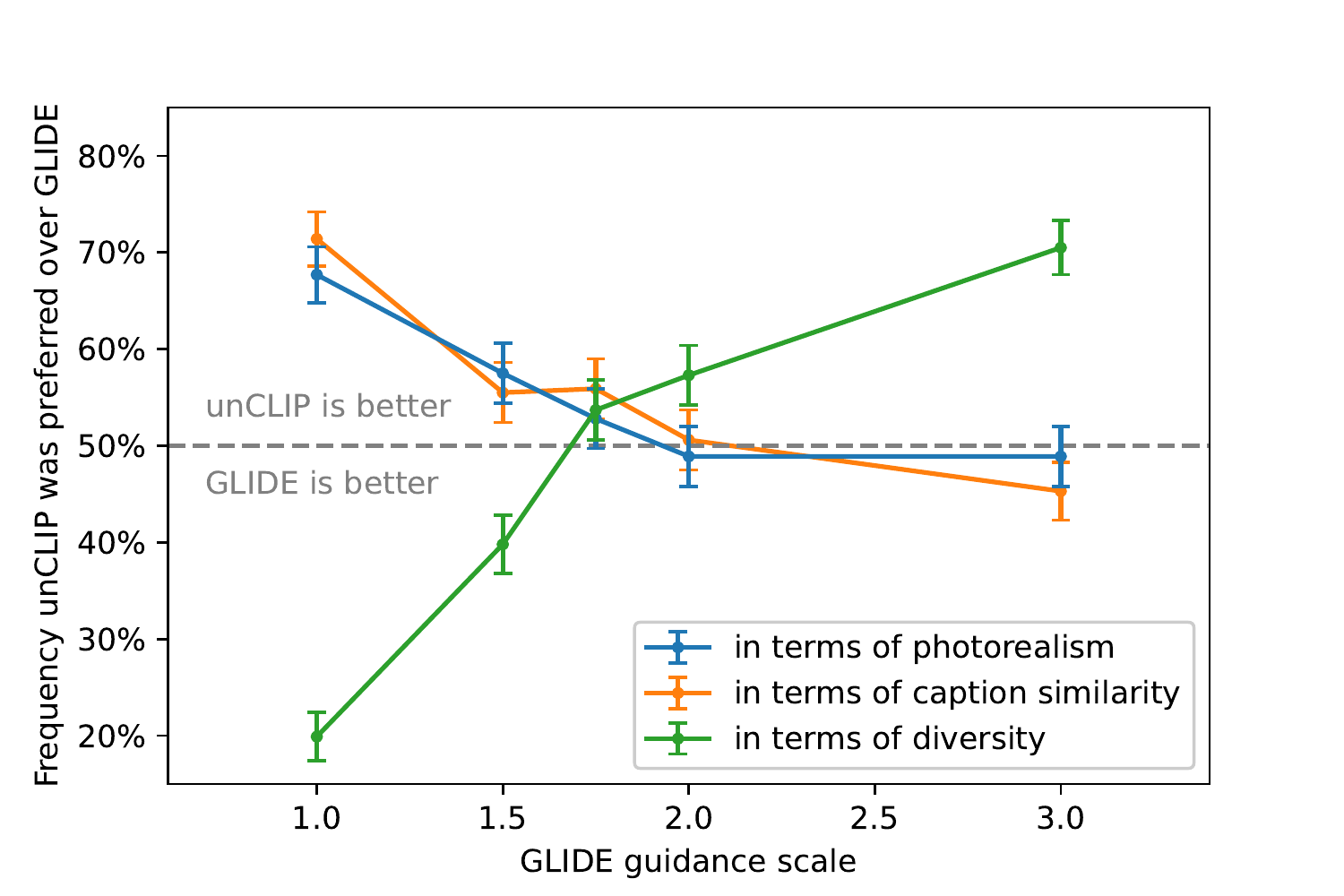}
    \end{center}
    \vskip -0.1in
    \caption{When comparing \modelname{} (with our best sampling settings) to various settings of guidance scale for GLIDE, \modelname{} was preferred by human evaluators on at least one axis among photorealism, caption similarity, and diversity for each comparison. At the higher guidance scales used to generate photorealistic images, unCLIP yields greater diversity for comparable photorealism and caption similarity.}
    \label{fig:humaneval_pareto}
    \vskip -0.1in
\end{figure}

\begin{figure}[t]
    \begin{center}
    \includegraphics[width=0.5\textwidth]{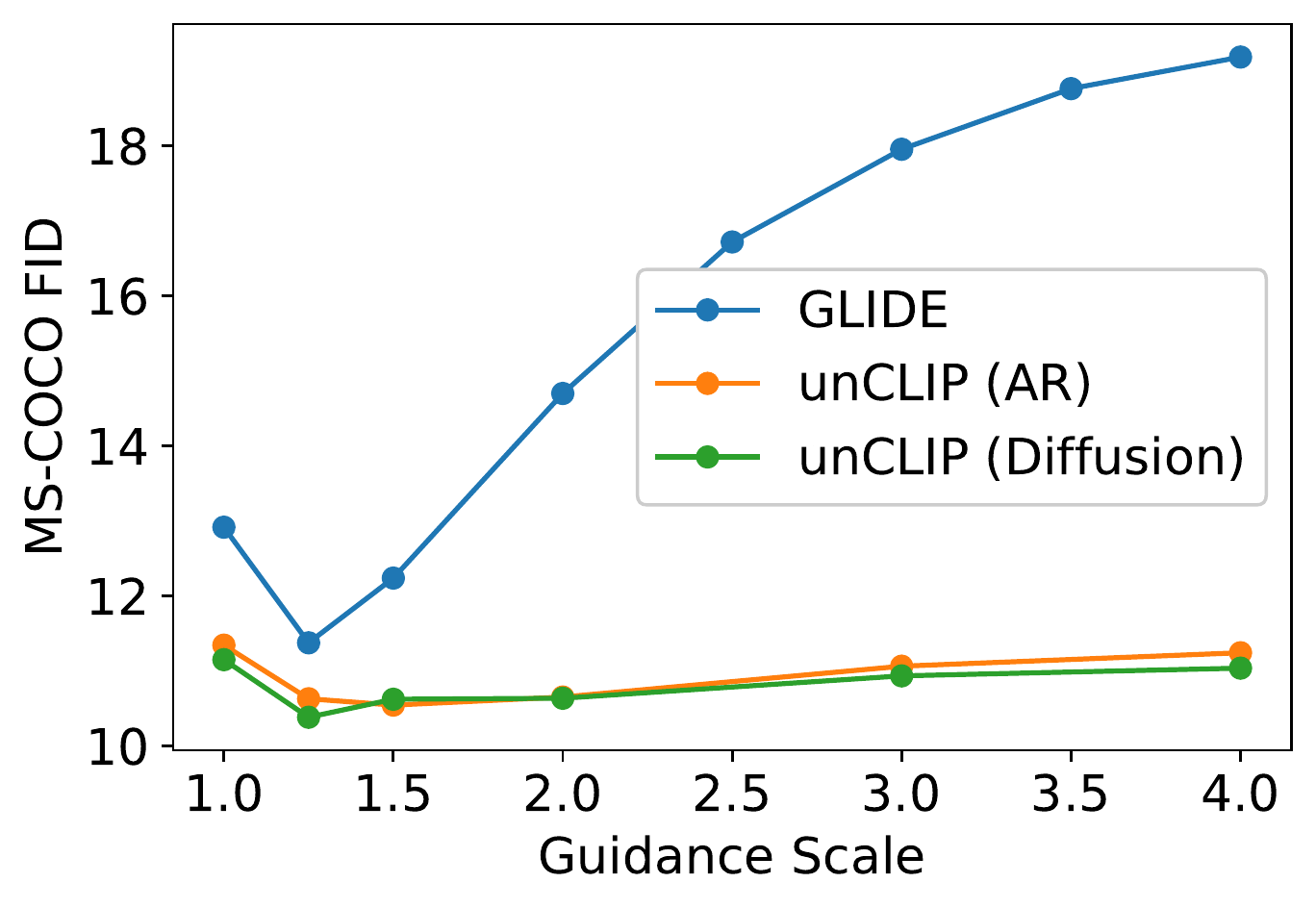}
    \end{center}
    \vskip -0.1in
    \caption{FID versus guidance scale for \modelname{} and GLIDE. For the \modelname{} priors, we swept over sampling hyperparameters and fixed to the settings with the best minimum FID.}
    \label{fig:fid_plot}
    \vskip -0.1in
\end{figure}

Compared to GLIDE, we qualitatively observe that unCLIP is able to generate more diverse images while leveraging the guidance technique to improve sample quality. To understand why, consider Figure \ref{fig:qualitative_diversity} where we increase guidance scale for both GLIDE and \modelname{}. For GLIDE, the semantics (camera angle, color, size) converge as we increase guidance scale, whereas for \modelname{} the semantic information of the scene is frozen in the CLIP image embedding and therefore does not collapse when guiding the decoder. 

In Section \ref{sec:human_evals}, we observed that \modelname{} achieves similar photorealism as GLIDE while maintaining more diversity, but that its caption matching capabilities were slightly worse. It is natural to ask whether GLIDE's guidance scale can be lowered to obtain the same diversity level as \modelname{} while maintaining better caption matching. In Figure \ref{fig:humaneval_pareto}, we conduct a more careful study of this question by performing human evaluations across several GLIDE guidance scales. We find that GLIDE at guidance scale 2.0 is very close to the photorealism and caption similarity of \modelname{}, while still producing less diverse samples.

Finally, in Figure \ref{fig:fid_plot} we compute MS-COCO zero-shot FID \shortcite{fid} while sweeping over guidance scale for both \modelname{} and GLIDE, finding that guidance hurts the FID of \modelname{} much less so than for GLIDE. In this evaluation, we fix the guidance scale of the \modelname{} prior and only vary the guidance scale of the decoder. This is another indication that guidance hurts the diversity of GLIDE much more than \modelname{}, since FID heavily penalizes non-diverse generations.

\subsection{Comparison on MS-COCO}
\label{sec:fid}

\begin{table}[t]
    \centering
    \begin{center}
    \begin{small}
    \begin{tabular}{cccc}
    \toprule
    Model & FID & Zero-shot FID & Zero-shot FID (filt) \\
    \midrule
    AttnGAN \citep{attngan} & 35.49 & & \\
    DM-GAN \citep{dmgan} & 32.64 & & \\
    DF-GAN \citep{dfgan} & 21.42 & & \\
    DM-GAN + CL \citep{textcl} & 20.79 & & \\
    XMC-GAN \citep{xmcgan} & 9.33 & & \\
    LAFITE \citep{lafite} & 8.12 & & \\
    Make-A-Scene \citep{makeascene} & \textbf{7.55} & & \\
    \midrule
    DALL-E \citep{dalle} & & $\sim$ 28 & \\
    LAFITE \citep{lafite} & & 26.94 & \\
    GLIDE \citep{glide} & & 12.24 & 12.89 \\
    Make-A-Scene \citep{makeascene} & & & 11.84 \\
    \modelname{} (AR prior) & & 10.63 & 11.08 \\
    \modelname{} (Diffusion prior) & & \textbf{10.39} & \textbf{10.87} \\
    \bottomrule
    \end{tabular}
    \end{small}
    \end{center}
    \caption{\label{tab:mscoco_fids} Comparison of FID on MS-COCO $256 \times 256$. We use guidance scale 1.25 for the decoder for both the AR and diffusion prior, and achieve the best results using the diffusion prior.}
    \vskip -0.2in
\end{table}

\begin{figure}[ht!]
    \centering
    \setlength{\tabcolsep}{2.0pt}
    \begin{tabular}{cccccc}
        \rotatebox{90}{\scriptsize\phantom{AAAA} Real Image} &
        \includegraphics[width=0.19\textwidth]{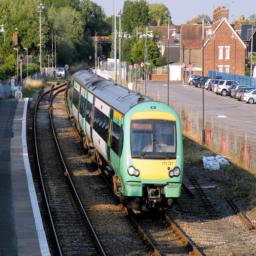} &
        \includegraphics[width=0.19\textwidth]{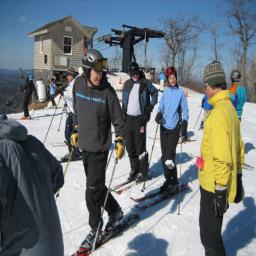} &
        \includegraphics[width=0.19\textwidth]{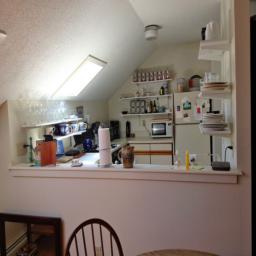} &
        \includegraphics[width=0.19\textwidth]{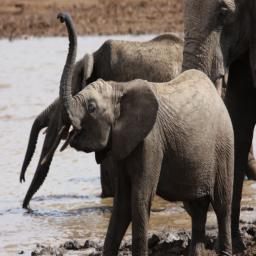} &
        \includegraphics[width=0.19\textwidth]{figures/model_comparison/original/4.jpg} \\

        \rotatebox{90}{\scriptsize\phantom{AAAAA} DALL-E} &
        \includegraphics[width=0.19\textwidth]{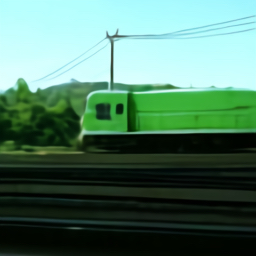} &
        \includegraphics[width=0.19\textwidth]{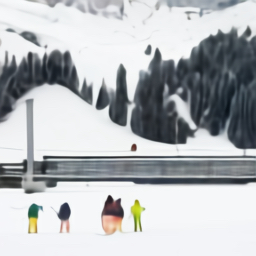} &
        \includegraphics[width=0.19\textwidth]{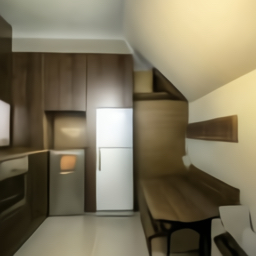} &
        \includegraphics[width=0.19\textwidth]{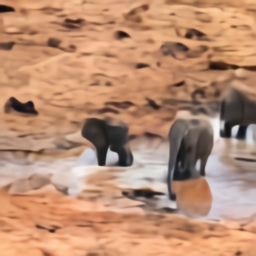} &
        \includegraphics[width=0.19\textwidth]{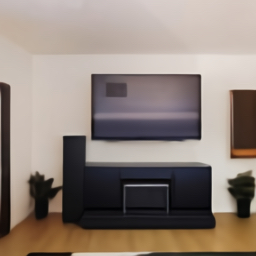} \\

        \rotatebox{90}{\scriptsize\phantom{AAAAA} GLIDE} &
        \includegraphics[width=0.19\textwidth]{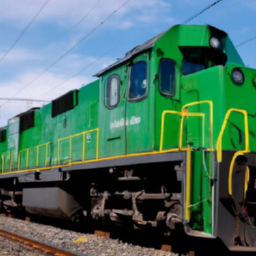} &
        \includegraphics[width=0.19\textwidth]{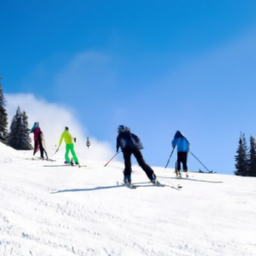} &
        \includegraphics[width=0.19\textwidth]{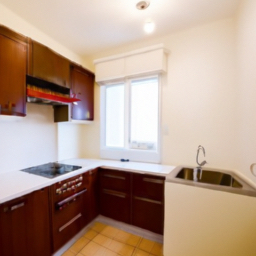} &
        \includegraphics[width=0.19\textwidth]{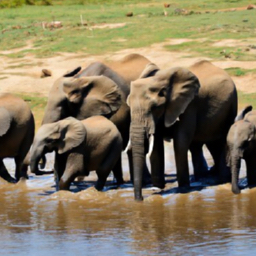} &
        \includegraphics[width=0.19\textwidth]{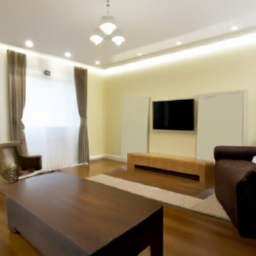} \\
        
        \rotatebox{90}{\scriptsize\phantom{AAA} Make-A-Scene} &
        \includegraphics[width=0.19\textwidth]{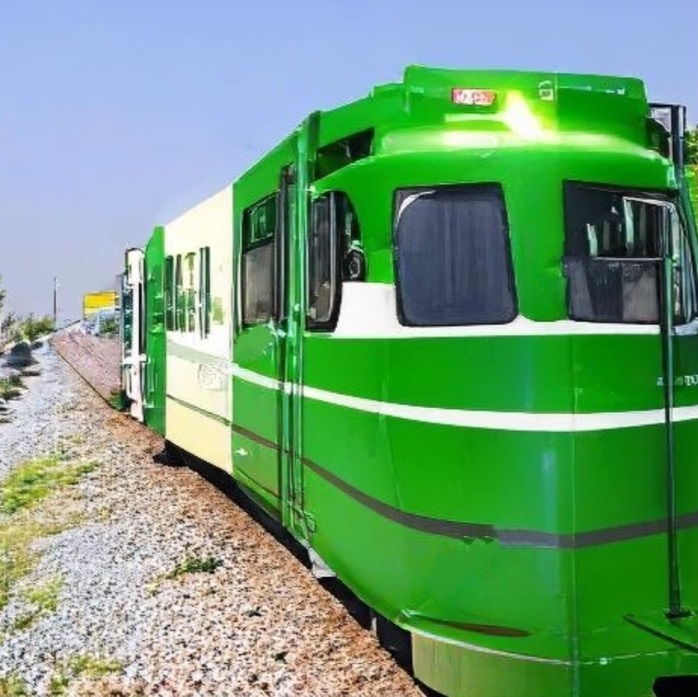} &
        \includegraphics[width=0.19\textwidth]{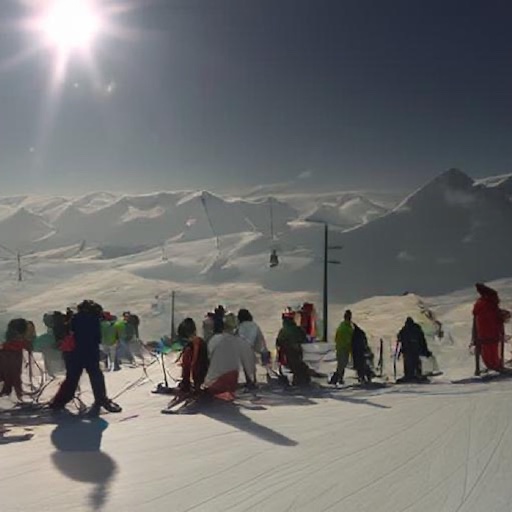} &
        \includegraphics[width=0.19\textwidth]{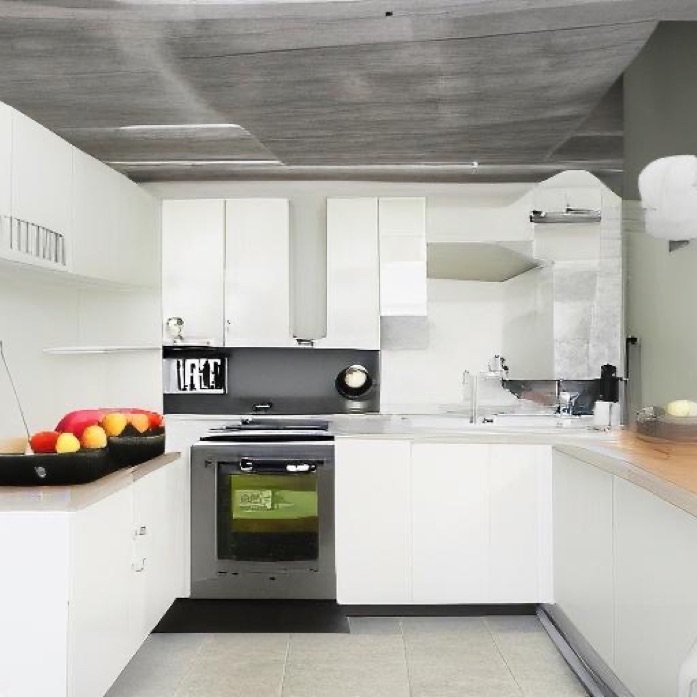} &
        \includegraphics[width=0.19\textwidth]{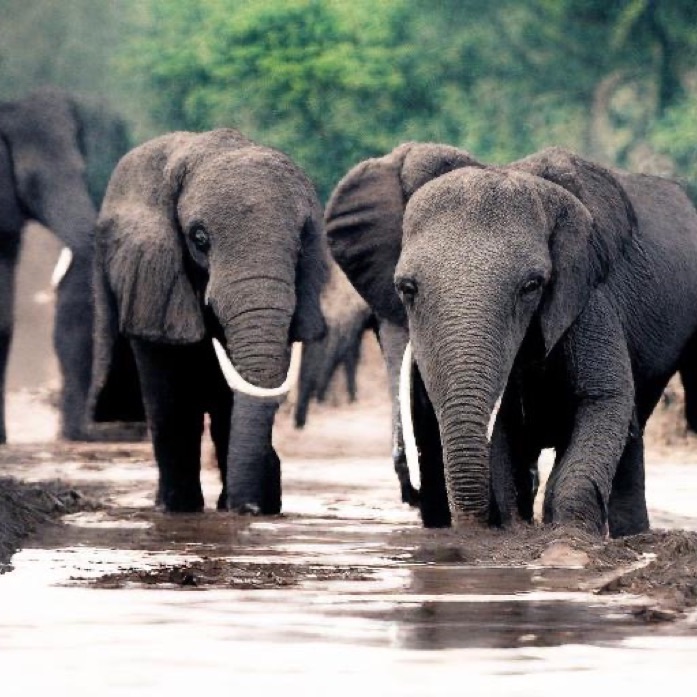} &
        \includegraphics[width=0.19\textwidth]{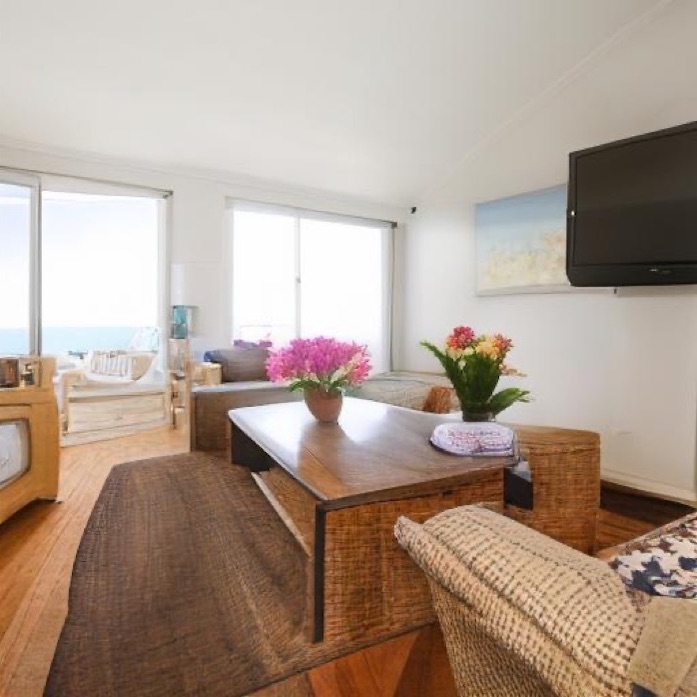} \\
        
        \rotatebox{90}{\scriptsize\phantom{AAAA} \modelname{}} &
        \includegraphics[width=0.19\textwidth]{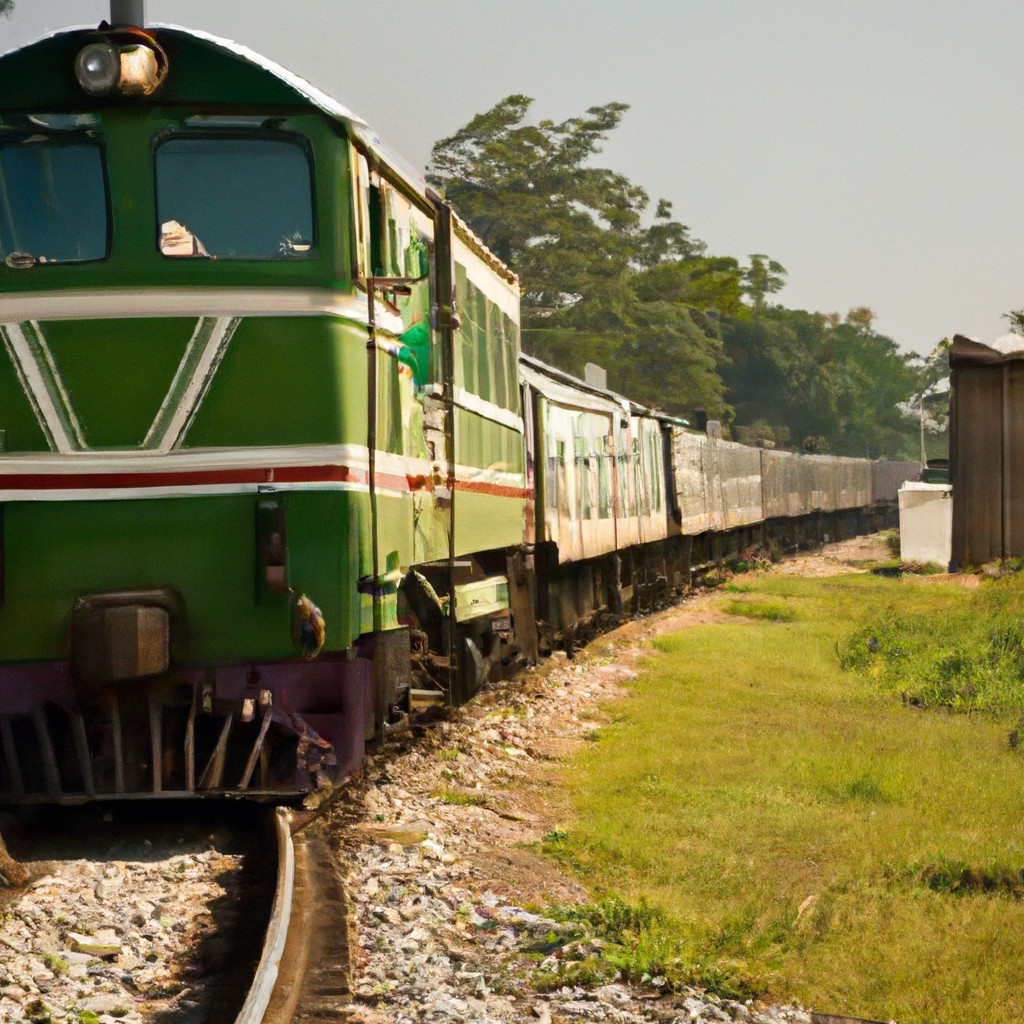} &
        \includegraphics[width=0.19\textwidth]{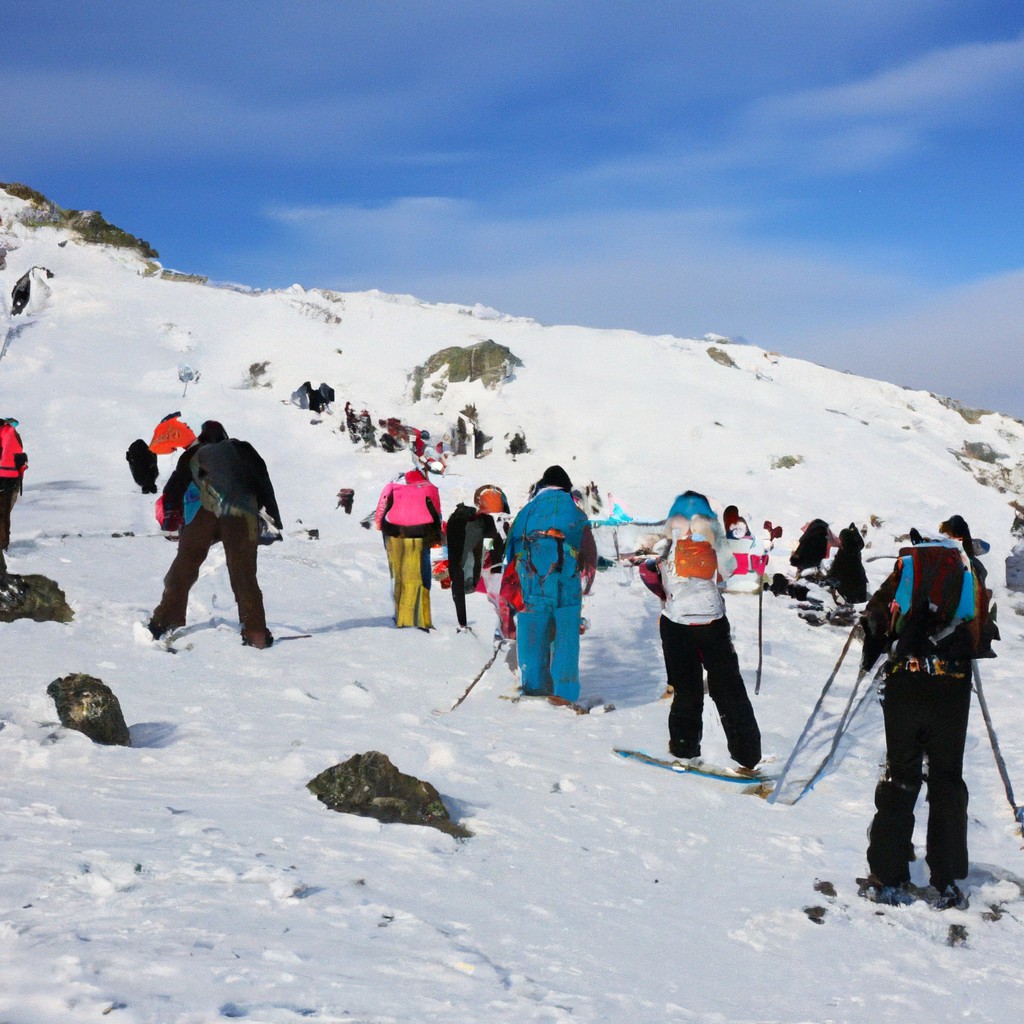} &
        \includegraphics[width=0.19\textwidth]{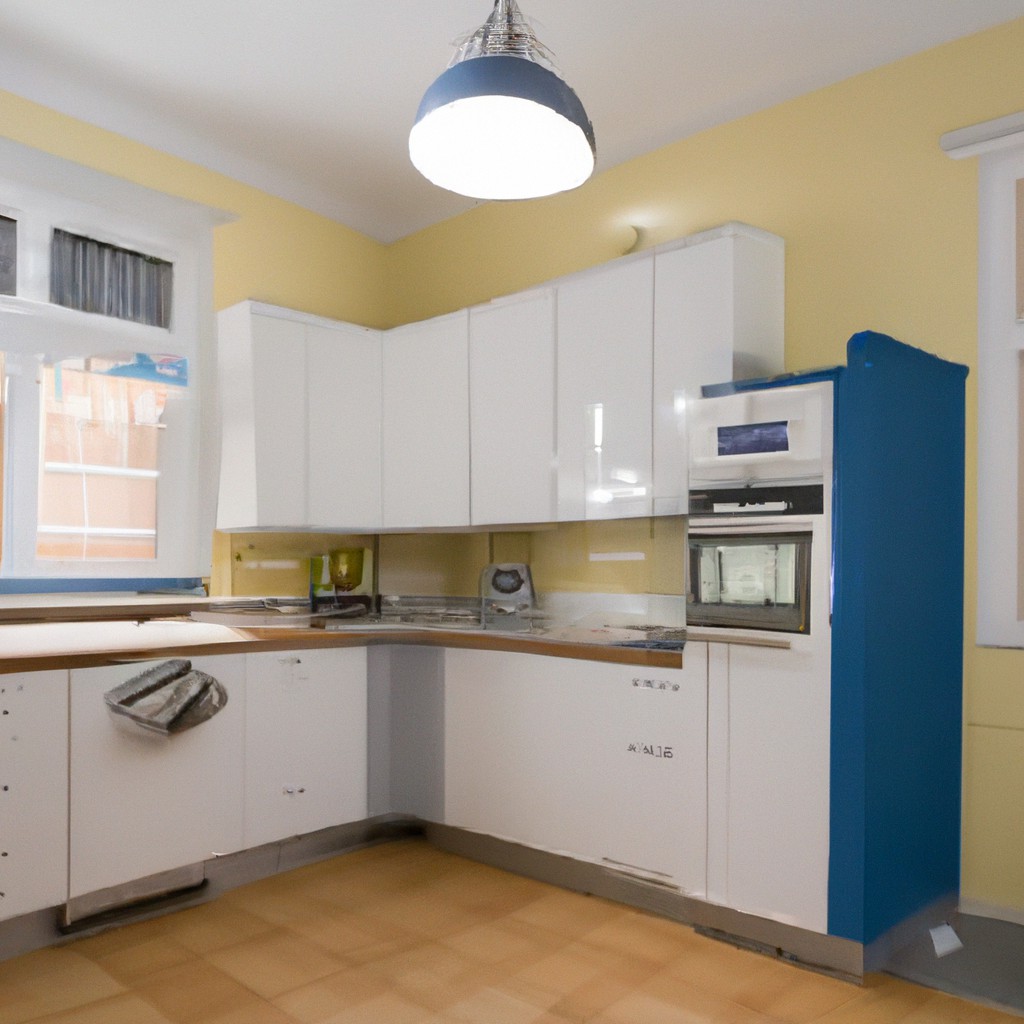} &
        \includegraphics[width=0.19\textwidth]{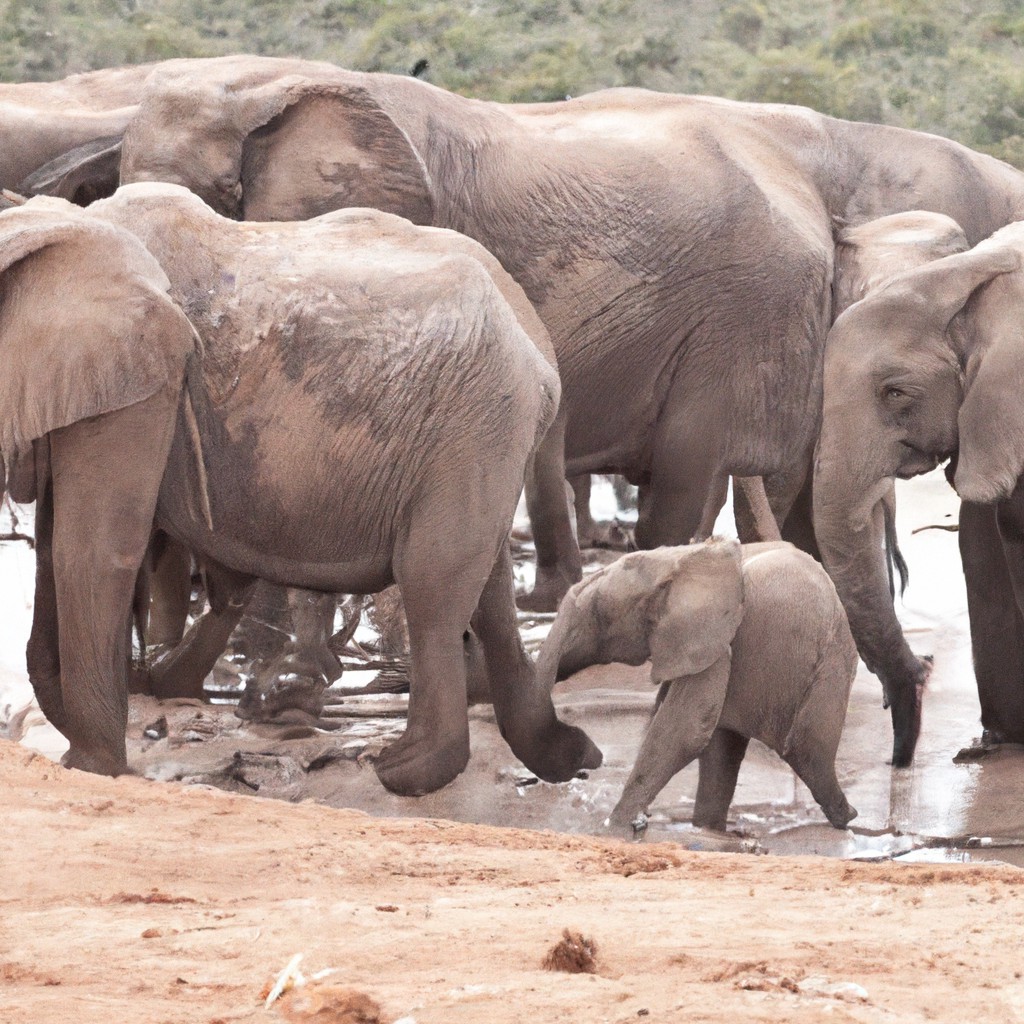} &
        \includegraphics[width=0.19\textwidth]{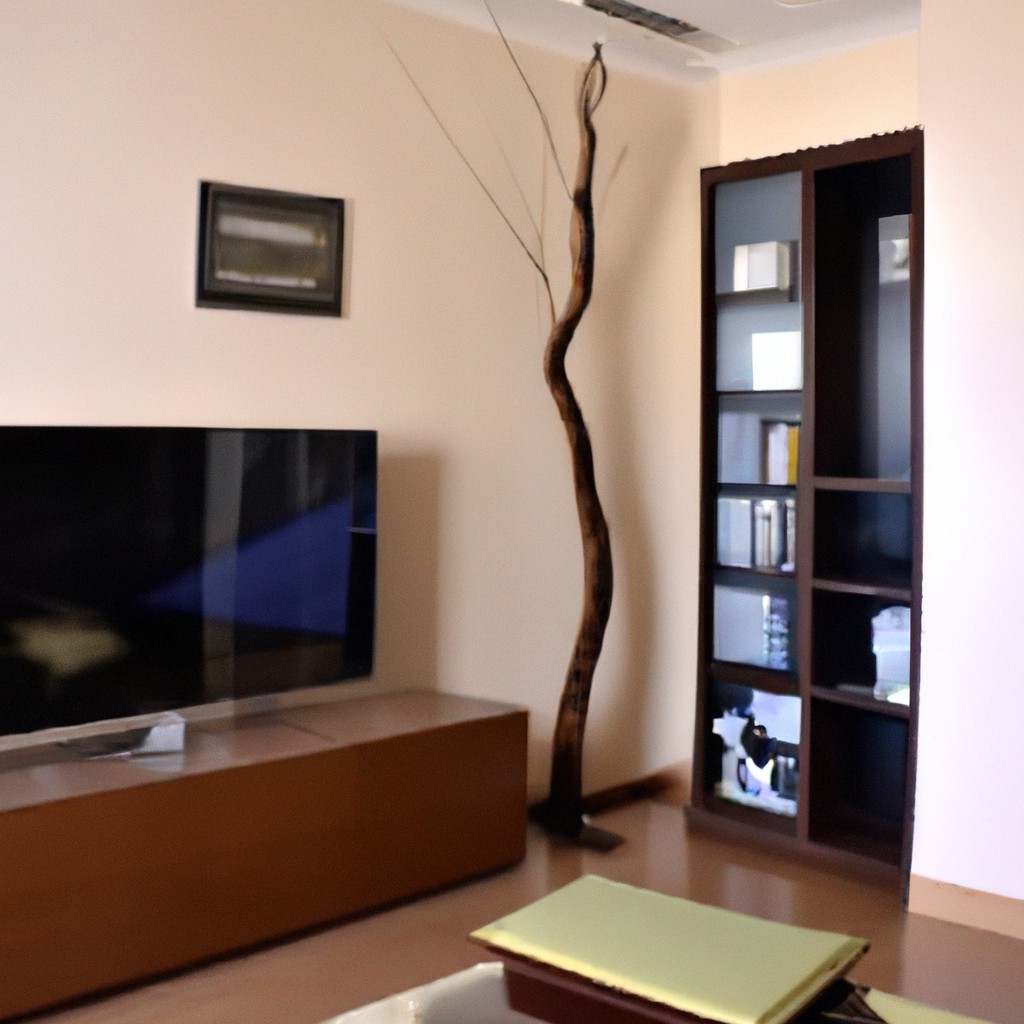} \\
        
        \rotatebox{90}{\scriptsize\phantom{AAAA} \modelname{} (prod.)} &
        \includegraphics[width=0.19\textwidth]{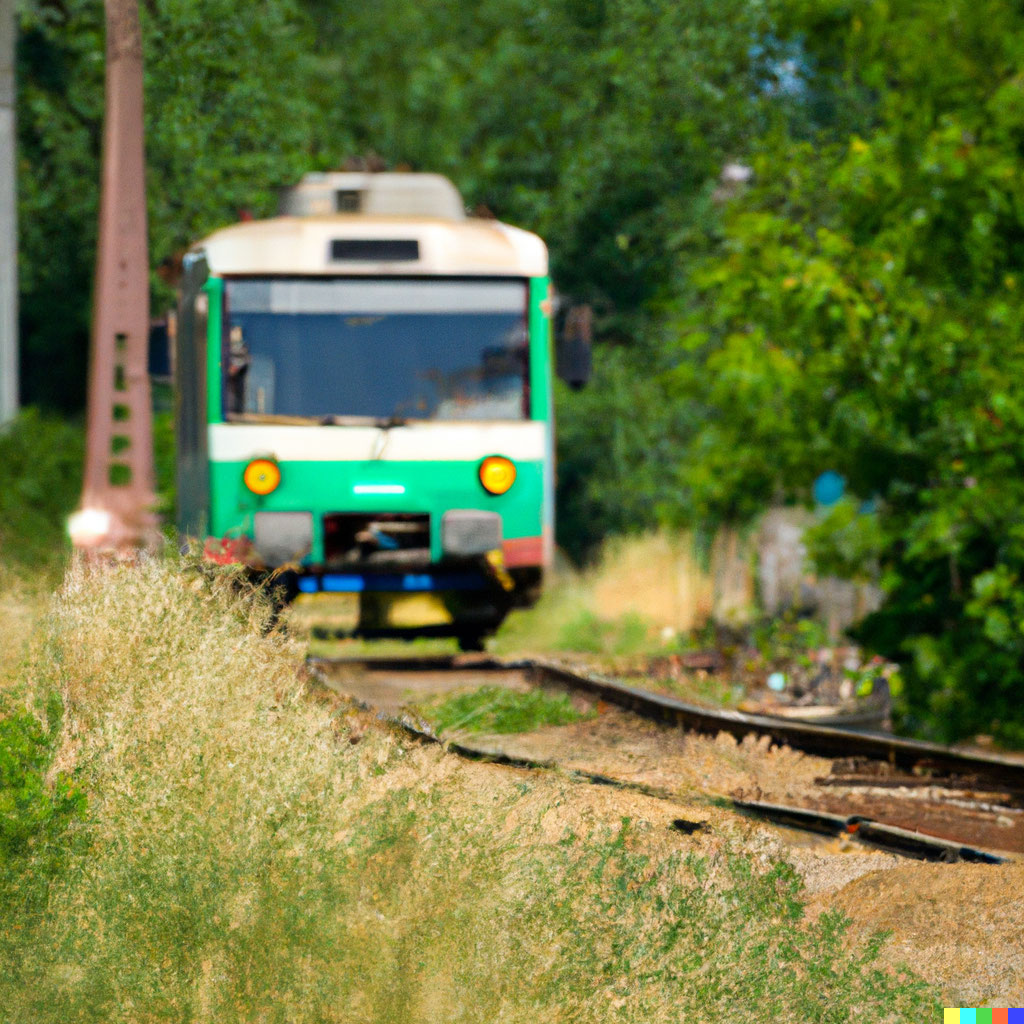} &
        \includegraphics[width=0.19\textwidth]{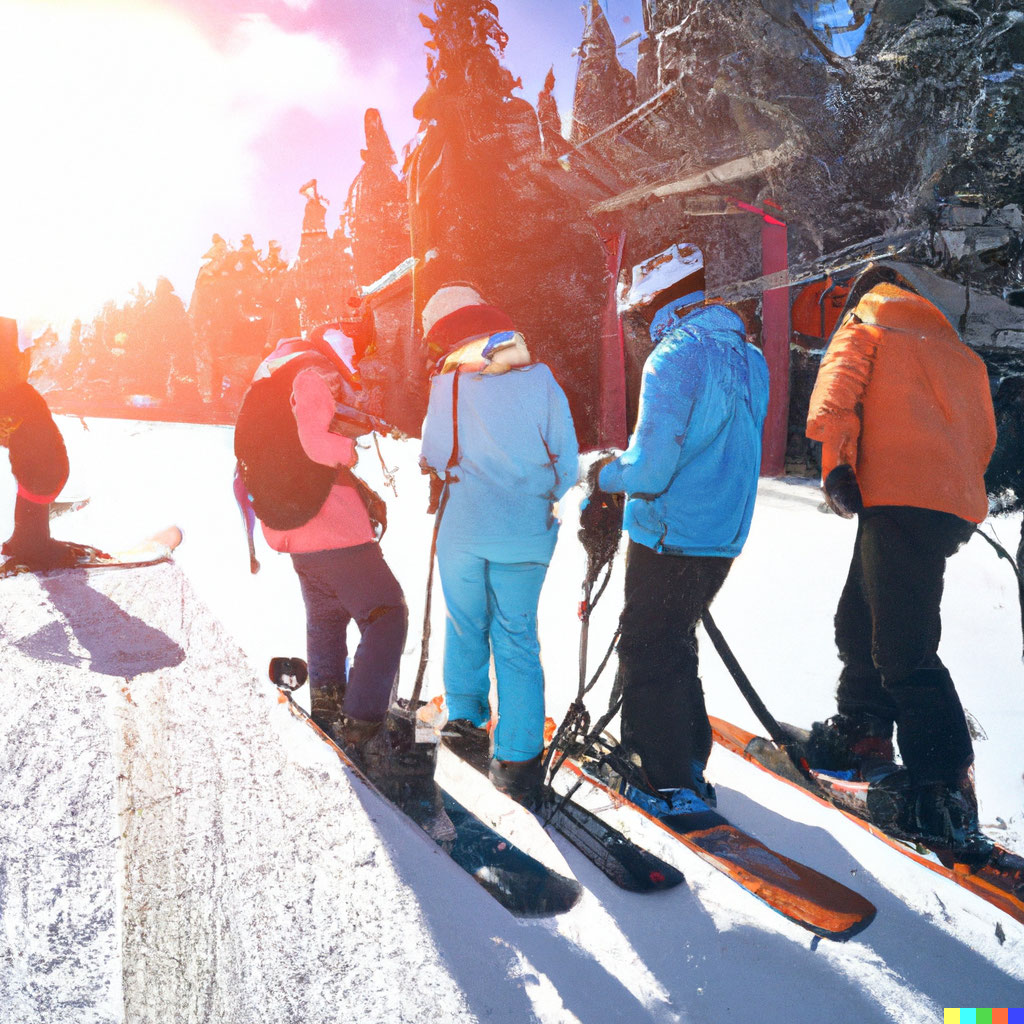} &
        \includegraphics[width=0.19\textwidth]{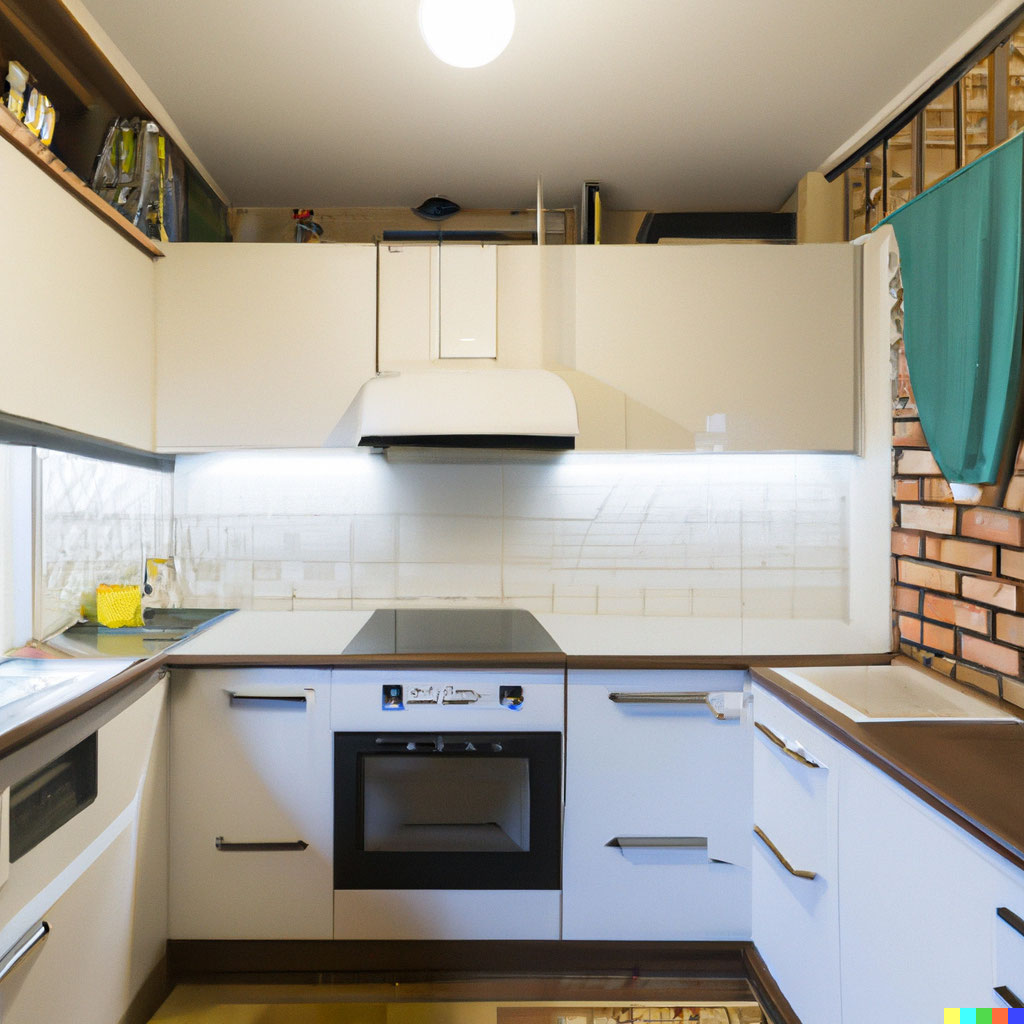} &
        \includegraphics[width=0.19\textwidth]{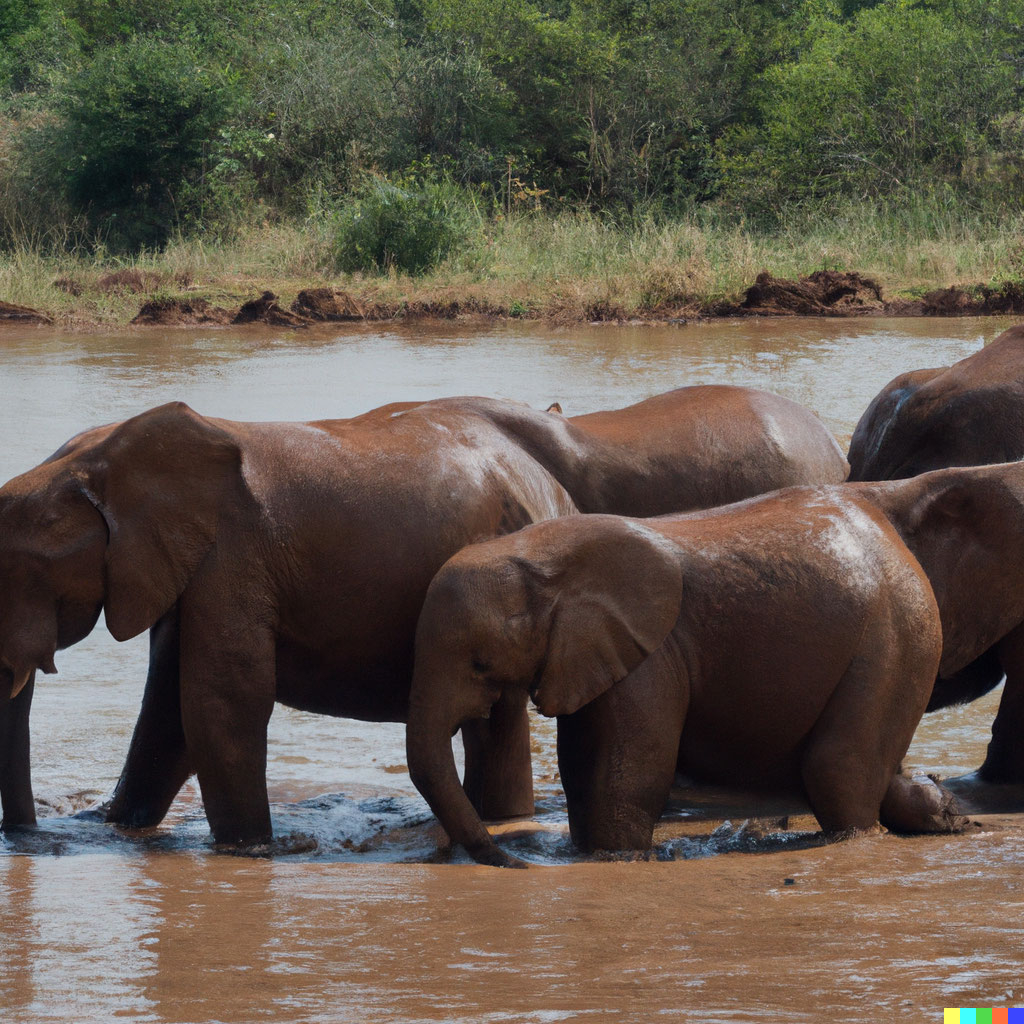} &
        \includegraphics[width=0.19\textwidth]{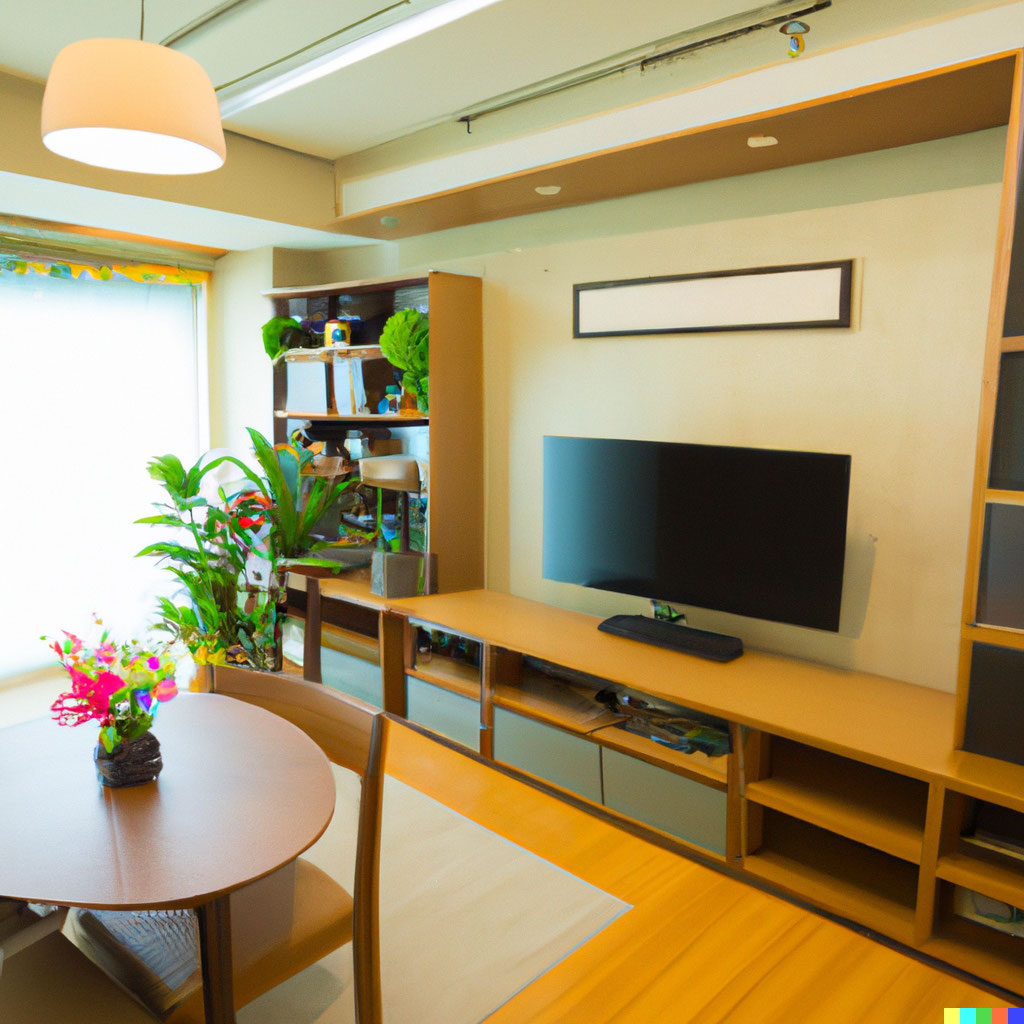} \\

        & \scriptsize \makecell{``a green train is coming \\ down the tracks''}
        & \scriptsize \makecell{``a group of skiers are \\ preparing to ski down \\ a mountain.''}
        & \scriptsize \makecell{``a small kitchen with \\ a low ceiling''}
        & \scriptsize \makecell{``a group of elephants \\ walking in muddy \\ water.''}
        & \scriptsize \makecell{``a living area with a \\ television and a table''}
    \end{tabular}
    \caption{Random image samples on MS-COCO prompts.}
    \label{fig:coco_model_comparison}
    \vskip -0.2in
\end{figure}

In the text-conditional image generation literature, it has become standard practice to evaluate FID on the MS-COCO \shortcite{mscoco} validation set. We present results on this benchmark in Table \ref{tab:mscoco_fids}. Like GLIDE and DALL-E, \modelname{} is not directly trained on the MS-COCO training set, but can still generalize to the validation set zero-shot. We find that, compared to these other zero-shot models, \modelname{} achieves a new state-of-the-art FID of 10.39 when sampling with the diffusion prior. In Figure \ref{fig:coco_model_comparison}, we visually compare \modelname{} to various recent text-conditional image generation models on several captions from MS-COCO. We find that, like the other methods, \modelname{} produces realistic scenes that capture the text prompts.

\subsection{Aesthetic Quality Comparison}
\label{sec:aesthetic}

We additionally perform automated aesthetic quality evaluations comparing \modelname{} to GLIDE. Our goal with this evaluation is to assess how well each model produces artistic illustrations and photographs. To this end, we generated 512 ``artistic'' captions using GPT-3 \shortcite{gpt3} by prompting it with captions for existing artwork (both real and AI generated). Next, we trained a CLIP linear probe to predict human aesthetic judgments using the AVA dataset \shortcite{avadataset} (Appendix \ref{app:linear_probes}). For each model and set of sampling hyperparameters, we produce four images for each prompt, and report the mean predicted aesthetic judgment over the full batch of 2048 images.

\begin{figure}[t]
    \begin{center}
    \begin{subfigure}{0.475\textwidth}
        \centering
        \includegraphics[width=\textwidth]{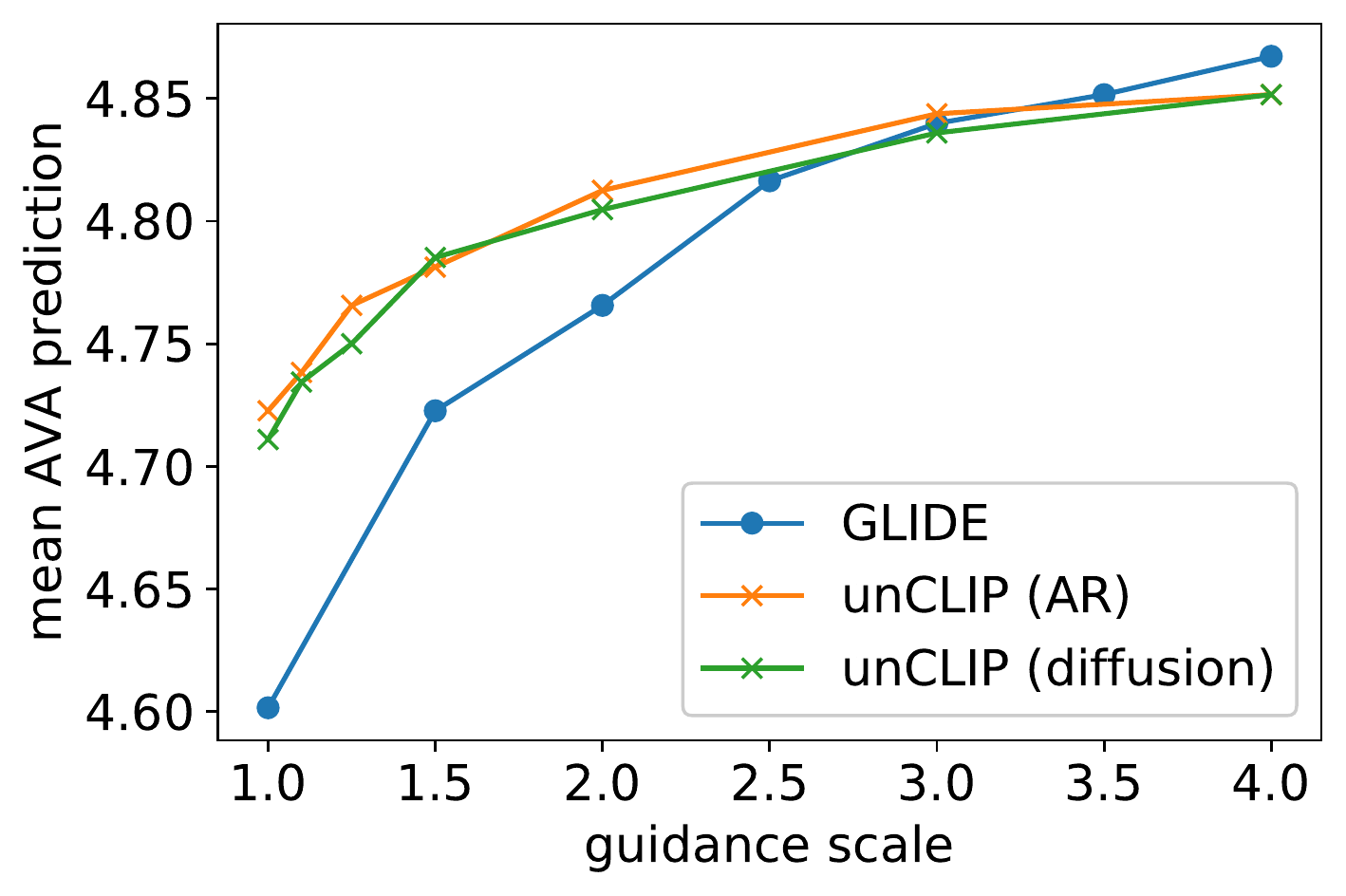}
    \end{subfigure}
    \begin{subfigure}{0.475\textwidth}
        \centering
        \includegraphics[width=\textwidth]{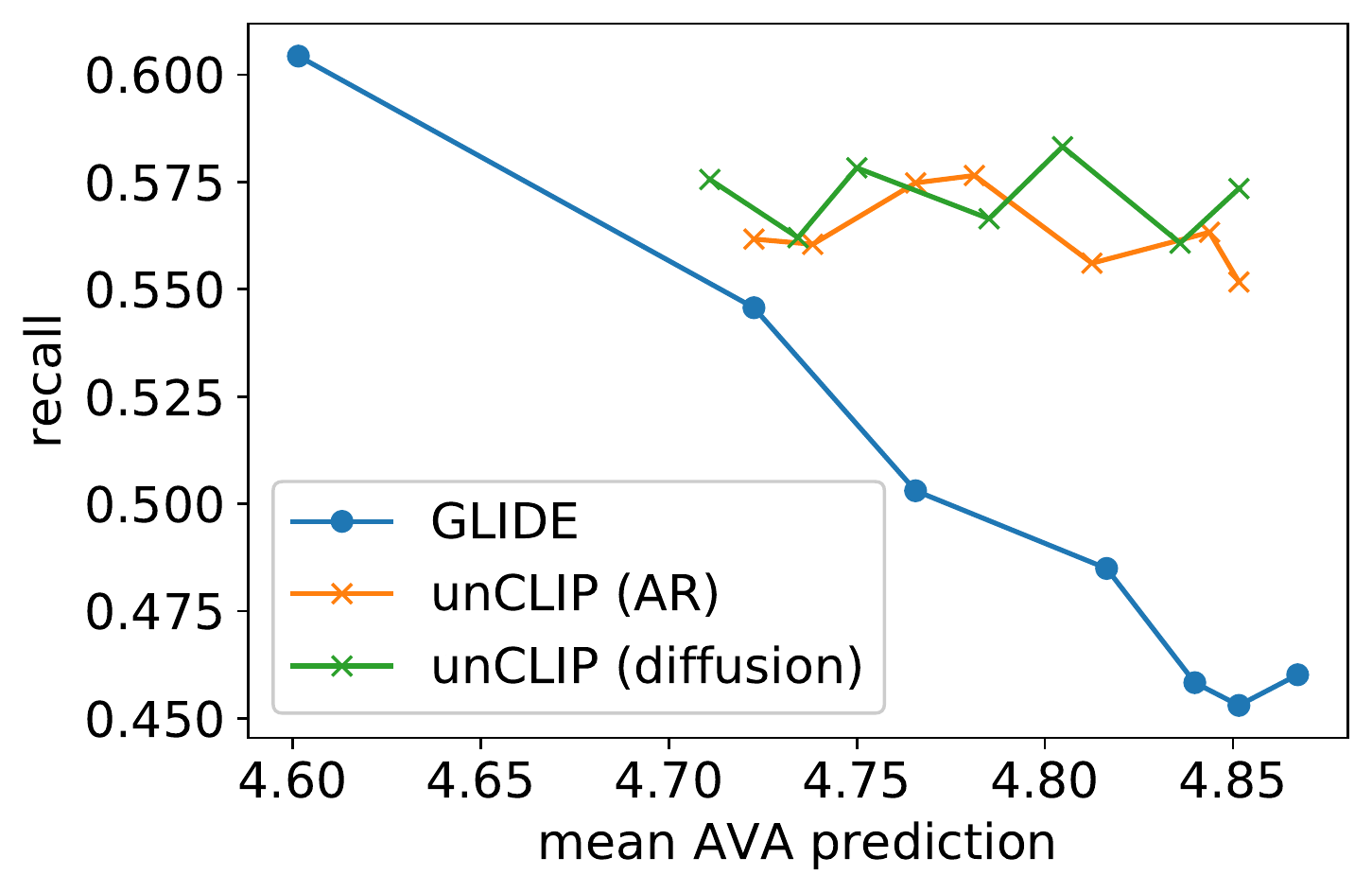}
    \end{subfigure}
    \end{center}
    \vskip -0.1in
    \caption{Aesthetic quality evaluations comparing GLIDE and \modelname{} using 512 auto-generated artistic prompts. We find that both models benefit from guidance, but \modelname{} does not sacrifice recall for aesthetic quality.}
    \label{fig:ava_results}
    \vskip -0.1in
\end{figure}

In Figure \ref{fig:ava_results}, we present results on our aesthetic quality evaluation. We find that guidance improves aesthetic quality for both GLIDE and \modelname{}. For \modelname{}, we only guide the decoder (we found that guiding the prior hurt results). We also plot the aesthetic quality against Recall\footnote{Recall is computed with respect to the training dataset.}, since guidance typically induces a trade-off 
\clearpage
\newpage
between fidelity and diversity. Interestingly, we find that guiding \modelname{} does not decrease Recall while still improving aesthetic quality according to this metric.
\section{Related Work}

Synthetic image generation is a well studied problem, and most popular techniques for unconditional image generation have also been applied to the text-conditional setting. Many previous works have trained GANs \shortcite{gan} on publicly available image captioning datasets to produce text-conditional image samples \shortcite{attngan,dmgan,dfgan,xmcgan,textcl}. Other works have adapted the VQ-VAE approach \shortcite{vqvae} to text-conditional image generation by training autoregressive transformers on sequences of text tokens followed by image tokens \shortcite{dalle,cogview,cm3}. Finally, some works have applied diffusion models to the problem, training either continuous \shortcite{glide} or discrete \shortcite{vqdiff} diffusion models with auxiliary text encoders to handle textual input.

Previous works have leveraged hierarchical generative processes to create high-quality synthetic images. \namecite{vqvae2} trains a multi-layer discrete autoencoder, allowing them to first sample coarse-grained latent codes and then use this as conditioning information when sampling higher-resolution latent codes. \namecite{vdvae,nvae} generate images using VAEs with a hierarchy of latent codes that increase progressively with resolution. Concurrently with our work, \namecite{makeascene} conditions a generative image model on segmentation masks, allowing for a generative process that first samples a semantic map of an image and then conditions the generated image on this information.

The computational benefits of using diffusion to model a latent space has been noted by previous works. \namecite{diffae} propose an autoencoder framework where diffusion models are used to render latent variables as images, and a second diffusion model is used to generate these latents (similar to our diffusion prior). \namecite{scorelatent} use a score-based model for the latent space of a VAE, while \namecite{latentdiffusion} use diffusion models on the latents obtained from a VQGAN \shortcite{vqgan} like autoencoder.

Since its release, CLIP \shortcite{clip} has been used extensively to steer generative image models towards text prompts. \namecite{clipglass,styleclip,bigsleep,stylegannada} guide GANs using gradients from a CLIP model. For diffusion models, \namecite{sotapaper} introduced classifier guidance as a way to use gradients from a classifier trained on noised images to steer the model towards higher quality generations. \namecite{glide} train a CLIP model on noised images and guide a text-conditional diffusion model, while \namecite{clipdiff,secondarymodelmethod} use an unnoised CLIP model to guide unconditional or class-conditional diffusion models. \namecite{uncond} introduced classifier-free guidance and showed that one can perform guidance implictly from the predictions of the model with and without the conditioning information, thus removing the need for a classifier. \namecite{glide} showed classifier-free guidance works more favorably than CLIP guidance for text conditional image generation. 

Several previous works have trained generative image models that are directly conditioned on CLIP embeddings. \namecite{lafite} condition GAN models on randomly perturbed CLIP image embeddings, finding that these models can generalize to CLIP text embeddings to produce text-conditional images. \namecite{vdiffusion} trained diffusion models conditioned on CLIP text embeddings, allowing for direct text-conditional image generation. \namecite{clipgen} train an autoregressive generative model conditioned on CLIP image embeddings, finding that it generalizes to CLIP text embeddings well enough to allow for text-conditional image synthesis.

\namecite{invertingssl} train diffusion models conditioned on image representations from contrastive models. While the diffusion models themselves cannot generate images unconditionally, the authors experimented with a simple approach for two-stage image generation by employing Kernel Density Estimation to sample image representations. By feeding these generated representations to the diffusion model, they can generate images end-to-end in a way similar to our proposed technique. However, our work differs from this in two ways: first, we use multimodal contrastive representations rather than image-only representations; second, we employ much more powerful generative models for the first stage of the generation hierarchy, and these generative models are conditioned on text.

\begin{figure}[t]
    \begin{center}
    \begin{subfigure}{0.47\textwidth}
        \centering
        \includegraphics[width=\textwidth]{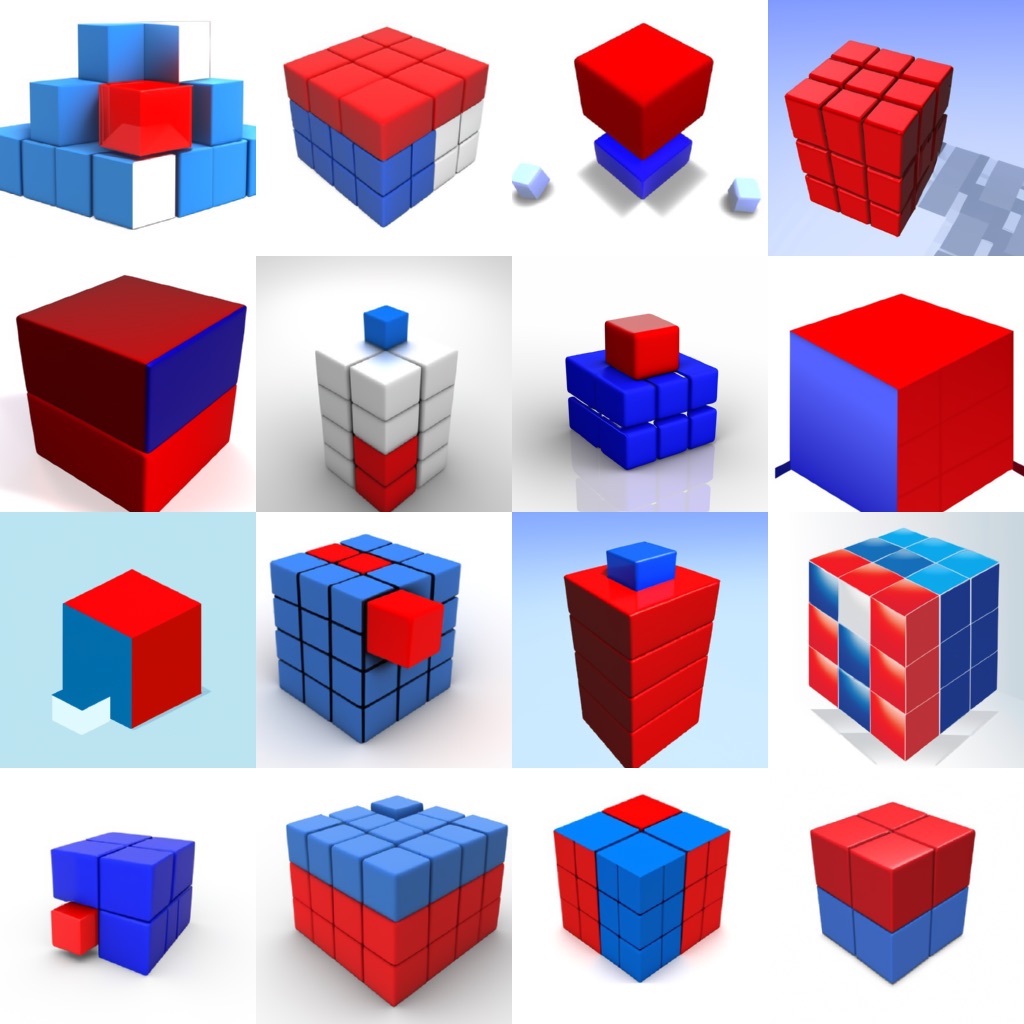}
        \caption{\modelname{}}
    \end{subfigure}
    \begin{subfigure}{0.47\textwidth}
        \centering
        \includegraphics[width=\textwidth]{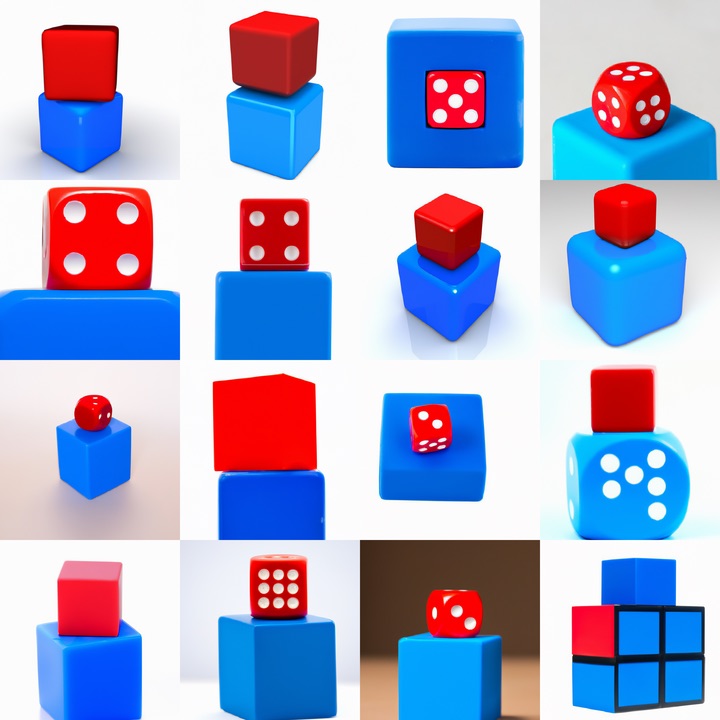}
        \caption{GLIDE}
    \end{subfigure}
    \end{center}
    \vskip -0.1in
    \caption{Samples from \modelname{} and GLIDE for the prompt ``a red cube on top of a blue cube''.}
    \label{fig:block_stacking}
    \vskip -0.1in
\end{figure}

\begin{figure*}[t]
    \centering
    \setlength{\tabcolsep}{10.0pt}
    \begin{tabular}{ccc}
        \includegraphics[width=0.2\textwidth]{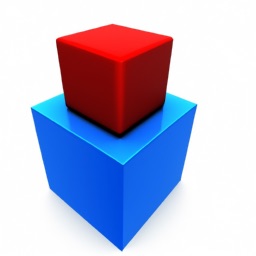} &
        \includegraphics[width=0.2\textwidth]{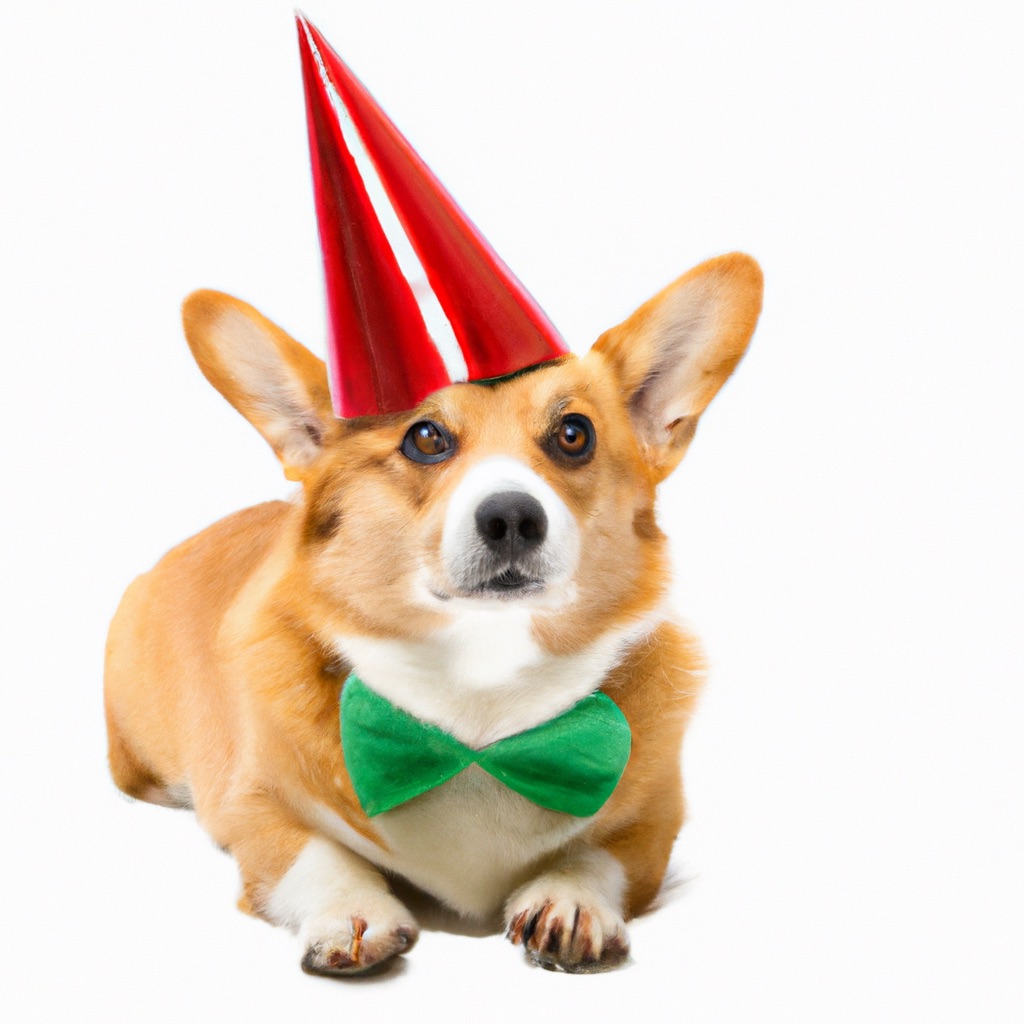} &
        \includegraphics[width=0.2\textwidth]{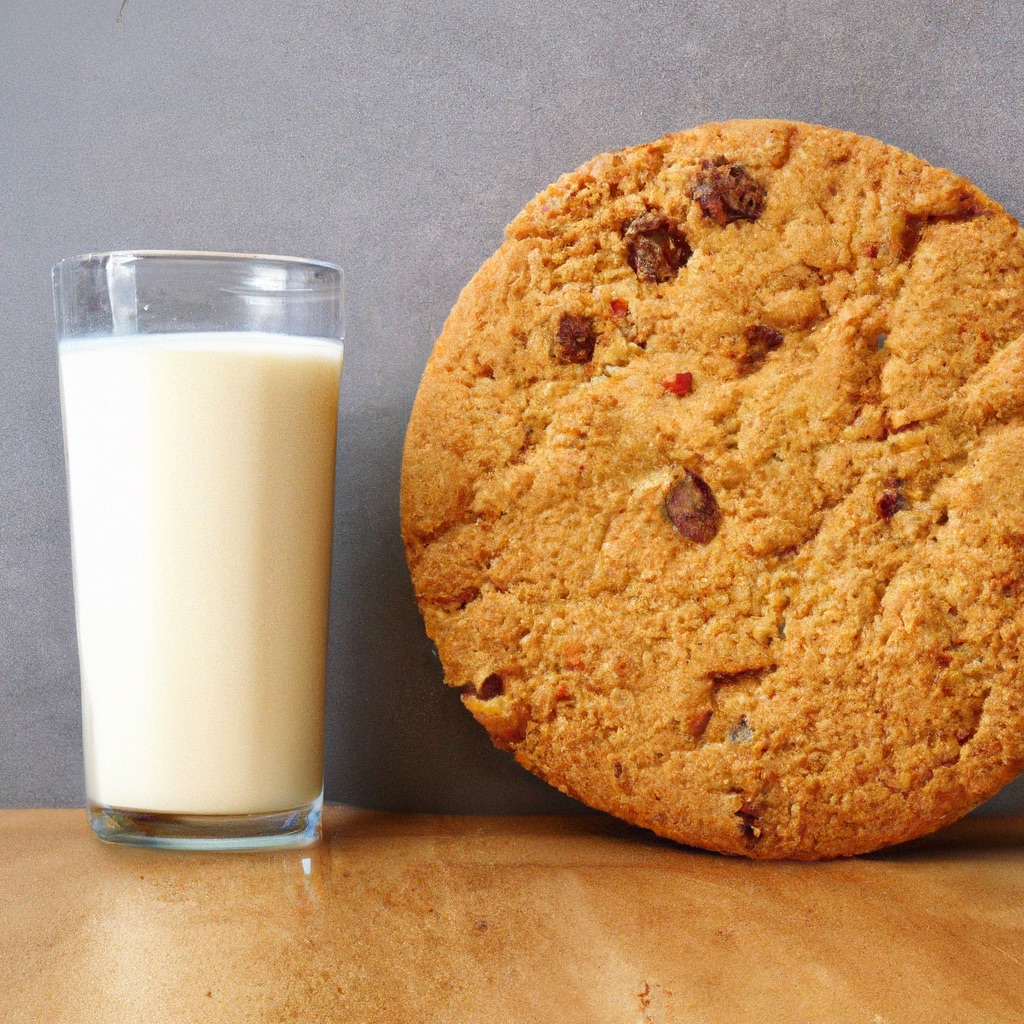} \\
        \rule{0pt}{0.0pt} \\
        \includegraphics[width=0.28\textwidth]{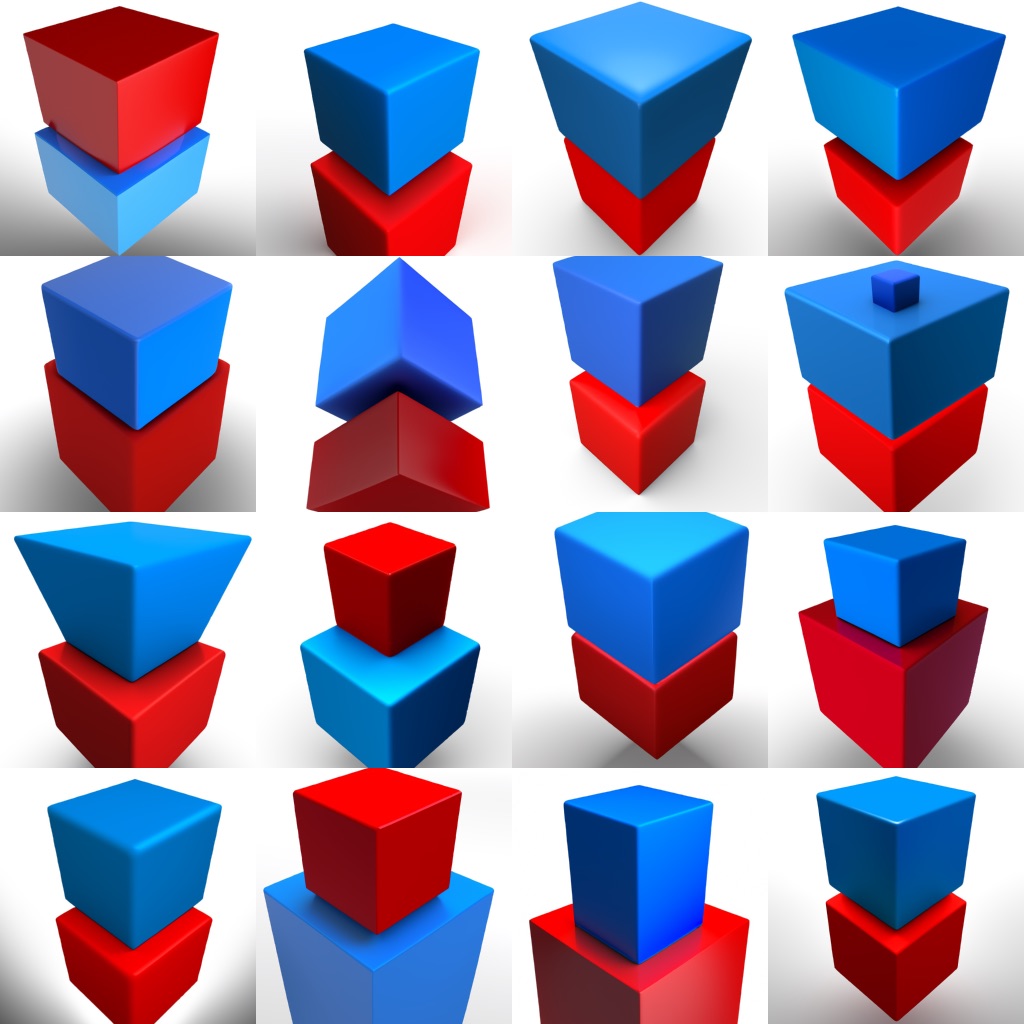} & \includegraphics[width=0.28\textwidth]{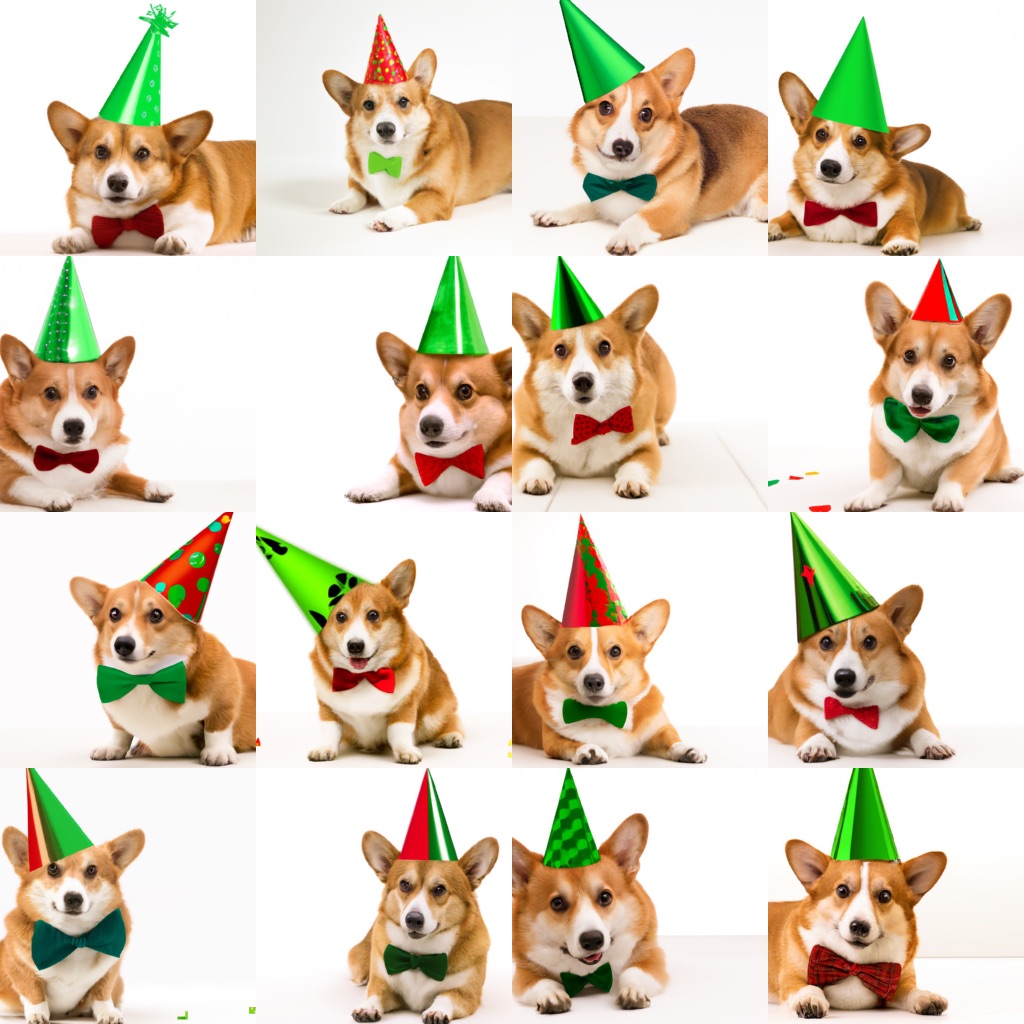} & \includegraphics[width=0.28\textwidth]{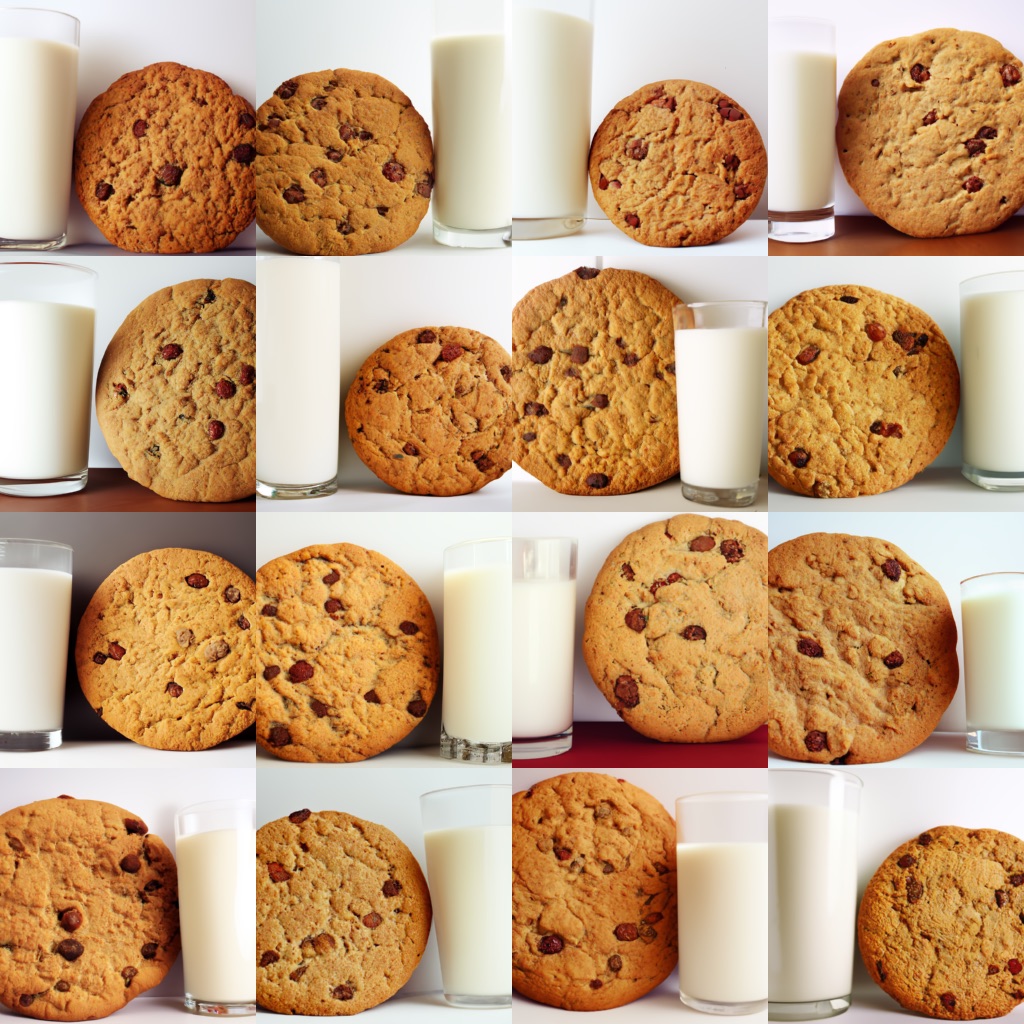}

        \rule{0pt}{0.5pt}
    \end{tabular}

    \caption{Reconstructions from the decoder for difficult binding problems. We find that the reconstructions mix up objects and attributes. In the first two examples, the model mixes up the color of two objects. In the rightmost example, the model does not reliably reconstruct the relative size of two objects.}
    \label{fig:binding_variations}
    \vskip -0.1in 
\end{figure*}

\begin{figure*}[t]
    \centering
    \setlength{\tabcolsep}{10.0pt}
    \begin{center}
    \includegraphics[width=0.85\textwidth]{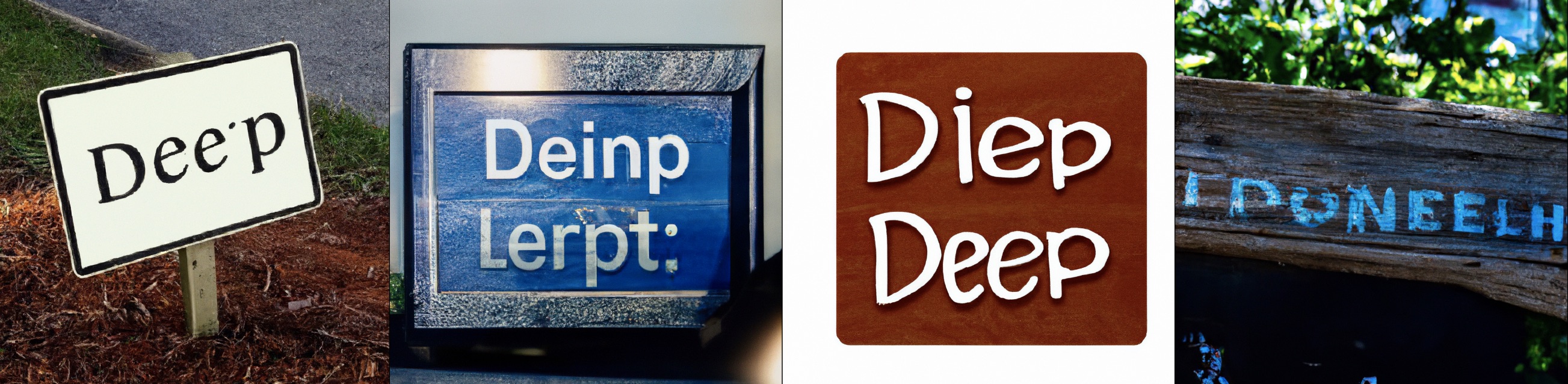}
    \end{center}

    \caption{Samples from \modelname{} for the prompt, ``A sign that says deep learning.''}
    \label{fig:ocr_samples}
\end{figure*}

\begin{figure}[ht!]
    \begin{center}
    \begin{subfigure}{0.85\textwidth}
        \centering
        \includegraphics[width=\textwidth]{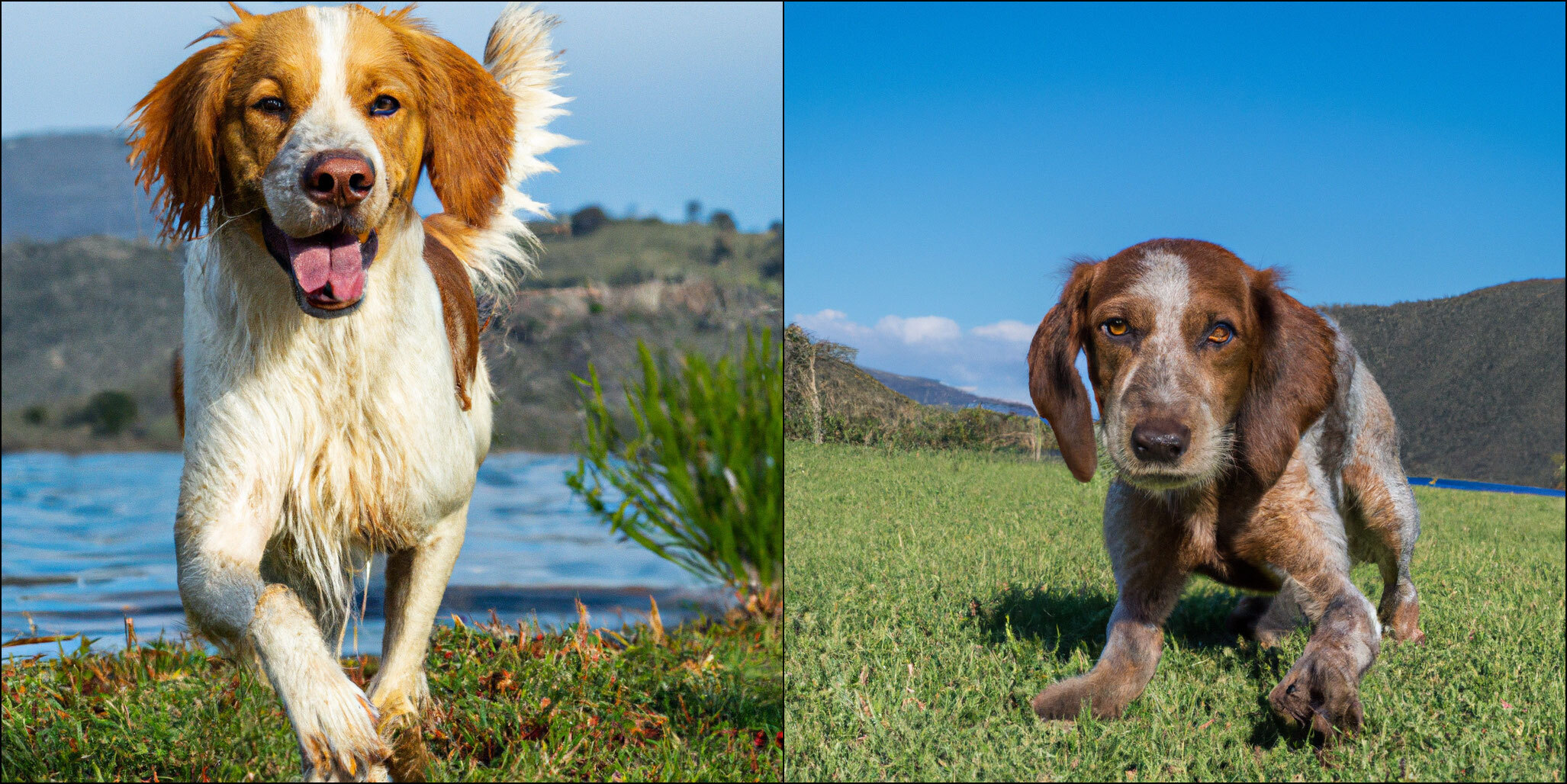}
        \caption{A high quality photo of a dog playing in a green field next to a lake.}
    \end{subfigure} \\
    \rule{0pt}{0.0pt} \\
    \begin{subfigure}{0.85\textwidth}
        \centering
        \includegraphics[width=\textwidth]{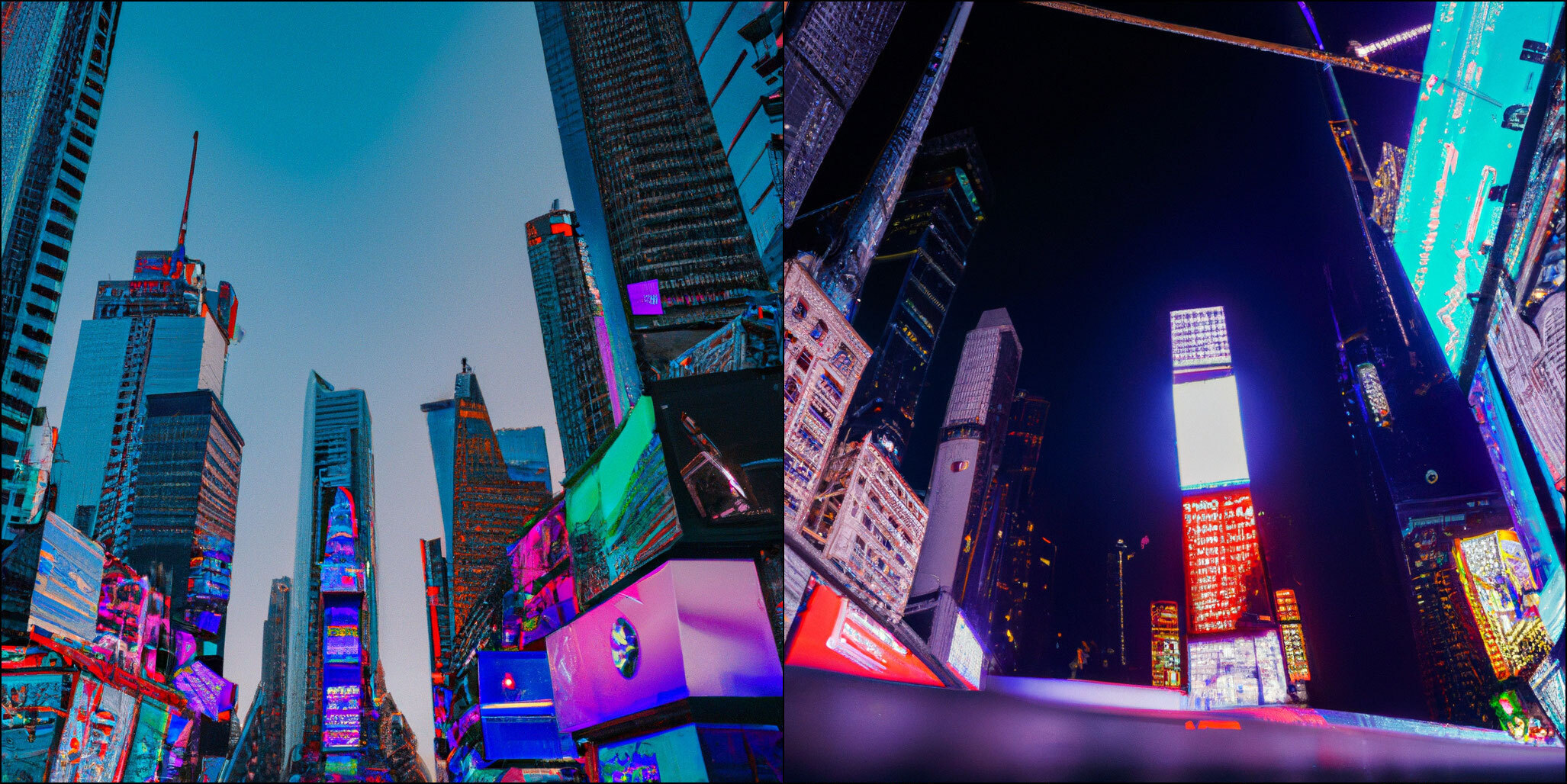}
        \caption{A high quality photo of Times Square.}
    \end{subfigure}
    \end{center}
    \caption{\modelname{} samples show low levels of detail for some complex scenes.}
    \label{fig:details}
    \vskip -0.2in
\end{figure}

\section{Limitations and Risks}
\label{sec:limitations}
Although conditioning image generation on CLIP embeddings improves diversity, this choice does come with certain limitations. In particular, \modelname{} is worse at binding attributes to objects than a corresponding GLIDE model. In Figure \ref{fig:block_stacking}, we find that \modelname{} struggles more than GLIDE with a prompt where it must bind two separate objects (cubes) to two separate attributes (colors). We hypothesize that this occurs because the CLIP embedding itself does not explicitly bind attributes to objects, and find that reconstructions from the decoder often mix up attributes and objects, as shown in Figure \ref{fig:binding_variations}. A similar and likely related issue is that \modelname{} struggles at producing coherent text, as illustrated in Figure \ref{fig:ocr_samples}; it is possible that the CLIP embedding does not precisely encode spelling information of rendered text. This issue is likely made worse because the BPE encoding we use obscures the spelling of the words in a caption from the model, so the model needs to have independently seen each token written out in the training images in order to learn to render it.

We also note that our stack still has a hard time producing details in complex scenes (Figure \ref{fig:details}). We hypothesize that this is a limitation of our decoder hierarchy producing an image at a base resolution of $64 \times 64$ and then upsampling it. Training our \modelname{} decoder at a higher base resolution should be able to alleviate this, at the cost of additional training and inference compute. 

As discussed in the GLIDE paper, image generation models carry risks related to deceptive and otherwise harmful content. \modelname{}'s performance improvements also raise the risk profile over GLIDE. As the technology matures, it leaves fewer traces and indicators that outputs are AI-generated, making it easier to mistake generated images for authentic ones and vice versa. More research is also needed on how the change in architecture changes how the model learns biases in training data.

The risks of these models should be assessed in relation to the particular deployment context, which includes training data, guardrails in place, the deployment space, and who will have access. A preliminary analysis of these issues in the context of the DALL·E 2 Preview platform (the first deployment of an unCLIP model), can be found in \namecite{mishkin2022risks}.

\section{Acknowledgements}
We’d like to thank Jong Wook Kim,  Hyeonwoo Noh, Alec Radford, Pranav Shyam, and Ilya Sutskever for helpful discussions and contributions to our work. We'd also like to thank Yunxin Jiao for creating several figures used in the paper. We are grateful to the Acceleration and Supercomputing teams at OpenAI for their work on software and hardware infrastructure this project used. 

\setcitestyle{numbers}
\bibliographystyle{plainnat}
\bibliography{main.bib}

\clearpage
\appendix
\section{Linear Probes for Evaluations}
\label{app:linear_probes}

For our evaluations, we leverage two new linear probes on top of a CLIP ViT-L/14 \shortcite{vit} model. To automate aesthetic quality evaluations, we follow the procedure used by \namecite{avaprobe}, training a linear regression model on images and mean ratings from the AVA dataset \shortcite{avadataset}. To reduce the cost of hyperparameter sweeps before conducting human evaluations, we train a logistic regression model to predict win probabilities between pairs of images. To train this model, we used 15,000 pairwise image comparisons gathered from all of our previous human evaluations. For each comparison~$i$, we computed CLIP image embeddings~$x_i$ and~$y_i$ for the two images in the pair. We then trained a linear model~$f(x)$ such that $1 / (1+\exp{(f(x_i) - f(y_i))})$ approximates the probability that a human prefers the image for~$y_i$. This can be reduced to a logistic regression problem with inputs equal to $y_i - x_i$.

\section{Error Bars for Human Evaluation}
\label{app:error_bars}

When computing error bars for human evaluations, we use the normal approximation interval with~$p=0.95$. We expect the normal approximation to be accurate for such a large sample size of~$n=1000$.

\section{Training Details}
\label{app:hps}
The unCLIP models used for the experiments in this paper were trained with the hyperparameters described below, unless otherwise noted. We additionally trained a production version of unCLIP using similarly sized models but with modified architectures and trained for longer; we include changes to accommodate product and safety requirements (e.g. inpainting, preventing unwanted memorization), and train on a larger dataset that is filtered for aesthetic quality and safety. We report model and training hyperparameters for the paper models in Table~\ref{tab:hps}. All models were trained using Adam \shortcite{adam} with corrected weight decay~\shortcite{adamw} and momentum $\beta_1 = 0.9$.

Our CLIP model uses a ViT-H/16~\shortcite{vit} image encoder that consumes $256 \times 256$ resolution images, and has width~1280 with 32~Transformer~\shortcite{transformer} blocks. The text encoder also follows the architecture described in~\namecite{clip}: it is a Transformer~\shortcite{transformer} with a causal attention mask, with width~1024 and 24~Transformer blocks. Both models are trained with learning rate~$3\times10^{-4}$ and SAM~\shortcite{sam} with~$\rho = 0.1$, where the perturbations are applied independently by the replicas, each of which uses batch size~64. The remaining hyperparameters are the same as those reported in~\namecite{clip}.

When training the encoder, we sample from the CLIP \shortcite{clip} and DALL-E \shortcite{dalle} datasets (approximately 650M~images in total) with equal probability. When training the decoder, upsamplers, and prior, we use only the DALL-E dataset~\shortcite{dalle} (approximately 250M~images). Incorporating the noisier CLIP dataset while training the generative stack negatively impacted sample quality in our initial evaluations. 

Our decoder architecture is the 3.5~billion parameter GLIDE model, with the same architecture and diffusion hyperparameters as in~\namecite{glide}. We train with learned sigma and sample with $250$ strided sampling steps as in \namecite{improved}. 

We use the ADMNet architecture \shortcite{sotapaper} for the upsamplers. In the first upsampling stage, we use a cosine noising schedule, $320$~channels and a depth of $3$~resblocks per resolution inside the ADMNet. We also apply gaussian blur (kernel size~$3$, sigma~$0.6$) as described in~\namecite{sr3}. In the second upsampling stage, we use a linear noising schedule, $192$~channels, a depth of $2$~resblocks per resolution, and train with the BSR degradation from~\namecite{latentdiffusion}. Neither upsampler uses attention. To reduce inference time, we use DDIM~\shortcite{ddim} and manually tune the number of steps, with 27~steps for $256 \times 256$ model, and 15~steps for the $1024 \times 1024$ model.

For the AR~prior, we use a Transformer text encoder with width~$2048$ and 24~blocks and a decoder with a causal attention mask, width~$1664$, and 24~blocks. For the diffusion prior, we use a Transformer with width~$2048$ and 24~blocks, and sample with Analytic DPM~\shortcite{analyticddpm} with 64~strided sampling steps. To reuse hyperparameters tuned for diffusion noise schedules on images from \namecite{sotapaper}, we scale the CLIP embedding inputs by~$17.2$ to match the empirical variance of RGB~pixel values of ImageNet images scaled to~$[-1, 1]$. 
\begin{table}[h]
    \setlength\tabcolsep{4pt}
    \begin{center}
    \begin{small}
    \begin{tabular}{lccccc}
    \toprule
                         & AR prior & Diffusion prior & $64$ & $64 \rightarrow 256$ & $256 \rightarrow 1024$\\
    \midrule
    Diffusion steps      & -         & 1000    & 1000    & 1000    & 1000 \\
    Noise schedule       & -         & cosine  & cosine  & cosine  & linear \\
    Sampling steps       & -         & 64      & 250     & 27      & 15 \\
    Sampling variance method & -     & analytic \shortcite{analyticddpm} & learned \shortcite{improved} & DDIM \shortcite{ddim}    & DDIM \shortcite{ddim} \\
    Crop fraction        & -         & -       & -       & 0.25    & 0.25 \\
    Model size           & 1B        & 1B      & 3.5B    & 700M    & 300M \\
    Channels             & -         & -       & 512     & 320     & 192 \\
    Depth                & -         & -       & 3       & 3       & 2 \\
    Channels multiple    & -         & -       & 1,2,3,4 & 1,2,3,4 & 1,1,2,2,4,4 \\
    Heads channels       & -         & -       & 64      & -       & - \\
    Attention resolution & -         & -       & 32,16,8 & -       & - \\
    Text encoder context    & 256       & 256     & 256     & -       & - \\
    Text encoder width      & 2048      & 2048    & 2048    & -       & - \\
    Text encoder depth      & 24        & 24      & 24      & -       & - \\ 
    Text encoder heads      & 32        & 32      & 32      & -       & - \\
    Latent decoder context  & 384       & -       & -       & -       & - \\
    Latent decoder width    & 1664      & -       & -       & -       & - \\
    Latent decoder depth    & 24        & -       & -       & -       & - \\ 
    Latent decoder heads    & 26        & -       & -       & -       & - \\
    Dropout              & -         & -       & 0.1     & 0.1     & - \\
    Weight decay         & 4.0e-2    & 6.0e-2  & -       & -       & - \\
    Batch size           & 4096      & 4096    & 2048    & 1024    & 512 \\
    Iterations           & 1M        & 600K    & 800K    & 1M      & 1M \\
    Learning rate        & 1.6e-4    & 1.1e-4  & 1.2e-4  & 1.2e-4  & 1.0e-4 \\
    Adam $\beta_2$       & 0.91      & 0.96    & 0.999   & 0.999   & 0.999 \\
    Adam $\epsilon$      & 1.0e-10     & 1.0e-6   & 1.0e-8    & 1.0e-8    & 1.0e-8  \\
    EMA decay            & 0.999     & 0.9999  & 0.9999  & 0.9999  & 0.9999 \\
    \bottomrule
    \end{tabular}
    \end{small}
    \end{center}
    \caption{Hyperparameters for the models}
    \label{tab:hps}
    \vskip -0.2in
\end{table}

\clearpage
\section{Random samples}
\begin{figure*}[h]
    \begin{center}
    \includegraphics[width=\textwidth]{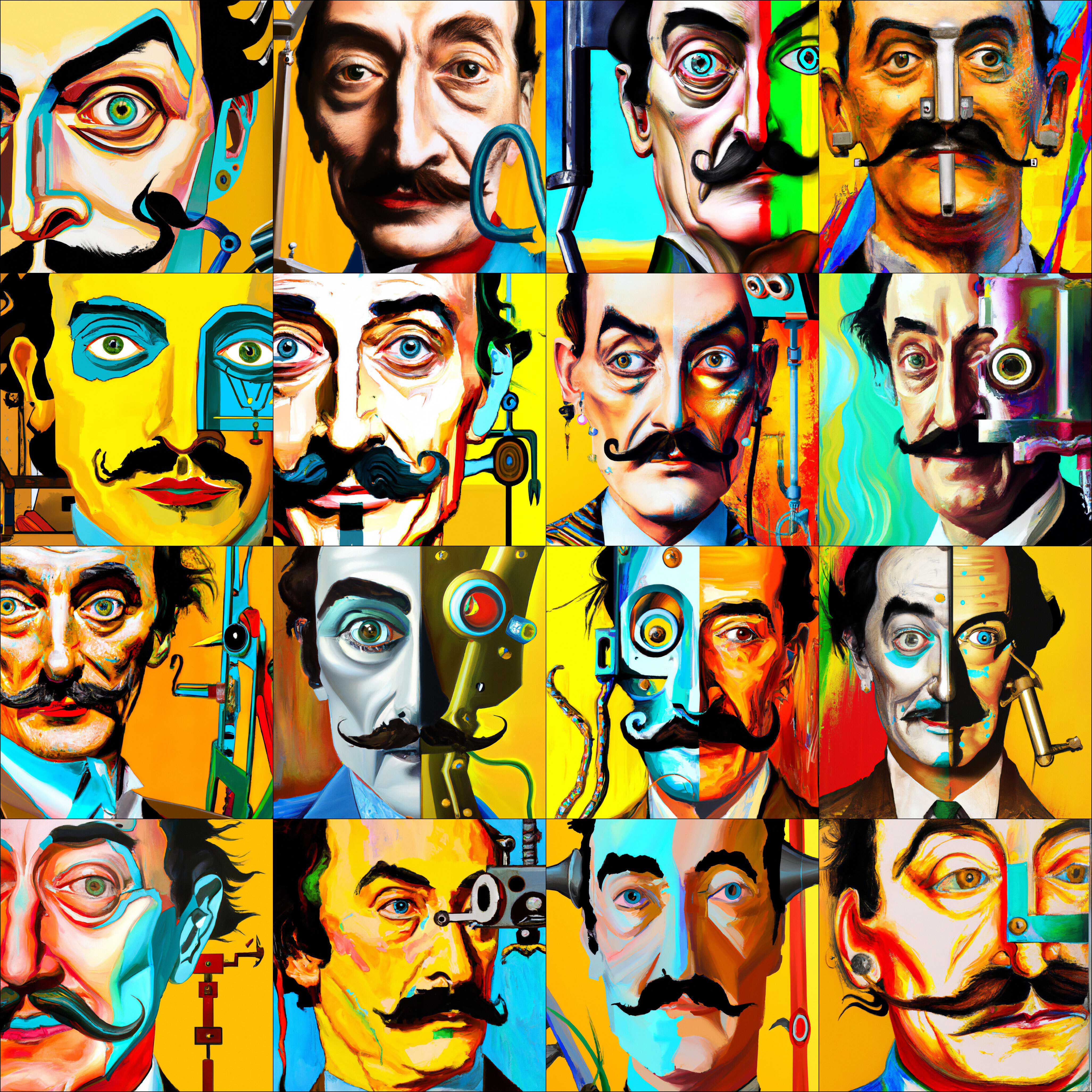}
    \end{center}
    \caption{Random samples from \modelname{} for prompt ``Vibrant portrait painting of Salvador Dali with a robotic half face''}
    \label{fig:rnd1} 
    \vskip -0.1in
\end{figure*}

\begin{figure*}[h]
    \begin{center}
    \includegraphics[width=\textwidth]{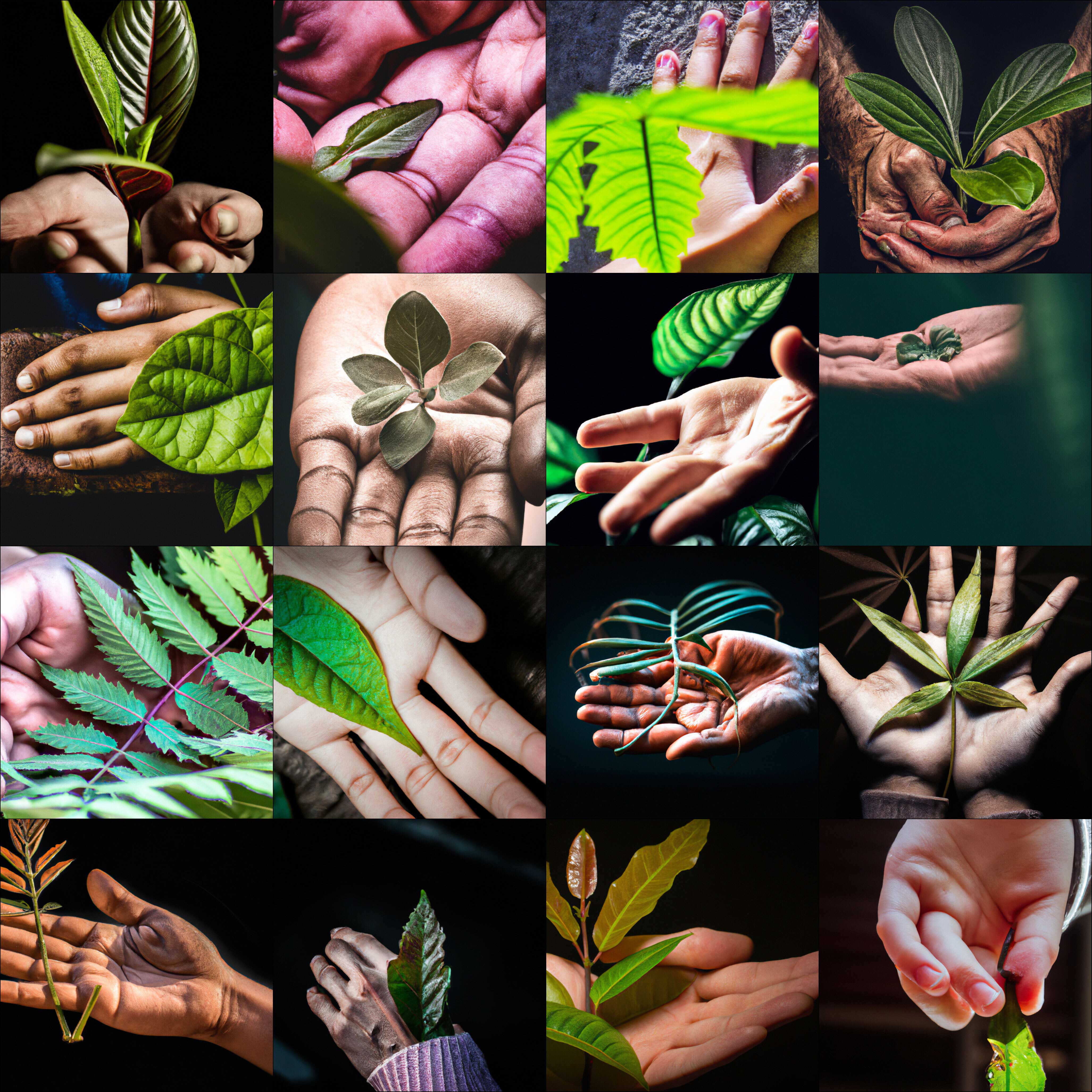}
    \end{center}
    \caption{Random samples from \modelname{} for prompt ``A close up of a handpalm with leaves growing from it.''}
    \label{fig:rnd2}
    \vskip -0.1in
\end{figure*}

\begin{figure*}[h]
    \begin{center}
        \includegraphics[width=\textwidth]{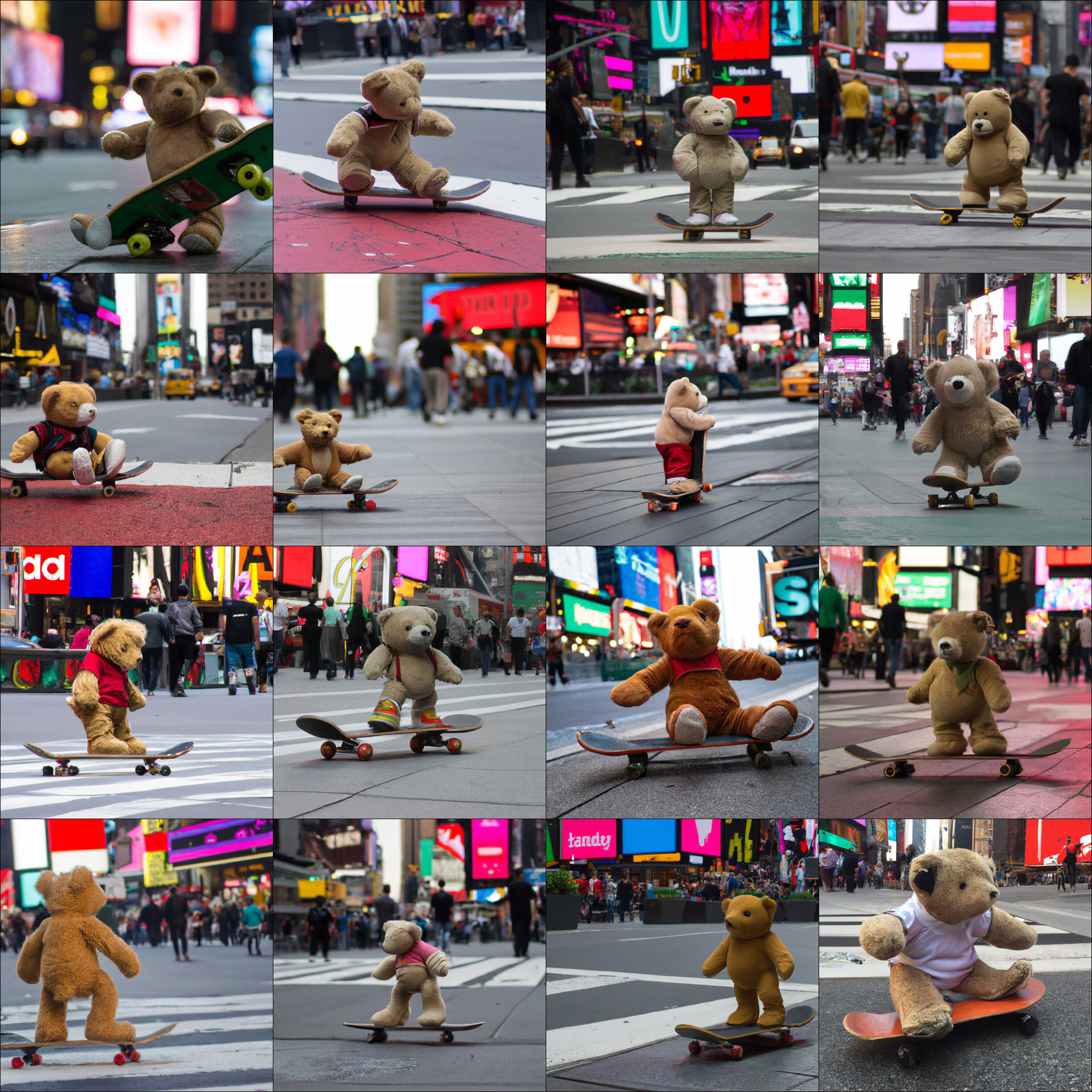}
    \end{center}
    \caption{Random samples from \modelname{} for prompt ``A teddybear on a skateboard in Times Square.''}
    \label{fig:rnd3}
    \vskip -0.1in
\end{figure*}

In Figures \ref{fig:rnd1}, \ref{fig:rnd2} and \ref{fig:rnd3} we show random samples from our production model for some of the prompts from Figure \ref{fig:header_samples}. 

\end{document}